%% file: main.tex
\documentclass[pdflatex,sn-aps]{sn-jnl}

\usepackage{setspace}

\usepackage{amsfonts}
\usepackage{amsmath, bm}
\usepackage{upgreek}
\usepackage{graphicx}

\renewcommand{\thesubsection}{\Alph{subsection}}
\makeatletter
\renewcommand{\p@subsection}{}
\makeatother

\usepackage[dvipsnames]{xcolor}

\input{macros}

\usepackage{graphicx}%
\usepackage{multirow}%
\usepackage{amsmath,amssymb,amsfonts}%
\usepackage{amsthm}%
\usepackage{mathrsfs}%
\usepackage[title]{appendix}%

\renewcommand{\thesubsection}{\Alph{subsection}.}
\makeatletter
\renewcommand{\p@subsection}{}
\makeatother

\usepackage{textcomp}%
\usepackage{manyfoot}%
\usepackage{booktabs}%
\usepackage{algorithm}%
\usepackage{algorithmicx}%
\usepackage{algpseudocode}%
\usepackage{listings}%

\theoremstyle{thmstyleone}%

\theoremstyle{thmstyletwo}%

\theoremstyle{thmstylethree}%

\begin{document}

\title{\bf\large Radio-Frequency Convolutional Neural Networks}

\author{\normalsize Zhihui Gao\textsuperscript{1}, Shi-Yuan Ma\textsuperscript{2}, Yiran Chen\textsuperscript{1}, Dirk Englund\textsuperscript{2}, and Tingjun Chen\textsuperscript{1}}

\affil[1]{\small\it Department of Electrical and Computer Engineering, Duke University, Durham, NC 27708, USA}

\affil[2]{\small\it Research Laboratory of Electronics, Massachusetts Institute of Technology, Cambridge, MA 02139, USA}

\maketitle

\input{tex/abstract}

\input{tex/article}

\input{tex/methods}

\input{tex/acks}

\clearpage

{
\centering{\bf\large Supplementary Information: \\
Radio-Frequency Convolutional Neural Networks}
\vspace{2.0ex}

\centering{
Zhihui Gao\textsuperscript{1}, Shi-Yuan Ma\textsuperscript{2}, Yiran Chen\textsuperscript{1}, Dirk Englund\textsuperscript{2}, and Tingjun Chen\textsuperscript{1}}
\vspace{1.0ex}

\centering{
\textsuperscript{1}\textit{Department of Electrical and Computer Engineering, Duke University, Durham, NC 27708, USA}}

\centering{
\textsuperscript{2}\textit{Research Laboratory of Electronics, Massachusetts Institute of Technology, Cambridge, MA 02139, USA}}

}

\setcounter{section}{0}
\setcounter{page}{1}
\vspace{2.0ex}
\tableofcontents

\newpage

\setcounter{figure}{0}
\renewcommand{\thefigure}{S\arabic{figure}}
\setcounter{table}{0}
\renewcommand{\thetable}{S\arabic{table}}
\setcounter{equation}{0}
\renewcommand{\theequation}{S\arabic{equation}}
\setcounter{subsection}{0}
\renewcommand{\thesubsection}{\Alph{subsection}}

\onehalfspacing

\clearpage
\input{tex/Supplementary_theory}

\clearpage
\input{tex/Supplementary_experiment}

\clearpage
\input{tex/Supplementary_figure}

\bibliography{reference}

\end{document}

%% file: macros.tex
\newcommand{\autorefmethod}{\hyperref[sec:methods]{Methods}~section}

\newcommand{\myparatight}[1]{\vspace{1.0ex}\noindent\textbf{#1.~~}}
\newcommand{\supplementary}{Supplementary Text Section}

\newcommand{\iu}{{j}}
\newcommand{\eu}{{e}}
\newcommand{\complexity}{\mathcal{O}\!}
\newcommand{\abs}[1]{\left|{#1}\right|}
\newcommand{\ang}[1]{\angle{#1}}
\newcommand{\conj}[1]{\overline{#1}}

\newcommand{\fft}[1]{\textsf{FFT}\left\{#1\right\}}
\newcommand{\ifft}[1]{\textsf{iFFT}\left\{#1\right\}}

\newcommand{\powerTx}{P_{x}}
\newcommand{\PSDTx}{p_{x}}

\newcommand{\name}{{\sc RF-CNN}}
\newcommand{\namebf}{{\sc\textbf{RF-CNN}}}
\newcommand{\convonedim}{{\sf conv1d}}
\newcommand{\convonedimbf}{{\sf \textbf{conv1d}}}
\newcommand{\convtwodim}{{\sf conv2d}}
\newcommand{\convtwodimbf}{{\sf \textbf{conv2d}}}

\newcommand{\hiddenInNum}{H_{\textrm{in}}}
\newcommand{\hiddenOutNum}{H_{\textrm{out}}}

\newcommand{\inputTensor}{\mathbf{X}}
\newcommand{\inputScalar}[1]{x[#1]}
\newcommand{\inputElem}[1]{x_{#1}}
\newcommand{\weightTensor}{\mathbf{W}}
\newcommand{\weightScalar}[1]{w[#1]}
\newcommand{\weightElem}[1]{w_{#1}}
\newcommand{\outputTensor}{\mathbf{Y}}
\newcommand{\outputTensorEst}{\widehat{\mathbf{Y}}}
\newcommand{\outputScalar}[1]{y[#1]}
\newcommand{\outputScalarEst}[1]{\widehat{y}[#1]}
\newcommand{\outputElem}[1]{y_{#1}}
\newcommand{\chanNum}{C}
\newcommand{\chanInNum}{C_{\textrm{in}}}
\newcommand{\chanInIdx}{c_{\textrm{in}}}
\newcommand{\chanOutNum}{C_{\textrm{out}}}
\newcommand{\chanOutIdx}{c_{\textrm{out}}}
\newcommand{\imageSize}{I}
\newcommand{\imageIdx}{i}
\newcommand{\imageSizeX}{I_{h}}
\newcommand{\imageIdxX}{i_{h}}
\newcommand{\imageSizeY}{I_{w}}
\newcommand{\imageIdxY}{i_{w}}
\newcommand{\kernelSize}{K}
\newcommand{\kernelIdx}{k}
\newcommand{\kernelSizeX}{K_{h}}
\newcommand{\kernelIdxX}{k_{h}}
\newcommand{\kernelSizeY}{K_{w}}
\newcommand{\kernelIdxY}{k_{w}}
\newcommand{\paddingSize}{P}
\newcommand{\paddingSizeX}{P_{h}}
\newcommand{\paddingSizeY}{P_{w}}

\newcommand{\conv}[1]{\textsf{conv}\left( #1 \right)}
\newcommand{\convDimenNum}{D}

\newcommand{\convInputVec}{\mathbf{S}_1}
\newcommand{\convInputScalar}{S_1}
\newcommand{\convInputLen}{L_1}
\newcommand{\convWeightVec}{\mathbf{S}_2}
\newcommand{\convWeightScalar}{S_2}
\newcommand{\convWeightLen}{L_2}
\newcommand{\convOutputVec}{\mathbf{S}_o}
\newcommand{\convOutputScalar}{S_o}
\newcommand{\convOutputLen}{L_o}
\newcommand{\convolution}{\ast}
\newcommand{\correlation}{\star}
\newcommand{\convIdx}{l}
\newcommand{\convIdxTemp}{l^{\prime}}
\newcommand{\convSampInputVec}{\mathbf{s}_1}
\newcommand{\convSampInputScalar}{s_1}
\newcommand{\convWaveInput}{s_1}
\newcommand{\convSampWeightVec}{\mathbf{s}_2}
\newcommand{\convWaveWeight}{s_2}
\newcommand{\convSampOutputVec}{\mathbf{s}_o}
\newcommand{\convSampOutputScalar}{s_o}
\newcommand{\convWaveOutput}{s_o}
\newcommand{\convOutputSubVec}{\mathbf{S}_{\downarrow}}
\newcommand{\convOutputSubScalar}{S_{\downarrow}}
\newcommand{\convOutputSubLen}{L_{\downarrow}}
\newcommand{\convOutputSubStart}{l_1}
\newcommand{\convOutputSubEnd}{l_2}
\newcommand{\convSampOutputSubVec}{\mathbf{s}_{\downarrow}}
\newcommand{\convSampOutputSubScalar}{s_{\downarrow}}
\newcommand{\convWaveOutputSub}{s_{\downarrow}}
\newcommand{\convWaveNoise}{n}
\newcommand{\PSDNoise}{p_n}

\newcommand{\convChannelFunc}[1]{\mathcal{H}\left( #1 \right)}
\newcommand{\convChannelInputVec}{\mathbf{H}_1}
\newcommand{\convChannelInputScalar}{H_1}
\newcommand{\convChannelWeightVec}{\mathbf{H}_2}
\newcommand{\convChannelWeightScalar}{H_2}
\newcommand{\convChannelOutputVec}{\mathbf{H}_o}
\newcommand{\convChannelOutputScalar}{H_o}
\newcommand{\convCalibInputVec}{\mathbf{C}_1}
\newcommand{\convCalibInputScalar}{C_1}
\newcommand{\convCalibWeightVec}{\mathbf{C}_2}
\newcommand{\convCalibWeightScalar}{C_2}
\newcommand{\convCalibOutputVec}{\mathbf{C}_o}
\newcommand{\convCalibOutputScalar}{C_o}

\newcommand{\symNum}{K}
\newcommand{\symIdx}{k}
\newcommand{\packMat}{\mathcal{S}}
\newcommand{\packScalar}{S}
\newcommand{\packPreMat}{\mathcal{S}^{\textrm{(pre)}}}
\newcommand{\packPreScalar}{S^{\textrm{(pre)}}}
\newcommand{\packPostMat}{\mathcal{S}^{\textrm{(post)}}}
\newcommand{\packPostScalar}{S^{\textrm{(post)}}}
\newcommand{\packPayMat}{\mathcal{S}^{\textrm{(payload)}}}
\newcommand{\packPayScalar}{S^{\textrm{(payload)}}}
\newcommand{\packExpMat}{\mathcal{S}^{\prime}}
\newcommand{\packExpScalar}{S^{\prime}}
\newcommand{\packPreExpMat}{\mathcal{S}^{\textrm{(pre)}\prime}}
\newcommand{\packPreExpScalar}{S^{\textrm{(pre)}\prime}}
\newcommand{\packPostExpMat}{\mathcal{S}^{\textrm{(post)}\prime}}
\newcommand{\packPostExpScalar}{S^{\textrm{(post)}\prime}}
\newcommand{\packPayExpScalar}{S^{\textrm{(payload)}\prime}}
\newcommand{\freqOffset}{\Delta F}
\newcommand{\timeOffset}{\Delta \tau}

\newcommand{\convSampVec}{\mathbf{s}}
\newcommand{\convSampScalar}{s}
\newcommand{\convLen}{L}
\newcommand{\convWave}{s}
\newcommand{\bit}{b}
\newcommand{\sqnr}[1]{\textsf{SQNR}\left\{#1\right\}}
\newcommand{\papr}[1]{\textsf{PAPR}\left\{#1\right\}}
\newcommand{\backoff}{\alpha}
\newcommand{\zadoffVec}{\mathbf{\Phi}^{(\textrm{ZC})}}
\newcommand{\zadoffScalar}{\phi^{(\textrm{ZC})}}

\newcommand{\layerIdx}{l}

\newcommand{\specInputVec}{\mathbf{S}_{x}}
\newcommand{\specInputScalar}{S_{x}}
\newcommand{\specInputLen}{L_x}
\newcommand{\specWeightVec}{\mathbf{S}_{w}}
\newcommand{\specWeightScalar}{S_{w}}
\newcommand{\specWeightLen}{L_w}
\newcommand{\specOutputVec}{\mathbf{S}_{y}}
\newcommand{\specOutputScalar}{S_{y}}
\newcommand{\specOutputLen}{L_y}
\newcommand{\specOutputStart}{l_1}
\newcommand{\specOutputEnd}{l_2}

\newcommand{\waveLen}{T}
\newcommand{\waveIdx}{t}
\newcommand{\freqSub}{\Delta f}
\newcommand{\band}{B}
\newcommand{\bandInput}{B_x}
\newcommand{\bandWeight}{B_W}
\newcommand{\bandOutput}{B_y}
\newcommand{\energyMVM}{E}
\newcommand{\energyMVMEnc}{E_{\textrm{enc}}}
\newcommand{\energyMVMDAC}{E_{\textrm{dac}}}
\newcommand{\energyMVMADC}{E_{\textrm{adc}}}
\newcommand{\energyMVMDec}{E_{\textrm{dec}}}
\newcommand{\energyMAC}{e}
\newcommand{\energyMACEnc}{e_{\textrm{enc}}}
\newcommand{\energyMACDAC}{e_{\textrm{dac}}}
\newcommand{\energyMACADC}{e_{\textrm{adc}}}
\newcommand{\energyMACDec}{e_{\textrm{dec}}}
\newcommand{\energyMACTDL}{e_{\textrm{TDL}}}
\newcommand{\throughput}{\Lambda}

\newcommand{\energyPerMAC}{\epsilon_{\textrm{mac}}}
\newcommand{\energyPerADC}{\epsilon_{\textrm{adc}}}

%% file: tex/abstract.tex
\section*{Abstract}

Running artificial intelligence (AI) models directly on edge devices such as smartphones, wearables, and drones offers low latency, pervasive scalability, and data privacy, but these devices rarely carry the computing capability that modern neural networks demand. 
Edge accelerators have been developed in response, yet each adds computing hardware to devices already constrained in size, weight, power, and cost (SWaP-C). 
An alternative lies in what these devices already carry: the frequency mixer in every wireless radio multiplies signals in time, natively performing convolution in the frequency domain. 
Here we introduce radio-frequency convolutional neural networks ({\name}s), which repurpose existing communication hardware for CNN inference. 
Multi-channel convolutions are mapped onto frequency tones for a passive mixer to execute in a single pass. 
We experimentally demonstrate that {\name} runs deep CNNs up to {26.4} million parameters and nine layers from classification of wireless signals and images to controllable image generation, close to full-precision performance.
Because the weights arrive over the air and the analog hardware is shared with communication, the edge device spends energy only on data preparation and readout--down to {0.72}\,femtojoules per multiply-accumulate, two orders of magnitude less than it would cost on an added digital processor.
These results suggest that deployed wireless infrastructure can bring efficient, state-of-the-art AI inference to the billions of devices it already connects.

%% file: tex/article.tex
\section*{Introduction}
\label{sec: main-introduction}

Artificial intelligence (AI) is advancing through scaling: machine learning (ML) models grow, and the computation they demand grows faster than hardware efficiency improves~\cite{sevilla2022compute,thompson2020computational}. Much of this intelligence is needed at the network edge, where a data center cannot serve it for reasons such as latency, privacy, or bandwidth, so inference must run where the data is produced~\cite{satyanarayanan2017emergence,zhou2019edge}. The devices at the edge---smartphones, wearables, drones---operate on fixed budgets of battery, heat, size, and cost~\cite{xu2018scaling}. The prevailing response has been to equip them with dedicated accelerators; yet every addition spends from the same budgets that define the edge, and the obstacle it faces is physical rather than algorithmic: within these budgets, the hardware the computation requires cannot be carried where the computation must occur.

\begin{figure*}[!t]
    \centering
    \includegraphics[width=1.0\columnwidth]{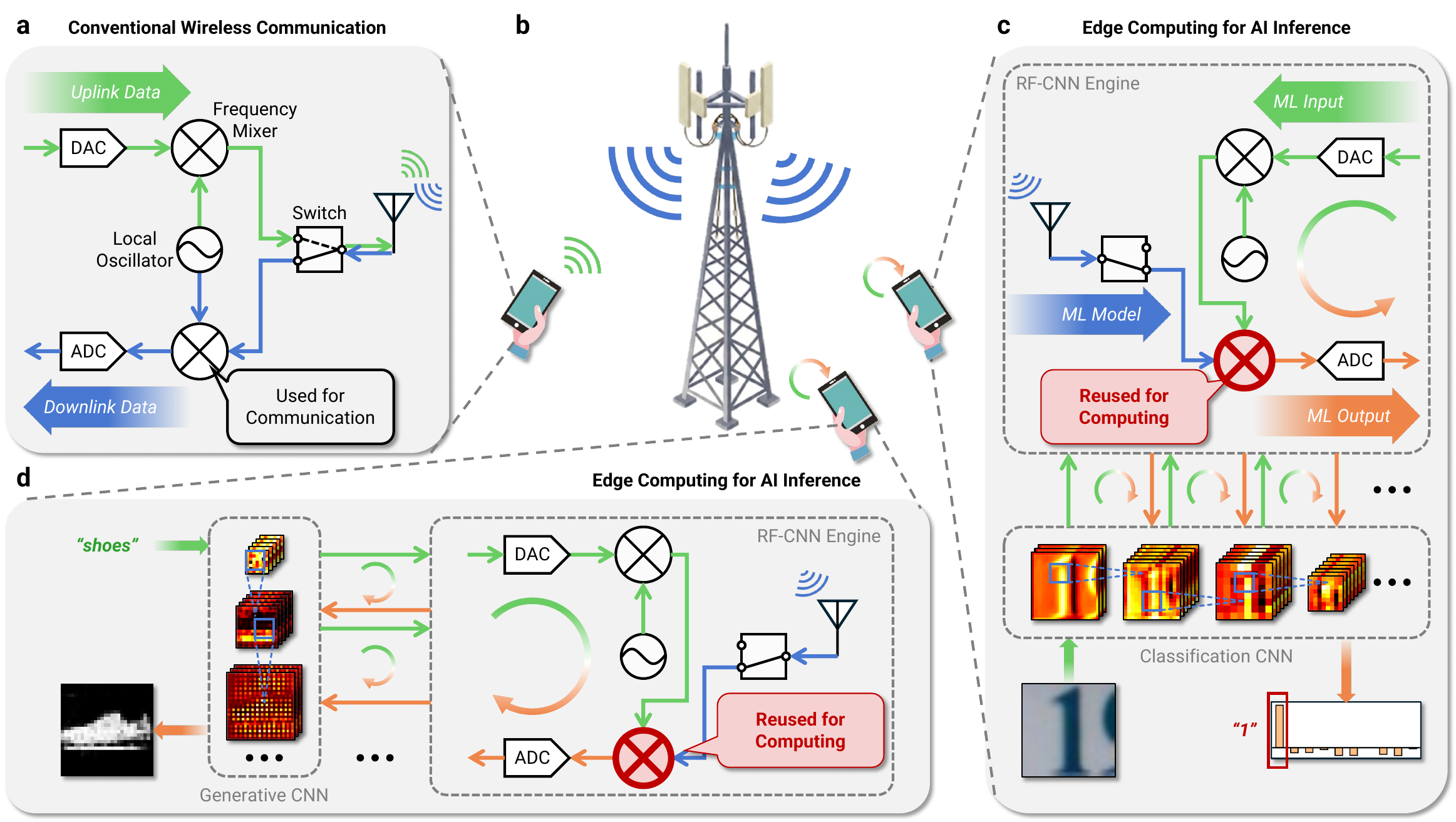}
    \caption{\textbf{{\namebf}: repurposing wireless communication hardware for edge AI computing.}
    \textbf{a}, In a conventional wireless transceiver on the edge devices, radio frequency (RF) front end components, including frequency mixers, digital-to-analog converters (DACs), and analog-to-digital converters (ADCs), are dedicated to uplink and downlink communications.
    \textbf{b}, In {\name}, a remote radio infrastructure (e.g., a 5G base station or Wi-Fi access point) can dynamically support either conventional wireless communication or distributed AI computation for nearby edge devices.
    \textbf{c}, The same RF hardware is repurposed to execute convolutional neural network (CNN) inference directly in the analog/RF domain: the wirelessly received signals are replaced by machine learning (ML) weights ($\weightTensor$), and the inputs ($\inputTensor$) and outputs ($\outputTensor$) signals are generated/received locally, enabling efficient classification tasks.
    \textbf{d}, The {\name} framework is also applicable to generative models, where RF hardware performs CNN-based feature transformations throughout the generation pipeline.
    }
    \label{fig:figure-architecture}
\end{figure*}

The hardware to resolve this constraint is, however, already carried by every edge device that communicates. Every wireless edge device contains radio frequency (RF) front ends including frequency mixers and local oscillators (LOs), data converters, and antennas~\cite{razavi2012rf,weinreich20220} (Fig.~\ref{fig:figure-architecture}a).
The frequency mixer multiplies two signals in the time domain and, by the convolution theorem, this multiplication is equivalent to a correlation in the frequency domain~\cite{oppenheim1999discrete}---the same operation communication systems use to translate signals between frequency bands.
It is also the operation that a convolutional layer performs, namely the cross-correlation of an input with a kernel. The operation that a frequency mixer executes is therefore the operation on which convolutional inference depends. A wireless front end is, without modification, a convolutional processor.

Yet this correspondence has not been fully exploited. The dedicated accelerators that have driven efficient edge inference---digital processors~\cite{jouppi2017datacenter,sze2017efficient,khwa2025mixed}, analog in-memory computing~\cite{yao2020fully,ambrogio2023analog,wan2022compute,zhang2023edge,jung2022crossbar,song2024programming}, and photonic processors~\cite{shen2017deep,feldmann2021parallel,xu202111tops,chen2023all,ahmed2025universal,hua2025integrated}---tailor the hardware to the model, each adding computing hardware alongside the device's existing components. A complementary body of work computes through an innate physical mechanism~\cite{grollier2020neuromorphic,wright2022deep,zolfagharinejad2025analogue,wu2026microring,ma2026machine}, where the computation is intrinsic to the physics; here the model is instead tailored to the mechanism, which need not match the operations a given ML architecture requires.

Computation has also been sought within wireless communication itself: in the channel between devices, and in the front ends inside them.
One route engineers the propagation environment so that the channel emulates a layer, using the multiple-access channel as an analog adder~\cite{nazer2007computation,zhu2021overtheair}, configuring intelligent surfaces to emulate convolutional and fully-connected layers~\cite{liu2022programmable,sanchez2023airnn,reusmuns2023airfc,hua2026cnns,stylianopoulos2026over}, or exploiting rich-scattering environments and time-modulated coding metasurfaces to compute in the wave domain~\cite{del2018leveraging,chen2026programmable}.
The other computes with the front-end circuitry itself, through reconfigurable RF or microwave circuits for analog linear algebra, including tunable matrix multipliers~\cite{zhu2024reconfigurable,keshavarz2025programmable}, microwave linear analog computers~\cite{nerini2025analog}, and oscillator-based microwave networks~\cite{govind2025integrated}. 
Most directly within this route, the passive mixer evaluates the matrix-vector multiplication of a fully-connected layer, with weights broadcast from a central radio~\cite{gao2026disaggregated} or analyzed in simulation~\cite{yu2026analog}. 
Across both routes, the computation is either emulated by engineering the radio environment, demonstrated only for single layers or shallow networks, or cast as a generic matrix-vector multiplication that discards the weight sharing and locality of convolution. 
The mixer's own correlation is not used to compute a convolutional layer directly.

Here we introduce \underline{R}adio-\underline{F}requency \underline{C}onvolutional \underline{N}eural \underline{N}etworks ({\name}s), which repurpose wireless communication hardware for edge inference of convolutional neural networks (CNNs) that underpin modern vision and signal processing~\cite{lecun2002gradient,simonyan2014very,lecun2015deep,liu2022convnet} (Fig.~\ref{fig:figure-architecture}b).
A passive frequency mixer, driven by a local input against weights received over the air from a central radio, performs a convolutional layer directly in the analog domain, using the mixer's native operation rather than an engineered emulation.
The converters, oscillators, and mixers that already serve communication are thereby reused as CNN accelerators, without dedicated hardware (Figs.~\ref{fig:figure-architecture}c--d). 

Because the mapping is native rather than approximate, it stays accurate as networks deepen: on a wireless testbed, {\name} runs wide CNNs with up to 1,024 channels, as well as deep CNNs with up to nine layers at test accuracy close to full-precision digital baselines, spanning classification and generative inference over both one- and two-dimensional inputs: classification of complex-valued wireless signals and of natural images such as CIFAR-10, and controllable image generation such as human faces. 
Because the computation is carried by reuse, this scheme avoids the two energy costs that dominate edge inference: weights are received over the air rather than stored and fetched on-device~\cite{horowitz2014computing,sze2017efficient}, and the front end is already powered for communication~\cite{weinreich20220}, so {\name} spends energy only on encoding/decoding and transmission/reception of the local input/output.
What remains is a sub-femtojoule-per-multiply-accumulate (fJ/MAC) budget, down to {0.72}\thinspace{fJ/MAC} on the edge side---more than two orders of magnitude below what the same inference would cost on an extra digital processor added to the device~\cite{reuther2025lincoln,anderson2023optical}.
These results show that deployed wireless infrastructure can be reused as a substrate for complete, deep convolutional inference at the edge, a step toward a unified communication and computing architecture for future networks~\cite{letaief2019roadmap,saad2020vision,feng2024integrated}.

\section*{Convolution by frequency mixing}

The architecture of {\name} is shown in Fig.~\ref{fig:figure-architecture}b. A central radio (e.g., a 5G base station or Wi-Fi access point) broadcasts the ML weights $\weightTensor$ to the edge devices in its range; each device generates its ML input $\inputTensor$ locally and obtains the output $\outputTensor$ through its own communication front end. The computation itself takes place in a passive frequency mixer, hereafter the computing mixer, into which the remotely received $\weightTensor$ and the locally transmitted $\inputTensor$ are fed (Fig.~\ref{fig:figure-scalability}a).
The operation the mixer performs is elementary. It multiplies its two incident waveforms in the time domain, and time-domain multiplication is frequency-domain correlation: encode complex-valued $\inputTensor$ and $\weightTensor$ as amplitudes and phases on two series of frequency tones, and the spectrum of the mixer's output contains, tone by tone, the cross-correlation of the two series ({\supplementary}~\ref{ssec: supplementary-theory-linear-convolution}).
A convolutional layer is a cross-correlation of the input with the kernel, so the mixer computes the layer by its physics; what remains is to arrange the high-dimensional tensors on the one-dimensional tone series.

This arrangement is the tone-mapping algorithm (Figs.~\ref{fig:figure-scalability}b--c), which casts a complex-valued, multi-channel, one- or two-dimensional convolution of arbitrary dimensions into a single one-dimensional cross-correlation. 
For brevity, we consider the {\convtwodim} layer with square kernels ($\kernelSize \times \kernelSize$) and images ($\imageSize \times \imageSize$), and the equal number of input and output channels as $\chanNum$ (see full derivation in {\supplementary}~\ref{ssec: supplementary-theory-convolution-basics}).
In {\name}, each input channel is unrolled row by row onto $\imageSize^2$ tones within one frequency sub-band, with zero-padded guard tones between rows so that correlations across row boundaries contribute nothing (Fig.~\ref{fig:figure-scalability}c). The $\chanNum$ input channels of $\inputTensor$, aligned with the corresponding channels of $\weightTensor$, then occupy $\chanNum$ adjacent sub-bands with $\chanNum\cdot\imageSize^2$ tones across the bandwidth (Fig.~\ref{fig:figure-scalability}b). The channel dimension is where the mapping pays off: every aligned sub-band pair deposits its correlation products onto the same output tones, so the sum over input channels is carried out by the mixing itself, and $\outputTensor$ emerges complete on a single output sub-band. One mixing operation thus executes an entire multi-channel convolution, accumulating $\kernelSize^2\cdot\chanNum$ products per output element; repeating it once per output channel, $\chanNum$ times in total with the same $\inputTensor$ against different $\weightTensor$, completes the layer.
Finally, the time domain waveforms can be transformed from/to these frequency-domain tone series by the (inverse) fast Fourier transform (IFFT/FFT).

\section*{Energy consumption: avoided costs and remaining costs}

Running {\name} on an edge device involves three energy contributions: delivering the ML weights, operating the analog RF hardware, and the digital processing at the analog-digital boundary. We argue that only the third should be charged to the edge device, and it is under this accounting that {\name} offers its energy benefit. 

First, the ML weights.
$\weightTensor$ is received over the air from the central radio rather than stored and fetched on-device, so the energy of weight storage and data movement, a dominant cost of digital accelerators~\cite{horowitz2014computing,sze2017efficient,anderson2023optical}, is not spent by the edge device (see channel and phase calibration in {\supplementary}s~\ref{ssec: supplementary-channel-calibration} and~\ref{ssec: supplementary-phase-calibration}). 
Second, the analog RF hardware. {\name} reuses the existing components of the communication transceiver, including the local oscillator and amplifiers, whose energy consumption is already expended for communication, thus incurring no additional cost for computing. 
Third, the digital boundary.
Encoding (the IFFT forming the time-domain waveform of $\inputTensor$, $\energyMACEnc$), the transmitting digital-to-analog converter (DAC) ($\energyMACDAC$), the receiving analog-to-digital converter (ADC) ($\energyMACADC$), and decoding ($\energyMACDec$) operate specifically for computation (Fig.~\ref{fig:figure-scalability}e); they define the energy per
MAC reported in this work:
\begin{align}
    \energyMAC = \energyMACEnc + \energyMACDAC + \energyMACADC + \energyMACDec.
\end{align}
This is an edge-side accounting: the first two contributions are displaced to the infrastructure or absorbed by communication.

Over-the-air weight delivery also provides additional flexibility to further reduce the counted terms: because the central radio can transmit $\weightTensor$ at high power, the power of $\inputTensor$ required for a given computing accuracy is reduced, saving energy on the edge device. Fig.~\ref{fig:figure-scalability}d quantifies this trade, mapping accuracy, expressed as effective number of bits (ENOB), over the power spectral densities (PSDs) of $\inputTensor$ and $\weightTensor$ (see full measurements in {\supplementary}~\ref{ssec: supplementary-experiment-benchmark}): raising the PSD of $\weightTensor$ reduces the PSD required of $\inputTensor$, so the input PSD serves as a single operating parameter through which energy is exchanged for accuracy, a property the end-to-end experiments below make use of at runtime.

Each counted term is further shared across the MACs a single mixing operation contains (Fig.~\ref{fig:figure-scalability}f). All four scale with $1/\kernelSize^2$, because one pass of the mixer's correlation covers every kernel position; $\energyMACEnc$, $\energyMACADC$, and $\energyMACDec$ decrease further with $\chanNum$, as the channel accumulation performed by the mixing spreads their cost; and $\energyMACDAC$ decreases with the reduced input PSD $\PSDTx$. The orthogonal frequency-division multiplexing (OFDM)-based waveforms support the tone counts this structure calls for, sustaining accumulation scales of $\kernelSize^2\cdot\chanNum$ in the magnitude of $10^4$ per operation. 

\section*{Single-layer accuracy and energy at scale}
 
We test this expectation on a software-defined radio testbed in a wireless setting, alongside numerical simulation ({\supplementary}s~\ref{ssec: supplementary-gaussian-simulation} and~\ref{ssec: supplementary-experiment-testbed}), asking two questions for
a single RF convolutional layer: how little energy a given accuracy requires, and how far the accuracy holds as the layer widens.
 
\begin{figure*}[!t]
    \centering
    \includegraphics[width=1.0\columnwidth]{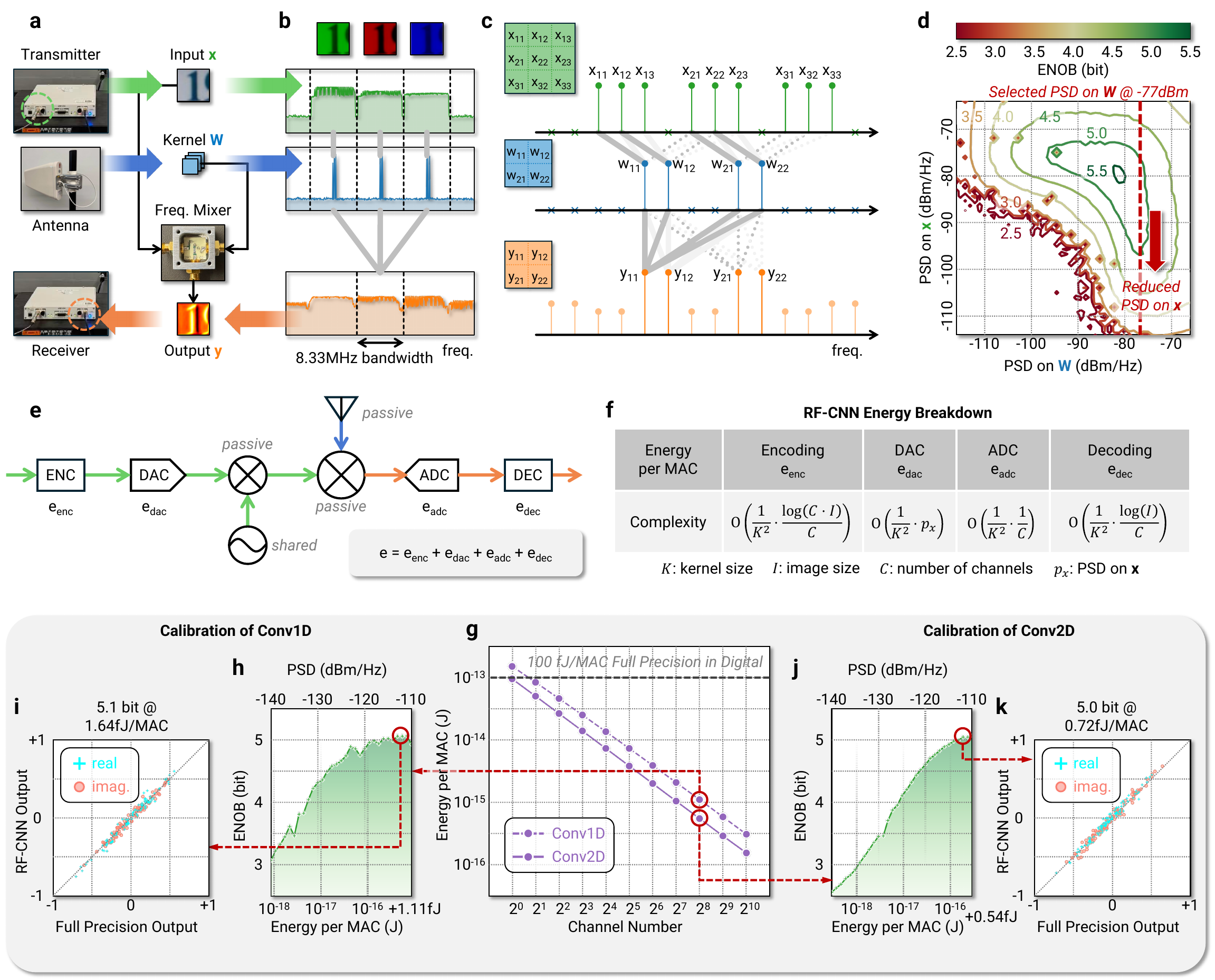}
    \caption{
    \textbf{The tone mapping algorithm and the characterization of a single convolutional layer.}
    \textbf{a}, On an edge device with {\name}, the ML input $\inputTensor$ transmitted by a radio TX and the ML weight $\weightTensor$ received through an antenna are mixed in a computing mixer to perform RF-domain correlation, and the output tensor $\outputTensor$ is received by a radio RX.
    \textbf{b}, Example operation of a single {\convtwodim} layer, where an RGB image ``1'' serving as $\inputTensor$ is mapped onto three frequency bands aligned with the three input channels of $\weightTensor$, and the output tensor $\outputTensor$ is extracted from the {8.33}\thinspace{MHz} output band after frequency-domain correlation.
    \textbf{c}, Illustration of the tone mapping algorithm within a single input channel, where the frequency-domain tones of $\inputTensor$ and $\weightTensor$ are arranged such that their correlation outputs include $\outputTensor$.
    \textbf{d}, Measured computing accuracy, expressed as effective number of bits (ENOB), over different power spectral density (PSD) combinations of $\inputTensor$ and $\weightTensor$.
    \textbf{e}, The energy consumption $\energyMAC$ is broken down into four modules: $\energyMACEnc$ for encoding, $\energyMACDAC$ for DAC, $\energyMACADC$ for ADC, and $\energyMACDec$ for decoding, while other modules are either passive or shared with other functionalities.
    \textbf{f}, The energy per multiply-accumulate (MAC) scaling of a {\convtwodim} layer, where $\chanNum$ is the number of input/output channels, $\kernelSize$ is the kernel size, and $\imageSize$ is the image size; $\PSDTx$ is the PSD of $\inputTensor$ that determines the SNR and thus the computing accuracy.
    \textbf{g}, Minimum energy per MAC operation required to achieve 4-bit ENOB for a single {\convonedim} or {\convtwodim} layer, showing that the energy per MAC decreases as the number of input/output channels increases.
    \textbf{h--j}, Measured ENOB as a function of energy per MAC for {\convonedim} (\textbf{h}) and {\convtwodim} (\textbf{j}) layers with {256} input/output channels.
    \textbf{i--k}, Comparison between the full-precision digital outputs and the outputs by {\name} for {\convonedim} (\textbf{i}) and {\convtwodim} (\textbf{k}), respectively, showing both real and imaginary components at an operating PSD of {$-$112}\thinspace{dBm/Hz}.
    }
    \label{fig:figure-scalability}
\end{figure*}

The expected scaling is what the measurements show.
Fig.~\ref{fig:figure-scalability}g plots the minimum energy per MAC for
4-bit ENOB against channel count: it follows $\complexity\left(\frac{1}{\kernelSize^2}\cdot\frac{\log\chanNum}{\chanNum}\right)$ for both layer types, with {\convtwodim} holding its kernel-area advantage over {\convonedim} throughout.
At $\chanNum=256$, a {\convtwodim} layer computes at {0.551}\thinspace{fJ/MAC} for 4-bit ENOB, more than two orders of magnitude
below what the same multiply-accumulate would cost on a dedicated digital accelerator added to the device~\cite{reuther2025lincoln,anderson2023optical}.
This trend can be scaled up to $\chanNum={1{,}024}$ without significant loss of computing accuracy: an accumulation of thousands of products per output element in one analog
pass, the regime of the widest layers in contemporary CNN backbones, and a scale no single operation can reach when the mixer is cast as a generic matrix-vector multiplier.
 
Accuracy follows the PSD dependence established above. For {256}-channel layers, ENOB rises monotonically with operating energy and saturates near 5 bits (Figs.~\ref{fig:figure-scalability}h,~\ref{fig:figure-scalability}j), with channel calibration and peak-to-average power ratio (PAPR) mitigation ({\supplementary}s~\ref{ssec: supplementary-channel-calibration} and~\ref{ssec: supplememtary-waveform-papr}) applied throughout to operate the hardware at its best achievable performance.
At an operating PSD of {$-$112}\thinspace{dBm/Hz}, the outputs match full-precision ground truth in both real and imaginary components (Figs.~\ref{fig:figure-scalability}i,~\ref{fig:figure-scalability}k): {5.1} bits at {1.64}\thinspace{fJ/MAC} for {\convonedim} and {5.0} bits at {0.72}\thinspace{fJ/MAC} for {\convtwodim}.
This 5-bit precision meets the demands of modern quantized
inference~\cite{bondarenko2021understanding}.
More measurements can be found in {\supplementary}~\ref{ssec: supplementary-experiment-scalability}.

\section*{Deep CNN inference on time-series signals and vision tasks}
 
\begin{figure*}[!t]
    \centering
    \includegraphics[width=1.0\columnwidth]{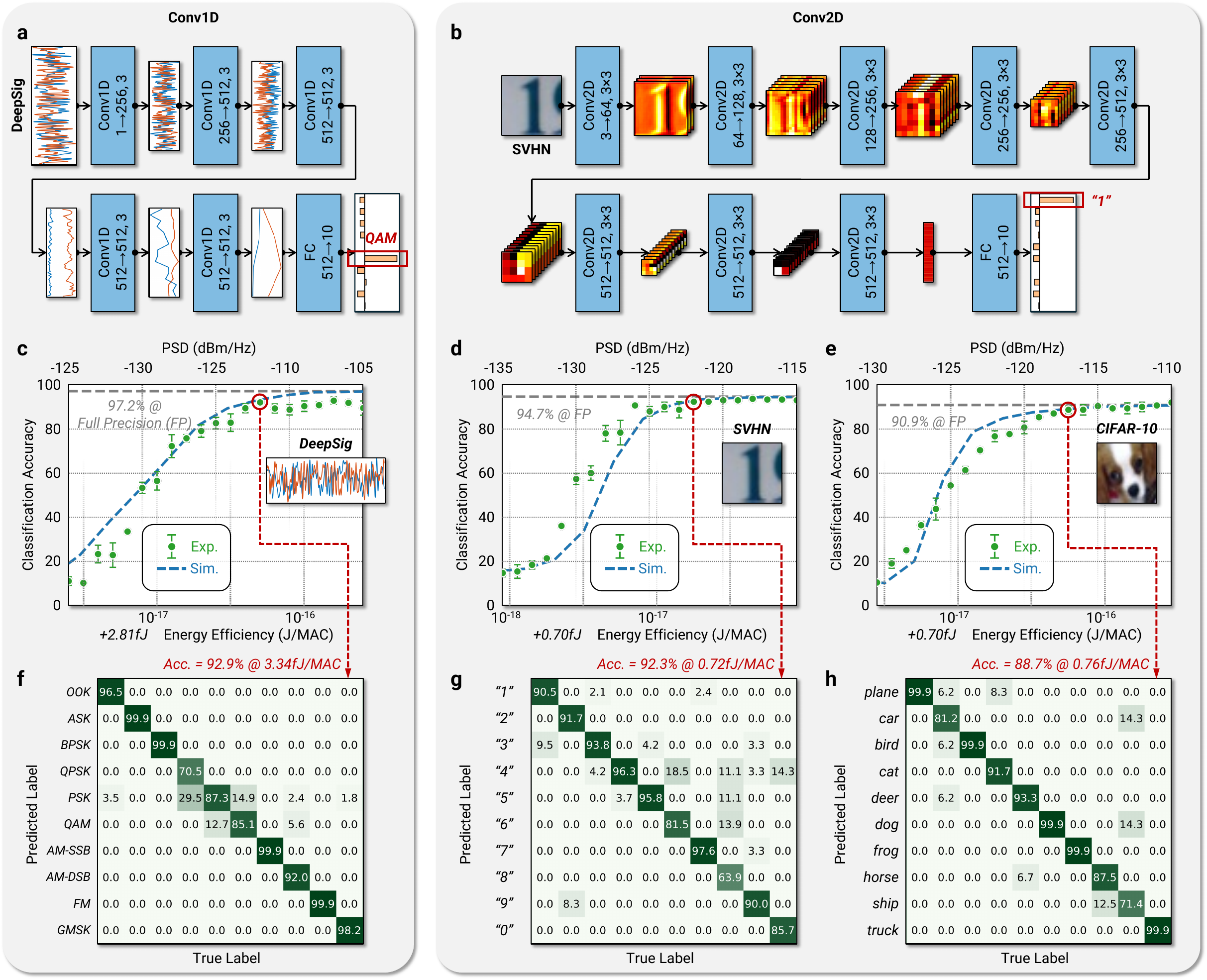}
    \caption{
    \textbf{{\namebf} inference for wireless communication and computer vision tasks.}
    \textbf{a}, Model architecture for wireless signal modulation classification on the DeepSig dataset, consisting of five {\convonedim} layers and one FC layer with {2.18} million parameters and {368.1} million MACs.
    \textbf{b}, Model architecture for image classification on the SVHN and CIFAR-10 datasets, consisting of eight {\convtwodim} layers and one FC layer with {9.22} million parameters and {611.1} million MACs.
    \textbf{c--e}, Classification accuracy as a function of PSD or energy efficiency for DeepSig (\textbf{c}), SVHN (\textbf{d}), and CIFAR-10 (\textbf{e}), including the experimental measurements and simulation results.
    \textbf{f--h}, Confusion matrices of {\name} inference results for DeepSig (\textbf{f}), SVHN (\textbf{g}), and CIFAR-10 (\textbf{h}), respectively.
    }
    \label{fig:figure-classifier}
\end{figure*}

The single-layer characterization establishes that one convolutional layer computes accurately; what practical inference requires is that many such layers chain, the analog output of each re-encoded as the input of the next layer, without errors accumulating into failure.
We therefore run \textit{complete, deep} CNNs of the scale used in modern edge applications: a network of five {\convonedim} layers and one fully-connected (FC) layer ({2.18} million parameters, {368.1} million MACs) for modulation classification of complex-valued wireless signals on DeepSig~\cite{o2018over}, and a network of eight {\convtwodim} layers and one FC layer ({9.22} million parameters, {611.1} million MACs) for image classification on SVHN and CIFAR-10
(Figs.~\ref{fig:figure-classifier}a--b, and more details in {\supplementary}~\ref{ssec: supplementary-experiment-model}).
The two architectures span the input domains a wireless edge device encounters: the one-dimensional,
complex-valued signals native to the radio itself, for which {\name}
performs inference on exactly the kind of waveform its hardware was built
to receive, and the two-dimensional natural images of standard computer
vision.
 
Across all three tasks, {\name} tracks its full-precision digital
counterpart closely. On DeepSig, {\name} distinguishes ten modulation
formats at {92.9\%} accuracy against a full-precision baseline of
{97.2\%}, operating at {3.34}\thinspace{fJ/MAC}
(Fig.~\ref{fig:figure-classifier}c). On SVHN it reaches {92.3\%} against
{94.7\%} at {0.72}\thinspace{fJ/MAC}, and on CIFAR-10, {88.7\%} against
{90.9\%} at {0.76}\thinspace{fJ/MAC}
(Figs.~\ref{fig:figure-classifier}d--e). More measurements can be found in {\supplementary}~\ref{ssec: supplementary-experiment-energy-classification}.
The small gap to full precision is
sustained through as many as eight successive analog convolutional layers:
the finite precision of each layer does not compound into degradation with
depth. Notably, the sub/few-fJ/MAC operating points at which these
accuracies are obtained are not a separate low-power mode but the same
regime characterized in Fig.~\ref{fig:figure-scalability}; accurate
deep inference and the two-orders-of-magnitude energy advantage are
achieved simultaneously, not traded against each other.
 
The confusion matrices show that {\name}'s residual errors are not an
artifact of analog computation but the ordinary errors of the underlying
model (Figs.~\ref{fig:figure-classifier}f--h). On DeepSig,
misclassification concentrates almost entirely among QPSK, PSK, and QAM, modulation formats with closely related constellations, while structurally distinct formats such as FM, GMSK, and AM-SSB are classified nearly perfectly.
The analog hardware, in other words, inherits the semantic error structure of the digital network rather than superimposing failure modes of its own.

Beyond validating the simulations, the accuracy-energy curves
(Figs.~\ref{fig:figure-classifier}c--e) expose an additional degree of
freedom: the trade between accuracy and energy is set by a single
parameter---the transmit PSD of the input---and can therefore be tuned at
runtime without retraining or reconfiguration.

\section*{Generative inference with latent-code control}

\begin{figure*}[!t]
    \centering
    \includegraphics[width=1.0\columnwidth]{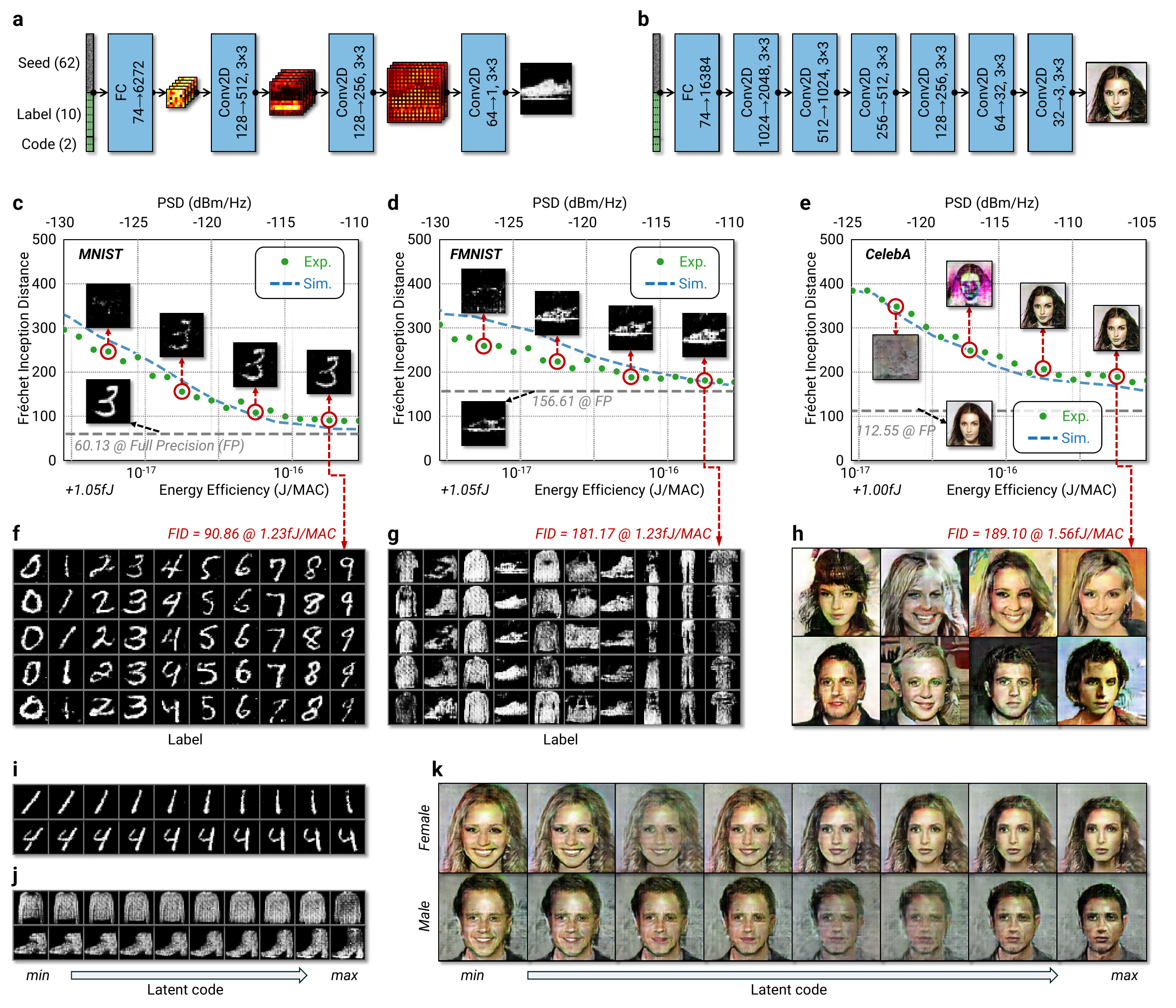}
    \caption{
    \textbf{{\namebf} inference for generative tasks.}
    \textbf{a}, The model architecture of the InfoGAN-based generative AI for either digit (MNIST) or fashion products (FMNIST), including one FC layer and three {\convtwodim} layers with {1.35} million parameters and {350.5} million MACs.
    \textbf{b}, The model architecture of the InfoGAN-based generative AI for high-resolution RGB human faces (CelebA), including one FC layer and six {\convtwodim} layers with {26.4} million parameters and {5,195} million MACs.
    \textbf{c--e}, Fréchet inception distance (FID) as a function of energy per MAC for MNIST (\textbf{c}), FMNIST (\textbf{d}), and CelebA (\textbf{e}), including both simulations and experimental results.
    \textbf{f--h}, Representative generated image examples from the {\name} for MNIST (\textbf{f}), FMNIST (\textbf{g}), and CelebA (\textbf{h}) at the selected high quality points.
    \textbf{i/j}, Generated MNIST and FMNIST samples with fixed class labels and varying latent codes, demonstrating smooth latent-space interpolation and semantic continuity.
    \textbf{k}, Conditional human face generation with varying latent codes for female and male conditions, showing continuous and controllable changes in facial appearance.
    }
    \label{fig:figure-gan}
\end{figure*}

Generative models pose a more stringent test of analog computing precision than classification, where quantization error is largely absorbed in the final class decision~\cite{bondarenko2021understanding,ma2025quantum}: the network's output is the image itself, produced pixel by pixel by the analog hardware, and any structured computational error is directly visible in the result. Beyond fidelity, generation on the edge demands controllability: a device that synthesizes content is useful only if the content can be specified. We therefore implement generative models based on InfoGAN~\cite{chen2016infogan}, whose generator input separates into three parts with distinct roles (Figs.~\ref{fig:figure-gan}a--b): a {62}-dimensional random seed providing sample diversity, a {10}-dimensional class label selecting \emph{what} is generated, and a {2}-dimensional latent code steering \emph{how}: a semantically interpretable attribute of the output. Whether this three-way separation survives execution on RF hardware is the question this section answers.

We evaluate two model scales on three datasets: a four-layer network (one FC and three {\convtwodim} layers; {1.35} million parameters, {350.5} million MACs) generating grayscale handwritten digits (MNIST) and fashion products (FMNIST), and a seven-layer network (one FC and six {\convtwodim} layers; {26.4} million parameters, {5.195} billion MACs) generating RGB human faces (CelebA), executing over five billion analog MACs for every face produced. More model details can be found in {\supplementary}~\ref{ssec: supplementary-experiment-model}.

The generated samples are well-formed across all three tasks (Figs.~\ref{fig:figure-gan}f--h). Measured over {100} generated images, {\name} reaches a Fr\'echet inception distance (FID) of {90.86} on MNIST at {1.23}\thinspace{fJ/MAC}, {181.17} on FMNIST at the same energy, and {189.10} on CelebA at {1.56}\thinspace{fJ/MAC}, approaching the full-precision baselines, with measured FID following the simulated energy-quality curve throughout the operating range (Figs.~\ref{fig:figure-gan}c--e).
More results can be found in {\supplementary}~\ref{ssec: supplementary-experiment-energy-generation}.
Along that curve, image quality degrades gracefully: at reduced energy the samples acquire noise but remain recognizable as digits, garments, and faces (insets of Figs.~\ref{fig:figure-gan}c--e), extending the runtime energy-quality trade established for classification to generative workloads.

The three-way separation of the generator input survives intact. The class label retains exclusive command of category: conditioning reliably produces the requested digit ``0'' through ``9'', the requested garment, and the requested gender of face. The latent code retains its independent, continuous handle on a single attribute: sweeping it from minimum to maximum while the label is held fixed tilts the slant of a digit, raises the height of a garment, and shifts the framing of a face, monotonically and without disturbing the class identity (Figs.~\ref{fig:figure-gan}i--k).
Five billion analog MACs at approximately 5-bit precision thus preserve not only the appearance of individual outputs but the disentangled structure the generator learned in digital training, the property that practical edge generation needs, from personalized content to adaptive human-machine interfaces and semantic scene synthesis.

\section*{Discussion and conclusion}

We have shown that the frequency mixer, the element at the core of almost every wireless edge device, is, by its physics, a convolution engine, and that this operation alone suffices to run complete CNNs end to end in hardware on the communication front end already present at the edge. On a wireless testbed, {\name} performs networks of up to nine layers and up to five billion analog MACs per inference---spanning modulation classification of complex-valued wireless signals, image classification, and image generation with latent-code control---at accuracy close to full-precision digital baselines and at sub-femtojoule-per-MAC energies, using no computing hardware beyond the reused front end.

The present implementation is a proof of concept, and its throughput, approximately {0.9}\thinspace{GOPS}, is far below that of contemporary digital processors. This reflects the testbed rather than the paradigm: the demonstration uses a single RF chain and a single frequency-mixing pathway over a limited bandwidth. Throughput scales directly with the wireless bandwidth and with the number of parallel mixing pathways, both of which are modest here and are routinely large in deployed radios~\cite{govind2024ultra}. Massive MIMO front ends, integrated mixer arrays, and wide bandwidth available at higher frequencies (millimeter-wave or sub-THz bands) offer a path to raising throughput by orders of magnitude without altering the underlying mapping.

Several boundaries of the approach should be stated plainly. The reported energies are edge-side: they exclude the power the central radio spends broadcasting weights, which is justified when that transmission is amortized across many receiving devices and drawn from grid-connected infrastructure, but which makes the advantage one of disaggregation rather than of total system energy.
The computing accuracy saturates near five ENOB, sufficient for the tasks demonstrated but not for those demanding higher precision. And the scheme presumes access to a radio that supplies weights over the air, so it suits devices operating within communication range rather than in isolation, and it addresses inference rather than training.

These constraints delimit rather than diminish the opportunity. Delivering weights over the air raises questions of confidentiality that the same wireless-physics toolbox is well placed to answer, including physical-layer and quantum-secured schemes~\cite{yin2020entanglement}. 
Integrating the mixer array on-chip would compound the energy and throughput gains while preserving compatibility with the communication stack. And because the mapping rests on a physical operation rather than a bespoke device, it invites extension beyond convolution toward the other structured linear operations that dominate modern models.

More broadly, {\name} reframes the relationship between communication and computation in which neither is built for the other. 
Correspondences between what hardware already does and what a computation requires are found rather than engineered, and the hardware need not be changed, only used for a second purpose alongside.
Much of the hardware already deployed may hold computational capability of its own.
Realized across the wireless infrastructure already deployed, this points toward an edge where communication and computation are not competing demands on separate hardware but two functions of a single physical substrate.

\clearpage

\clearpage

%% file: tex/methods.tex
\section*{Methods}
\label{sec:methods}

\subsection*{Tone Mapping Algorithm for Convolutional Layers}

The frequency encoding algorithm of {\name} leverages the convolutional layers (see definition in {\supplementary}~\ref{ssec: supplementary-theory-convolution-basics}) by cross-correlation between two series of frequency tones. This cross-correlation in the frequency domain is equivalent to the time-domain multiplication, which is inherently performed in the physics of the computing mixer.
We analyze the frequency encoding algorithm tailored for complex-valued 2-dimensional convolutional layers, abbreviated as {\convtwodim} layers, which can also be generalized to other dimensions. 

\myparatight{Notations of input, weight, and output in {\convtwodimbf} layers}
For the specific {\convtwodim} layer under consideration, let the kernel size be $\kernelSizeX \times \kernelSizeY$ (both $\kernelSizeX$ and $\kernelSizeY$ are odd numbers); by padding zeros around the input image of $\paddingSizeX=(\kernelSizeX-1)/2$ and $\paddingSizeY=(\kernelSizeY-1)/2$, we have the same input and output image dimensions, both of which are $\imageSizeX \times \imageSizeY$.
Additionally, let $\chanInNum$ represent the number of input channels and $\chanOutNum$ the number of output channels. 
Accordingly, the input, weight, and output tensors of this {\convtwodim} layer are represented as $\inputTensor \in \mathbb{C}^{\chanInNum \times \imageSizeX \times \imageSizeY}$ with $\inputScalar{\chanInIdx, \imageIdxX, \imageIdxY}$ as the $(\chanInIdx, \imageIdxX, \imageIdxY)$-th element, $\weightTensor \in \mathbb{C}^{\chanOutNum \times \chanInNum \times \kernelSizeX \times \kernelSizeY}$ with $\weightScalar{\chanInIdx, \chanOutIdx, \kernelIdxX, \kernelIdxY}$ as the $(\chanInIdx, \chanOutIdx, \kernelIdxX, \kernelIdxY)$-th element, and $\outputTensor \in \mathbb{C}^{\chanOutNum \times \imageSizeX \times \imageSizeY}$ with $\outputScalar{\chanOutIdx, \imageIdxX, \imageIdxY}$ as the $(\chanOutIdx, \imageIdxX, \imageIdxY)$-th element, respectively.
We further assume that the kernel size is no greater than the image size as $\kernelSizeX \leq \imageSizeX$ and $\kernelSizeY \leq \imageSizeY$.

\myparatight{Notations of frequency tone series}
On the other hand, the series of frequency tones representing the input $\inputTensor$ and weight $\weightTensor$ in the cross-correlation can be formulated by two complex-valued vectors, denoted as $\specInputVec \in \mathbb{C}^{\specInputLen}$ and $\specWeightVec \in \mathbb{C}^{\specWeightLen}$, whose lengths are $\specInputLen$ and $\specWeightLen$.
After the cross-correlation, we only examine a contiguous subset of frequency tones for the output $\outputTensor$, denoted as a complex-valued vector $\specOutputVec \in \mathbb{C}^{\specOutputLen}$ with a length of $\specOutputLen$, which is a sub-vector of the full cross-correlation output with a length of $\specInputLen + \specWeightLen - 1$ (see {\supplementary}~\ref{ssec: supplementary-theory-linear-convolution} for more details).

\myparatight{Single input-and-output channel case}
We start with the simple case of a single input and output channel, i.e., $\chanInNum = \chanOutNum = 1$, as shown in Fig.~\ref{fig:conv-algorithm}(a).
In this case, the input, weight, and output can be simplified as matrices: $\inputTensor \in \mathbb{C}^{\imageSizeX \times \imageSizeY}$, $\weightTensor \in \mathbb{C}^{\kernelSizeX \times \kernelSizeY}$, and $\outputTensor \in \mathbb{C}^{\imageSizeX \times \imageSizeY}$. Hence, their element indexing can be simplified as $\inputElem{\imageIdxX, \imageIdxY}$, $\weightElem{\kernelIdxX, \kernelIdxY}$, and $\outputElem{\imageIdxX, \imageIdxY}$, respectively.
For the input $\inputTensor$, we map it into the frequency tone series $\specInputVec$ with $\specInputLen=(\imageSizeX+\paddingSizeX) \cdot (\imageSizeY+\paddingSizeY)$ tones.
Specifically, the first row of $\inputTensor$, indexed by $\inputElem{0, 0}, \inputElem{1, 0}, \dots, \inputElem{\imageSizeX-1, 0}$, are mapped to the first $\imageSizeX$ tones indexed by $\specInputScalar[0, 1, \dots, \imageSizeX-1]$; then, a total of $\paddingSizeX$ zeros are padded next at $\specInputScalar[\imageSizeX, \dots, \imageSizeX+\paddingSizeX-1]$ on $\specInputVec$.
These $\paddingSizeX$ zeros correspond to the padded zero-pixels to the original image, which are reused for both the left padding for the first row and the right padding for the second row.
This process repeats for $\imageSizeY$ times for all the $\imageSizeY$ rows in $\inputTensor$.
The last $\paddingSizeY \cdot (\imageSizeX+\paddingSizeX)$ tones are all zeros, representing the additional zero-row padded to the original image.
To sum up, this mapping process can be formulated as:
\begin{align}
    \specInputScalar[i] &=
    \begin{cases}
        \inputElem{\imageIdxX, \imageIdxY}, & i = (\imageSizeX + \paddingSizeX) \cdot \imageIdxY + \imageIdxX, \\
        0, & \text{otherwise},
    \end{cases} \\
    \forall~\imageIdxX &= 0, 1, \dots, \imageSizeX - 1, ~\text{and}~ \imageIdxY = 0, 1, \dots, \imageSizeY - 1.
\end{align}
Similarly, there are $\specWeightLen=(\imageSizeX+\paddingSizeX) \cdot (\imageSizeY+\paddingSizeY)$ tones in $\specWeightVec$ for the weight $\weightTensor$.
The first row of $\weightTensor$, indexed by $\weightElem{0, 0}, \weightElem{1, 0}, \dots, \weightElem{\kernelSizeX-1, 0}$, are mapped to the first $\kernelSizeX$ tones indexed by $\specWeightScalar[0, 1, \dots, \kernelSizeX-1]$; to align the second row in $\weightTensor$ with the second row in $\inputTensor$, we pad $(\paddingSizeX+\imageSizeX-\kernelSizeX)$ zeros tones in $\specWeightVec$ indexed as $\specWeightScalar[\kernelSizeX, \dots, \imageSizeX+\paddingSizeX-1]$ on $\specWeightVec$.
This process repeats for $\kernelSizeY$ times for all the $\kernelSizeY$ rows in $\weightTensor$, while leaving all the remaining tones in $\specWeightVec$ as zeros.
This mapping process from $\weightTensor$ to $\specWeightVec$ can be formulated as:
\begin{align}
    \specWeightScalar[i] &=
    \begin{cases}
        \weightElem{\kernelIdxX, \kernelIdxY}, & i = (\imageSizeX + \paddingSizeX) \cdot \kernelIdxY + \kernelIdxX, \\
        0, & \text{otherwise},
    \end{cases} \\
    \forall~\kernelIdxX &= 0, 1, \dots, \kernelSizeX - 1, ~\text{and}~ \kernelIdxY = 0, 1, \dots, \kernelSizeY - 1.
\end{align}
Originally, the cross-correlation between the $\specInputLen$-point $\specInputVec$ and the $\specWeightLen$-point $\specWeightVec$ yields a $(\specInputLen + \specWeightLen - 1)$-point output vector.
We only examine $\specOutputLen$ contiguous elements on the cross-correlation output, indexed from $\left\{\specOutputStart, \specOutputStart+1, \dots, \specOutputEnd-1\right\}$, which is denoted as $\specOutputVec \in \mathbb{C}^{\specOutputLen}$.
Specifically, we set the starting and ending indices as
\begin{align}
    \begin{cases}
        \specOutputStart &= \specWeightLen -1 = (\imageSizeX + \paddingSizeX) \cdot (\imageSizeY + \paddingSizeY) - 1, \\
        \specOutputEnd &= \specInputLen + \specWeightLen -1 = 2 \cdot (\imageSizeX + \paddingSizeX) \cdot (\imageSizeY + \paddingSizeY) - 1,
    \end{cases}
    \Rightarrow \quad \specOutputLen &= \specOutputEnd - \specOutputStart = (\imageSizeX + \paddingSizeX) \cdot (\imageSizeY + \paddingSizeY).
\end{align}
Under this setting, we notice that the mapping between the $\outputTensor$ and $\specOutputVec$ is the reverse process of that between $\inputTensor$ and $\specInputVec$, as illustrated in Fig.~\ref{fig:conv-algorithm}(a).
The first row of $\outputTensor$, indexed by $\outputElem{0, 0}, \outputElem{1, 0}, \dots, \outputElem{\imageSizeX-1, 0}$, corresponds to the first $\imageSizeX$ tones indexed by $\specOutputScalar[0, 1, \dots, \imageSizeX-1]$. The subsequent $\paddingSizeX$ tones are ignored and discarded. This procedure is repeated for all $\imageSizeY$ rows in $\outputTensor$.
With the above mapping rules, we leverage the cross-correlation from $\specInputVec$ and $\specWeightVec$ to $\specOutputVec$, which realizes the single-input/output-channel {\convtwodim} from $\inputTensor$ and $\weightTensor$ to $\outputTensor$.

\myparatight{Multiple input and single output channel case}
When there are $\chanInNum$ input channels in the {\convtwodim}, we need to generate the frequency tone series per input channel for both $\inputTensor$ and $\weightTensor$.
As shown in Fig.~\ref{fig:conv-algorithm}(b), we concatenate these frequency tone series from all the $\chanInNum$ input channels into large frequency tone series as $\inputTensor$ and $\weightTensor$, which can be written as
\begin{align}
    \specInputVec &= \left[ \specInputVec \vert_{\chanInIdx=0}, \specInputVec\vert_{\chanInIdx=1}, \dots, \specInputVec\vert_{\chanInIdx=\chanInNum-1} \right], \\
    \specWeightVec &= \left[ \specWeightVec \vert_{\chanInIdx=0}, \specWeightVec\vert_{\chanInIdx=1}, \dots, \specWeightVec\vert_{\chanInIdx=\chanInNum-1} \right],
\end{align}
whose length becomes $\specInputLen=\specWeightLen=\chanInNum \cdot (\imageSizeX + \paddingSizeX) \cdot (\imageSizeY + \paddingSizeY)$.
When performing the cross-correlation between $\specInputVec$ and $\specWeightVec$, we only examine the contiguous subset of $\specOutputLen$ tones indexed from $\left\{\specOutputStart, \specOutputStart+1, \dots, \specOutputEnd-1\right\}$ as the output $\specOutputVec \in \mathbb{C}^{\specOutputLen}$, where
\begin{align}
    &\begin{cases}
        \specOutputStart &= \specWeightLen -1 = \chanInNum \cdot (\imageSizeX + \paddingSizeX) \cdot (\imageSizeY + \paddingSizeY) - 1, \\
        \specOutputEnd &= \specInputLen + (\imageSizeX+\paddingSizeX)\cdot(\imageSizeY+\paddingSizeY) -1 = (\chanInNum+1) \cdot (\imageSizeX + \paddingSizeX) \cdot (\imageSizeY + \paddingSizeY) - 1,
    \end{cases} \\
    & \Rightarrow \quad \specOutputLen = \specOutputEnd - \specOutputStart = (\imageSizeX + \paddingSizeX) \cdot (\imageSizeY + \paddingSizeY).
\end{align}
The mapping from $\specOutputVec$ to $\outputTensor$ remains the same as that in the single input/output channel case.
In practice, we add Zadoff-Chu phases on each input channel on $\specInputVec$ and inverse phases on $\specWeightVec$ to alleviate the peak-to-average power ratio (PAPR) of the waveforms while maintaining the calculation results (see {\supplementary}s~\ref{ssec: supplememtary-waveform-papr} and \ref{ssec: supplementary-experiment-waveform-amplitude}).

\myparatight{Multiple input-and-output channel case}
When there are multiple output channels ($\chanOutNum > 1$), we need to perform the above cross-correlation for $\chanOutNum$ times in serial, each of which corresponds to one output channel.
Specifically, the input frequency tone series $\specInputVec$ remains the same, while the weight frequency tone series $\specWeightVec$ corresponds to the $\chanOutIdx$-th output channel for the $\chanOutIdx$-th cross-correlation; the output of the $\chanOutIdx$-th cross-correlation corresponds to the $\chanOutIdx$-th output channel in $\outputTensor$.

\subsection*{Bandwidth and Computation Throughput Analysis}

Assuming an accessible bandwidth of $\band$ of the wireless channel, we first analyze the wired/wireless bandwidths of the $\specInputVec$, $\specWeightVec$, and $\specOutputVec$, and their waveform time $\waveLen$ under these bandwidths.
Then, we analyze the computation throughput, i.e., the number of MACs per time, denoted as $\throughput$.

To simplify, we further assume $\imageSizeX \gg \paddingSizeX$ and $\imageSizeY \gg \paddingSizeY$, and have the approximation that $\specInputLen = \specWeightLen \approx \chanInNum \cdot \imageSizeX \imageSizeY$, and that $\specOutputLen \approx \imageSizeX \imageSizeY$.

\myparatight{Bandwidth analysis}
In {\name}, the weight $\specWeightVec$ is transmitted over the wireless channel, meaning $\bandWeight = \band$.
Then, the tone spacing for $\specWeightVec$ can be given by
\begin{align}
    \freqSub = \frac{\band}{\specWeightLen} \approx \frac{\band}{\chanInNum \cdot \imageSizeX \imageSizeY}.
\end{align}
Note that this tone spacing $\freqSub$ is the same for $\specInputVec$ and $\specOutputVec$ (see {\supplementary}~\ref{ssec: supplementary-theory-linear-convolution}), so their bandwidths, denoted as $\bandInput$ and $\bandOutput$, can be given by
\begin{align}
    \bandInput = \freqSub \cdot \specInputLen = \band,
    \label{eq: method-computation-throughput-band-input}
\end{align}
and
\begin{align}
    \bandOutput = \freqSub \cdot \specOutputLen = \frac{\band}{\chanInNum}.
    \label{eq: method-computation-throughput-band-output}
\end{align}

\myparatight{Computation throughput analysis}
The waveform time of a single output channel can be given by $1/\freqSub$, which is to be repeated for $\chanOutNum$ times. Therefore, the total waveform time $\waveLen$ can be given by
\begin{align}
    \waveLen = \chanOutNum \cdot \frac{1}{\freqSub} \approx \frac{\chanInNum \cdot \chanOutNum \cdot \imageSizeX \imageSizeY}{\band}.
    \label{eq: method-computation-throughput-waveform-time}
\end{align}
During $\waveLen$, a total of $4 \chanInNum \cdot \chanOutNum \cdot \kernelSizeX \kernelSizeY \cdot \imageSizeX \imageSizeY$ real-valued MACs are computed. This corresponds to the computation throughput $\throughput$ of
\begin{align}
    \throughput = \frac{4 \chanInNum \cdot \chanOutNum \cdot \kernelSizeX \kernelSizeY \cdot \imageSizeX \imageSizeY}{\waveLen} \approx 4 \kernelSizeX \kernelSizeY \cdot \band.
\end{align}
This suggests that a faster computation throughput can be achieved by a larger wireless bandwidth $\band$. 
Also, the computation throughput of {\name} is scalable as the kernel size $\kernelSizeX \times \kernelSizeY$ increases.

\subsection*{Energy Efficiency Analysis}

Then, we analyze the energy efficiency of {\name} on the edge client.
In the following, we discuss the energy consumption of each module in detail.

Throughout the analysis, we denote the energy consumption per real-valued MAC of digital computing as $\energyPerMAC$ (i.e., $4\energyPerMAC$ for a complex-valued MAC), and the energy consumption per real-valued sample in ADC as $\energyPerADC$ (i.e., $2\energyPerADC$ for an I/Q sample).
Throughout the analysis, we set $\energyPerADC = {100}\thinspace{\textrm{fJ/sample}}$~\cite{adc_survey} and $\energyPerMAC = {100}\thinspace{\textrm{fJ/MAC}}$~\cite{choquette2023nvidia}.

\myparatight{The encoding module}
The encoding module performs the mapping from the input tensor $\inputTensor$ to the frequency tone series $\specInputVec$, and converts the frequency-domain $\specInputVec$ into the time-domain, where a $\specInputLen$-point IFFT is conducted in digital computing.
Therefore, the energy consumption of the encoding module per layer, denoted as $\energyMVMEnc$, is
\begin{align}
    \energyMVMEnc = 2 \specInputLen \log_2 \specInputLen \cdot \energyPerMAC \approx 2 \chanInNum \cdot \imageSizeX\imageSizeY \log_2(\chanInNum \cdot \imageSizeX\imageSizeY) \cdot \energyPerMAC.
    \label{eq: method-energy-efficiency-encoding-mvm}
\end{align}
Given the number of real-valued MACs per layer as $4 \chanInNum \cdot \chanOutNum \cdot \kernelSizeX \kernelSizeY \cdot \imageSizeX \imageSizeY$, the energy consumption per MAC in the encoding module, denoted as $\energyMACEnc$, is
\begin{align}
    \energyMACEnc &= \frac{\energyMVMEnc}{4 \cdot \chanInNum \cdot \chanOutNum \cdot \kernelSizeX \kernelSizeY \cdot \imageSizeX \imageSizeY} = \frac{1}{2} \cdot \frac{\log_2(\chanInNum \cdot \imageSizeX\imageSizeY)}{\chanOutNum \cdot \kernelSizeX\kernelSizeY} \cdot \energyPerMAC.
    \label{eq: method-energy-efficiency-encoding-mac}
\end{align}

\myparatight{The digital-to-analog converter (DAC) module}
On the edge client, the DAC module generates the analog waveform carrying the input $\inputTensor$, whose energy consumption, denoted as $\energyMVMDAC$, can be expressed as the product of the transmitted power (denoted as $\powerTx$) and the waveform time $\waveLen$ derived in equation~\eqref{eq: method-computation-throughput-waveform-time}.
Thereby, we have the energy consumption of the DAC module per MVM as
\begin{align}
    \energyMVMDAC = \powerTx \cdot \waveLen \approx \powerTx \cdot \frac{\chanInNum \cdot \chanOutNum \cdot \imageSizeX \imageSizeY}{\band} = \chanInNum \cdot \chanOutNum \cdot \imageSizeX \imageSizeY \cdot \PSDTx,
    \label{eq: method-energy-efficiency-dac-mvm}
\end{align}
where we define $\PSDTx = \powerTx / \band$ as the power spectral density (PSD) of the DACs.
By normalizing it with the number of real-valued MACs per layer, the energy consumption per MAC in the DAC module, denoted as $\energyMACDAC$, is
\begin{align}
    \energyMACDAC &= \frac{\energyMVMDAC}{4 \cdot \chanInNum \cdot \chanOutNum \cdot \kernelSizeX \kernelSizeY \cdot \imageSizeX \imageSizeY} \approx \frac{1}{4} \cdot \frac{1}{\kernelSizeX \kernelSizeY} \cdot \PSDTx.
    \label{eq: method-energy-efficiency-dac-mac}
\end{align}

\myparatight{The analog-to-digital converter (ADC) module}
The ADC module receives the output frequency tone series $\specOutputVec$ from the computing mixer.
Given the smaller bandwidth $\bandOutput$ of $\specOutputVec$ in equation~\eqref{eq: method-computation-throughput-band-output}, the ADC can also operate at a smaller sampling rate of $\bandOutput$ to fully capture $\specOutputVec$.
Throughout the total waveform time $\waveLen$, the total energy consumed by ADC, denoted as $\energyMVMADC$ is given by
\begin{align}
    \energyMVMADC = \waveLen \cdot \bandOutput \cdot 2 \energyPerADC \approx \frac{\chanInNum \cdot \chanOutNum \cdot \imageSizeX \imageSizeY}{\band} \cdot \frac{\band}{\chanInNum} \cdot 2 \energyPerADC = 2 \chanOutNum \cdot \imageSizeX \imageSizeY \cdot \energyPerADC.
    \label{eq: method-energy-efficiency-adc-mvm}
\end{align}
Normalized by the number of MACs computed, we have the energy per MAC term for the ADC module as
\begin{align}
    \energyMACADC = \frac{\energyMVMADC}{4 \cdot \chanInNum \cdot \chanOutNum \cdot \kernelSizeX \kernelSizeY \cdot \imageSizeX \imageSizeY} \approx \frac{1}{2} \cdot \frac{1}{\chanInNum \cdot \kernelSizeX \kernelSizeY} \cdot \energyPerADC.
    \label{eq: method-energy-efficiency-adc-mac}
\end{align}

\myparatight{The decoding module}
The decoding module performs FFT in digital to extract the computing output $\outputTensor$ from the frequency domain $\specOutputVec$.
Specifically, a $\specOutputLen$-point FFT is performed per output channel $\chanOutIdx$, involving the MAC number of $2 \specOutputLen \log_2 \specOutputLen$; the total $\chanOutNum$ repeats yield the total energy consumption of this decoding module as
\begin{align}
    \energyMVMDec = \chanOutNum \cdot 2 \specOutputLen \log_2 \specOutputLen \cdot \energyPerMAC \approx 2 \chanOutNum \cdot \imageSizeX \imageSizeY \log_2 (\imageSizeX \imageSizeY) \cdot \energyPerMAC.
    \label{eq: method-energy-efficiency-decoding-mvm}
\end{align}
Correspondingly, the energy per MAC is given by
\begin{align}
    \energyMACDec = \frac{\energyMVMDec}{4 \cdot \chanInNum \cdot \chanOutNum \cdot \kernelSizeX \kernelSizeY \cdot \imageSizeX \imageSizeY} \approx \frac{1}{2} \cdot \frac{\log_2 (\imageSizeX \imageSizeY)}{\chanInNum \cdot \kernelSizeX \kernelSizeY} \cdot \energyPerMAC.
    \label{eq: method-energy-efficiency-decoding-mac}
\end{align}

\myparatight{Total energy efficiency}
Putting equations~\eqref{eq: method-energy-efficiency-encoding-mvm}, \eqref{eq: method-energy-efficiency-dac-mvm}, \eqref{eq: method-energy-efficiency-adc-mvm}, and \eqref{eq: method-energy-efficiency-decoding-mvm} together, we have the total energy consumption per {\convtwodim} layer as
\begin{align}
    \energyMVM &\approx \underbrace{2 \chanInNum \cdot \imageSizeX\imageSizeY \log_2(\chanInNum \cdot \imageSizeX\imageSizeY) \cdot \energyPerMAC}_{\energyMVMEnc} + \underbrace{\chanInNum \cdot \chanOutNum \cdot \imageSizeX \imageSizeY \cdot \PSDTx}_{\energyMVMDAC} \\
    &\quad + \underbrace{2 \chanOutNum \cdot \imageSizeX \imageSizeY \cdot \energyPerADC}_{\energyMVMADC} + \underbrace{2 \chanOutNum \cdot \imageSizeX \imageSizeY \log_2 (\imageSizeX \imageSizeY) \cdot \energyPerMAC}_{\energyMVMDec}.
\end{align}
Similarly, summing up equations~\eqref{eq: method-energy-efficiency-encoding-mac}, \eqref{eq: method-energy-efficiency-dac-mac}, \eqref{eq: method-energy-efficiency-adc-mac} and \eqref{eq: method-energy-efficiency-decoding-mac} yields the energy consumption per MAC as
\begin{align}
    \energyMAC &\approx \underbrace{\frac{1}{2} \cdot \frac{\log_2(\chanInNum \cdot \imageSizeX\imageSizeY)}{\chanOutNum \cdot \kernelSizeX\kernelSizeY} \cdot \energyPerMAC}_{\energyMACEnc} + \underbrace{\frac{1}{4} \cdot \frac{1}{\kernelSizeX \kernelSizeY} \cdot \PSDTx}_{\energyMACDAC} + \underbrace{\frac{1}{2} \cdot \frac{1}{\chanInNum \cdot \kernelSizeX \kernelSizeY} \cdot \energyPerADC}_{\energyMACADC} + \underbrace{\frac{1}{2} \cdot \frac{\log_2 (\imageSizeX \imageSizeY)}{\chanInNum \cdot \kernelSizeX \kernelSizeY} \cdot \energyPerMAC}_{\energyMACDec} \\
    &= \frac{1}{2} \cdot \frac{1}{\kernelSizeX \kernelSizeY} \cdot \left( \frac{1}{2} \cdot \PSDTx + \frac{1}{\chanInNum} \cdot \energyPerADC + \left[ \frac{\log_2(\chanInNum \cdot \imageSizeX\imageSizeY)}{\chanOutNum} + \frac{\log_2(\imageSizeX \imageSizeY)}{\chanInNum}\right] \cdot \energyPerMAC \right).
    \label{eq: method-energy-efficiency-mac-total}
\end{align}
We notice that the energy efficiency of {\name} is also scalable for the kernel size $\kernelSizeX \times \kernelSizeY$.
In addition, the energy per MAC decreases as the computing scale, $\chanInNum$ and $\chanOutNum$ increase, and converges to the thermal dynamic limit (TDL), $\energyMACTDL$, as
\begin{align}
    \energyMACTDL = \lim_{\chanInNum, \chanOutNum \rightarrow \infty} \energyMAC = \frac{1}{4} \cdot \frac{1}{\kernelSizeX\kernelSizeY} \cdot \PSDTx.
\end{align}

%% file: tex/acks.tex
\section*{Acknowledgments}

\myparatight{Funding}
Z.G., Y.C., and T.C. acknowledge partial support from the NSF Athena AI Institute for Edge Computing under award \#2112562.
Z.G., D.E., and T.C. acknowledge partial support from ARO under grant W911NF-25-1-0241. 
D.E. acknowledges support from AFRL under Cooperative Agreement No. FA8750-25-2-0500.
S.-Y.M. acknowledges support from the DARPA INSPIRED program (HR001123S0052) and from the NSF Engineering Research Center for Quantum Networks (CQN) under award \#1941583.

\myparatight{Author Contributions}
Z.G., D.E., and T.C. developed the original concept and system architecture. Z.G., S.-Y.M., and T.C. designed the experimental setup and evaluation plan. Z.G. conducted the experiments and analyzed the data. 
Z.G., S.-Y.M., and T.C. prepared the manuscript with the inputs from Y.C. and D.E..

\myparatight{Correspondence}
Requests for information should be directed to Tingjun Chen (\href{mailto:tingjun.chen@duke.edu}{tingjun.chen@duke.edu}).

%% file: tex/Supplementary_theory.tex
\section*{\textbf{Supplementary Information: Theory}}

\subsection{Convolutional Layer Basics and Formulation}
\label{ssec: supplementary-theory-convolution-basics}

{\name} is designed to accelerate general convolutional layers of any dimensions.
Without loss of generality, we start with the basics of the (complex-valued) 2D convolutional layer (abbreviated as {\convtwodim}) that is widely used in modern DL models (e.g., object detection, image classification, semantic segmentation, and generative models).
Then, we present the extension to convolutional layers with higher dimensions.

\myparatight{2D convolutional layer modeling}
We define the input to a {\convtwodim} layer as a (complex-valued) three-dimensional tensor, $\inputTensor \in \mathbb{C}^{\chanInNum \times \imageSizeX \times \imageSizeY}$, where $\chanInNum$ is the number of channels (e.g., $\chanInNum=3$ for an RGB-colored image), and $\imageSizeX \times \imageSizeY$ is the dimension of the input's height and width.
For the $\chanInIdx$-th channel, the element in the $\imageIdxX$-th row and $\imageIdxY$-th column is denoted as $\inputScalar{\chanInIdx, \imageIdxX, \imageIdxY} \in \mathbb{C}$.
The trainable weights in this {\convtwodim} layer are defined as a four-dimensional tensor, $\weightTensor \in \mathbb{C}^{\chanInNum \times \chanOutNum \times \kernelSizeX \times \kernelSizeY}$, where $\chanOutNum$ is the number of output channels, and $\kernelSizeX \times \kernelSizeY$ is the dimension of the 2D kernel along the height and width dimensions, respectively.
For the $\chanInIdx$-th input channel to the $\chanOutIdx$-th output channel, the kernel element in the $\kernelIdxX$-th row and $\kernelIdxY$-th column is denoted as $\weightScalar{\chanInIdx, \chanOutIdx, \kernelIdxX, \kernelIdxY} \in \mathbb{C}$.
Without loss of generality, we assume that the kernel dimensions, $\kernelSizeX$ and $\kernelSizeY$, are both odd numbers, and we further assume a zero-padding size $\paddingSizeX=(\kernelSizeX-1)/2$ and $\paddingSizeY = (\kernelSizeY-1)/2$ along the two dimensions. 
Under this definition, the height and width dimensions of the output are identical to those of the input, while the number of channels changes from $\chanInNum$ to $\chanOutNum$.
Therefore, the output of the {\convtwodim} layer can be defined as $\outputTensor \in \mathbb{C}^{\chanOutNum \times \imageSizeX \times \imageSizeY}$, whose $\chanOutIdx$-th channel's element in the $\imageIdxX$-th row and $\imageIdxY$-th column is $\outputScalar{\chanOutIdx, \imageIdxX, \imageIdxY} \in \mathbb{C}$.
Using such notations, the {\convtwodim} operation, denoted as $\conv{\cdot}$, can be formulated as
\begin{align}
    & \outputTensor = \conv{\weightTensor, \inputTensor},\ \text{where} \nonumber \\
    & \outputScalar{\chanOutIdx, \imageIdxX, \imageIdxY} = \sum_{\chanInIdx=1}^{\chanInNum} \sum_{\kernelIdxX=1}^{\kernelSizeX} \sum_{\kernelIdxY=1}^{\kernelSizeY} \weightScalar{\chanInIdx, \chanOutIdx, \kernelIdxX, \kernelIdxY} \cdot \inputScalar{\chanInIdx, \imageIdxX-\paddingSizeX+\kernelIdxX, \imageIdxY-\paddingSizeY+\kernelIdxY}.
    \label{eq: conv2d-operation}
\end{align}
Typically, there are a total number of $(\chanInNum\chanOutNum \cdot \kernelSizeX\kernelSizeY)$ complex-valued trainable parameters in $\weightTensor$ in such a {\convtwodim} layer, corresponding to a total number of $(\chanInNum\chanOutNum \cdot \kernelSizeX\kernelSizeY \cdot \imageSizeX\imageSizeY)$ complex-valued MACs, or $(4 \cdot \chanInNum \chanOutNum \cdot \kernelSizeX\kernelSizeY \cdot \imageSizeX\imageSizeY)$ real-valued MACs.

\myparatight{Arbitrary-dimension convolutional layer}
We further extend the definition for a {\convtwodim} layer to a general $\convDimenNum$-dimensional convolutional layer, where the input is a $(\convDimenNum+1)$-dimensional tensor $\inputTensor \in \mathbb{C}^{\chanInNum \times \imageSize_1 \times \cdots \times \imageSize_{\convDimenNum}}$, the weight is a $(\convDimenNum+2)$-dimensional tensor $\weightTensor \in \mathbb{C}^{\chanInNum \times \chanOutNum \times \kernelSize_1 \times \cdots \times \kernelSize_{\convDimenNum}}$, and the output is a $(\convDimenNum+1)$-dimensional tensor $\outputTensor \in \mathbb{C}^{\chanOutNum \times \imageSize_1 \times \cdots \times \imageSize_{\convDimenNum}}$.
Following equation~\eqref{eq: conv2d-operation}, the $\convDimenNum$-dimensional convolution operation can be written as
\begin{align}
    & \outputTensor = \conv{\weightTensor, \inputTensor},\ \text{where} \nonumber \\
    & \outputScalar{\chanOutIdx, \imageIdx_1, \dots, \imageIdx_{\convDimenNum}}
    = \sum_{\chanInIdx=1}^{\chanInNum} \sum_{\kernelIdx_1=1}^{\kernelSize_1} \dots \sum_{\kernelIdx_{\convDimenNum}=1}^{\kernelSize_{\convDimenNum}} \weightScalar{\chanInIdx, \chanOutIdx, \kernelIdx_{1}, \dots, \kernelIdx_{\convDimenNum}} \cdot \inputScalar{\chanInIdx, \imageIdx_1-\paddingSize_1+\kernelIdx_1, \dots, \imageIdx_{\convDimenNum}-\paddingSize_{\convDimenNum}+\kernelIdx_{\convDimenNum}}.
    \label{eq: convXd-operation}
\end{align}

\myparatight{Fully-connected (FC) layer as a special case of convolutional layer}
A fully-connected (FC) layer with an input size of $\hiddenInNum$ and an output size of $\hiddenOutNum$ can be formulated as a matrix-vector multiplication that $\outputTensor = \weightTensor \cdot \inputTensor$, where $\inputTensor \in \mathbb{C}^{\hiddenInNum}$, $\weightTensor \in \mathbb{C}^{\hiddenInNum}$, and $\outputTensor \in \mathbb{C}^{\hiddenOutNum}$.
There are a total number of $\hiddenInNum \hiddenOutNum$ complex-valued trainable parameters in $\weightTensor$, corresponding to a total number of $\hiddenInNum \hiddenOutNum$ complex-valued MACs, or $4 \hiddenInNum \hiddenOutNum$ real-valued MACs.
This FC layer can be viewed as a special case of a {\convtwodim} layer with $\chanInNum=\hiddenInNum$ and $\chanOutNum=\hiddenOutNum$ with $\kernelSizeX=\kernelSizeY=\imageSizeX=\imageSizeY=1$.
Therefore, inferring an FC layer can either be accelerated by {\name} or the existing work~\cite{gao2026disaggregated}.

\subsection{Using Frequency Mixer for Convolution and Cross-Correlation}
\label{ssec: supplementary-theory-linear-convolution}

We conduct analog computing for complex-valued convolution using a (passive) frequency mixer, which is the core computing operation to drive the convolution layers in {\name}.
According to the convolution theorem~\cite{oppenheim1999discrete, gao2026disaggregated}, element-wise multiplication in the time domain is equivalent to the convolution in the frequency domain; when one of the operands is conjugated in the time domain, the convolution in the frequency domain becomes cross-correlation.

\myparatight{Convolution and cross-correlation formulation}
We consider the convolution between two complex-valued vectors with lengths $\convInputLen$ and $\convWeightLen$, respectively, denoted as $\convInputVec \in \mathbb{C}^{\convInputLen}$ and $\convWeightVec \in \mathbb{C}^{\convWeightLen}$.
The output of this convolution, denoted as $\convOutputVec \in \mathbb{C}^{\convOutputLen}$, where $\convOutputLen = \convInputLen + \convWeightLen - 1$, can be formulated as
\begin{align}
    \convOutputVec = \convInputVec \convolution \convWeightVec, ~\text{where}~
    \convOutputScalar[\convIdx] = \sum_{\convIdxTemp=\max\{0, \convIdx-\convWeightLen+1\}}^{\min\{\convInputLen-1, \convIdx\}} \convInputScalar[\convIdxTemp] \cdot \convWeightScalar[\convIdx-\convIdxTemp],\
    \forall \convIdx = 0, 1, \dots, \convInputLen+\convWeightLen-2.
    \label{eq: supplementary-linear-convolution-linear-convolution}
\end{align}
This convolution contains a total number of $\convInputLen\convWeightLen$ complex-valued MACs, or $4 \convInputLen\convWeightLen$ real-valued MACs.
Similarly, one can formulate the cross-correlation operation, whose only difference from the convolution is that $\convOutputVec$ is not flipped, given by
\begin{align}
    \convOutputVec^{\prime} = \convInputVec \correlation \convWeightVec, ~\text{where}~
    \convOutputScalar^{\prime}[\convIdx] = \sum_{\convIdxTemp=\max\{0, \convIdx-\convWeightLen+1\}}^{\min\{\convInputLen-1, \convIdx\}} \convInputScalar[\convIdxTemp] \cdot \convWeightScalar[\convIdxTemp+\convWeightLen-\convIdx-1],\
    \forall \convIdx = 0, 1, \dots, \convInputLen+\convWeightLen-2.
    \label{eq: supplementary-linear-convolution-cross-correlation}
\end{align}

\myparatight{Waveform generation using IFFT}
The two vectors, $\convInputVec$ and $\convWeightVec$, are mapped to two series of frequency tones in the frequency domain.
We assume the analog computing process is completed within a time duration of $\waveLen$, which corresponds to a tone spacing of $\freqSub = 1/\waveLen$ for each frequency tone series.
For $\convInputVec$, we map the $\convInputLen$ elements to $\convInputLen$ tones, where the $\convIdx$-th element, $\convInputScalar[\convIdx]$, is assigned on the $\convIdx$-th tone at the frequency of $\convIdx \cdot \freqSub$.
Its corresponding time domain I/Q sample sequence, or \emph{I/Q waveform}, is an $\convInputLen$-point vector, $\convSampInputVec \in \mathbb{C}^{\convInputLen}$, obtained by conducting an $\convInputLen$-point IFFT, i.e.,
\begin{align}
    \convSampInputVec = \ifft{\convInputVec},\ \text{where}~
    \convSampInputScalar[\convIdx] = \frac{1}{\convInputLen} \sum_{\convIdxTemp=0}^{\convInputLen-1} \convInputScalar[\convIdxTemp] \cdot \eu^{\iu 2 \pi \frac{\convIdxTemp \convIdx}{\convInputLen}}, ~\forall \convIdx = 0, 1, \dots, \convInputLen-1.
    \label{eq: supplementary-linear-convolution-encoding}
\end{align}
This $\convInputLen$-point IFFT operation contains a total number of $\frac{\convInputLen}{2} \cdot \log_2 \convInputLen$ complex-valued MACs, or $2 \convInputLen \cdot \log_2 \convInputLen$ real-valued MACs.
Then, this I/Q waveform $\convSampInputVec$, which carries the information of $\convInputVec$, is streamed to a DAC at the sampling rate of $\convInputLen \cdot \freqSub$, generating a continuous-time waveform $\convWaveInput(\waveIdx)$ lasting for a time duration of $\waveLen$.
This waveform occupies a bandwidth of $\convInputLen \cdot \freqSub$ in the frequency domain.
Similarly, this process can be applied to $\convWeightVec$.
First, an $\convWeightLen$-point IFFT with $2\convWeightLen \cdot \log_2 \convWeightLen$ MACs provides the time-domain I/Q waveform $\convSampWeightVec \in \mathbb{C}^{\convWeightLen}$.
Then, a DAC operating at a sampling rate of $\convWeightLen \cdot \freqSub$ generates the analog waveform of $\convWaveWeight(\waveIdx)$ with a time duration of $\waveLen$, occupying a bandwidth of $\convWeightLen \cdot \freqSub$.
This waveform generation process is illustrated in the left half of Fig.~\ref{fig:supplementary-convolution-diagram}(a).

\myparatight{Waveform analysis for the convolution output}
The two generated analog waveforms are fed into a frequency mixer, or, namely, a computing mixer in {\name}.
When performing up-conversion, this computing mixer yields the output waveform as a time-domain multiplication of the two input waveforms~\cite{gao2026disaggregated} as
\begin{align}
    \convWaveOutput(\waveIdx) = \convWaveInput(\waveIdx) \cdot \convWaveWeight(\waveIdx), ~\forall \waveIdx \in [0, \waveLen).
    \label{eq: supplementary-linear-convolution-up-convertion}
\end{align}
Based on the convolution theorem, the frequency domain representation of $\convWaveOutput(\waveIdx)$ is the convolution between $\convInputVec$ and $\convWeightVec$, which contains a total number of $\convOutputLen$ tones spaced at $\freqSub$ and spanning over a bandwidth of $\convOutputLen \cdot \freqSub$.
Therefore, to fully receive this signal without frequency aliasing, an ADC operating at a sampling rate of $\convOutputLen \cdot \freqSub$ is required to sample this output waveform, which yields an $\convOutputLen$-point I/Q sample sequence in digital as $\convSampOutputVec \in \mathbb{C}^{\convOutputLen}$.
The convolution results, $\convOutputVec$, between $\convInputVec$ and $\convWeightVec$, can be extracted from $\convSampOutputVec$ by performing an $\convOutputLen$-point digital FFT, i.e.,
\begin{align}
    \convOutputVec = \fft{\convSampOutputVec},\ \text{where}~
    \convOutputScalar[\convIdx] = \sum_{\convIdxTemp=0}^{\convOutputLen-1} \convSampOutputScalar[\convIdxTemp] \cdot \eu^{-\iu 2 \pi \frac{\convIdxTemp \convIdx}{\convOutputLen-1}}, ~\forall \convIdx = 0, 1, \dots, \convOutputLen-1.
    \label{eq: supplementary-linear-convolution-decoding-whole}
\end{align}
This $\convOutputLen$-point FFT contains a total number of $(\frac{\convOutputLen}{2} \cdot \log_2 \convOutputLen)$ complex-valued MACs, or $(2 \convOutputLen \cdot \log_2 \convOutputLen)$ real-valued MACs.
The waveform analysis diagram for $\convOutputVec$ is shown in the right half of Fig.~\ref{fig:supplementary-convolution-diagram}(a).

\myparatight{Waveform analysis for partial convolution output}
In practice, only a subset of elements in the output is needed, which are usually contiguous.
For example, we consider the desired output vector $\convOutputSubVec \in \mathbb{C}^{\convOutputSubLen}$ elements indexed at $\left\{\convOutputSubStart, \convOutputSubStart+1, \dots, \convOutputSubEnd-1\right\}$ on the full output vector $\convOutputVec$, where $0 \leq \convOutputSubStart < \convOutputSubEnd \leq \convOutputLen$ and $\convOutputSubLen = \convOutputSubEnd-\convOutputSubStart$.
Hereby, we only need to examine the middle $\convOutputSubLen$ tones indexed at $\left\{\convOutputSubStart, \convOutputSubStart+1, \dots, \convOutputSubEnd-1\right\}$ in the frequency domain. 
To achieve this, we first apply an additional frequency mixer, which down-converts $\convWaveOutput(\waveIdx)$ by $\convOutputSubStart \cdot \freqSub$. 
This shifts the desired $\convOutputSubLen$ tones to the baseband at the frequencies of $\{0, \freqSub, \dots, \convOutputSubLen\cdot \freqSub\}$. 
Then, a low-pass filter (LPF) with a stop band at $\convOutputSubLen \cdot \freqSub$ is applied, which only passes the desired $\convOutputSubLen$ tones.
We denote this filtered waveform as $\convWaveOutputSub(\waveIdx)$, which still lasts for a time duration of $\waveLen$ but occupies a reduced bandwidth of $\convOutputSubLen \cdot \freqSub$.
Given its smaller bandwidth, a pair of ADCs operating at a sampling rate of $\convOutputSubLen \cdot \freqSub$ is sufficient to sample this down-converted waveform $\convWaveOutputSub(\waveIdx)$ using I/Q demodulation, which yields an $\convOutputSubLen$-point I/Q sample sequence in digital, denoted by $\convSampOutputSubVec \in \mathbb{C}^{\convOutputSubLen}$.
Then, an $\convOutputSubLen$-point FFT is applied on $\convSampOutputSubVec$ to extract the desired output vector $\convOutputSubVec$ as
\begin{align}
    \convOutputSubVec = \fft{\convSampOutputSubVec},\ \text{where}~
    \convOutputSubScalar[\convIdx] = \sum_{\convIdxTemp=0}^{\convOutputSubLen-1} \convSampOutputSubScalar[\convIdxTemp] \cdot \eu^{-\iu 2 \pi \frac{\convIdxTemp \convIdx}{\convOutputSubLen-1}}, ~\forall \convIdx = 0, 1, \dots, \convOutputSubLen-1,
    \label{eq: supplementary-linear-convolution-decoding-partial}
\end{align}
which requires a total number of $(2 \convOutputSubLen \cdot \log_2 \convOutputSubLen)$ real-valued MACs.
The whole waveform analysis diagram for $\convOutputSubVec$ is illustrated in Fig.~\ref{fig:supplementary-convolution-diagram}(b).

\myparatight{Cross-correlation by down-conversion}
When the computing mixer is configured and used for down-conversion, the output waveform becomes the time-domain multiplication of $\convWaveInput(\waveIdx)$ and the conjugate of $\convWaveWeight(\waveIdx)$~\cite{gao2026disaggregated}.
In this case, equation~\eqref{eq: supplementary-linear-convolution-up-convertion} can be written as
\begin{align}
    \convWaveOutput^{\prime}(\waveIdx) = \convWaveInput(\waveIdx) \cdot \conj{\convWaveWeight}(\waveIdx), ~\forall \waveIdx \in [0, \waveLen).
    \label{eq: supplementary-linear-convolution-down-convertion}
\end{align}
This conjugate in the time domain reflects vector-flipping in the frequency domain for $\specWeightVec$, and therefore the convolution becomes the cross-correlation defined in equation~\eqref{eq: supplementary-linear-convolution-cross-correlation}.
This means that the output of equation~\eqref{eq: supplementary-linear-convolution-decoding-whole} becomes $\convOutputVec^{\prime} = \convInputVec \correlation \convWeightVec$, and the output of equation~\eqref{eq: supplementary-linear-convolution-decoding-partial} becomes $\convOutputSubVec^{\prime}$, which is a subset of $\convOutputVec^{\prime}$.

\subsection{Physical Channel Calibration}
\label{ssec: supplementary-channel-calibration}

The transmission of waveforms, $\convWaveInput(\waveIdx)$, $\convWaveWeight(\waveIdx)$, and $\convWaveOutput(\waveIdx)$, or their frequency-domain tone series, $\convInputVec$, $\convWeightVec$, and $\convOutputVec$, through a physical channel introduces distortions that must be calibrated before convolution computing.
This distortion comes from the unflattened frequency response introduced by the embedded anti-aliasing filters for the wired channel, or the multipath effect or timing delay for the wireless channel.

\myparatight{Physical channel modeling}
We model the physical channel in the frequency domain, which element-wisely multiplies to the corresponding tone series.
We formulate this process as a function as $\convCalibOutputVec = \convChannelFunc{\convCalibInputVec, \convCalibWeightVec}$, where $\convChannelInputVec \in \mathbb{C}^{\convInputLen}$ and $\convChannelWeightVec \in \mathbb{C}^{\convWeightLen}$ are the input tone series, and $\convCalibOutputVec \in \mathbb{C}^{\convOutputLen}$ is the output tone series after the physical channel.
Specifically, we define $\convChannelInputVec \in \mathbb{C}^{\convInputLen}$, $\convChannelWeightVec \in \mathbb{C}^{\convWeightLen}$, and $\convChannelOutputVec \in \mathbb{C}^{\convOutputLen}$ as the channel frequency response for tone series $\convCalibInputVec$, $\convCalibWeightVec$ and $\convCalibOutputVec$, respectively.
In addition, we denote the element-wise multiplication (or Hadamard multiplication) between two equal dimension vectors by $\odot$, as well as the element-wise division (or Hadamard division) by $\oslash$.
Generally, these channel responses are complex-valued, and their amplitudes and phases are continuous and smooth over the frequency.
Then, the distorted version of equation~\eqref{eq: supplementary-linear-convolution-linear-convolution} including the physical channel distortion can be rewritten as
\begin{align}
    \convCalibOutputVec &= \convChannelFunc{\convCalibInputVec, \convCalibWeightVec} = \convChannelOutputVec \odot \left( \left( \convChannelInputVec \odot \convCalibInputVec \right) \convolution \left( \convChannelWeightVec \odot \convCalibWeightVec \right) \right) \\
    \text{where}~\convCalibOutputScalar[\convIdx] &= \convChannelOutputScalar[\convIdx] \cdot \sum_{\convIdxTemp=\max\{0, \convIdx-\convWeightLen+1\}}^{\min\{\convInputLen-1, \convIdx\}} \convChannelInputScalar[\convIdxTemp] \cdot \convCalibInputScalar[\convIdxTemp] \cdot \convChannelWeightScalar[\convIdx-\convIdxTemp] \cdot \convCalibWeightScalar[\convIdx-\convIdxTemp], \\
    \forall \convIdx &= 0, 1, \dots, \convInputLen+\convWeightLen-2.
\end{align}

\myparatight{Pre- and post-channel calibration}
To calculate the convolution $\convOutputVec = \convInputVec \convolution \convWeightVec$, we pre-compensate the physical channel distortion by pre-calibrating the input tone series $\convCalibInputVec$ and $\convCalibWeightVec$ before the transmission as
\begin{align}
    \begin{cases}
        \convCalibInputVec = \convInputVec \oslash \convChannelInputVec, \quad \text{where}~ \convCalibInputScalar[\convIdx] = \convInputScalar[\convIdx] / \convChannelInputScalar[\convIdx], \quad \forall \convIdx = 0, 1, \dots, \convInputLen-1 \\
        \convCalibWeightVec = \convWeightVec \oslash \convChannelWeightVec, \quad \text{where}~ \convCalibWeightScalar[\convIdx] = \convWeightScalar[\convIdx] / \convChannelWeightScalar[\convIdx], \quad \forall \convIdx = 0, 1, \dots, \convWeightLen-1,
    \end{cases}
\end{align}
as well as post-calibrating the output tone series $\convCalibOutputVec$ after the reception by
\begin{align}
    \convOutputVec = \convCalibOutputVec \oslash \convChannelOutputVec, \quad \text{where}~ \convOutputScalar[\convIdx] = \convCalibOutputScalar[\convIdx] / \convChannelOutputScalar[\convIdx], \quad \forall \convIdx = 0, 1, \dots, \convOutputLen-1.
\end{align}
This process is illustrated in Fig.~\ref{fig:supplementary-convolution-diagram}(c).
This indicates that the key to the physical channel calibration is to estimate the channel response vectors, $\convChannelInputVec$, $\convChannelWeightVec$, and $\convChannelOutputVec$.
In practice, these pre- and post-calibration can be fused with the IFFT and FFT operations in equations~\eqref{eq: supplementary-linear-convolution-encoding} and~\eqref{eq: supplementary-linear-convolution-decoding-whole} without introducing additional digital computing costs.

\myparatight{Physical channel estimation using pre-defined training sequences}
The physical channels, $\convChannelInputVec$, $\convChannelWeightVec$, and $\convChannelOutputVec$ can be estimated using pre-defined training sequences sent through them and the minimum mean squared error (MMSE) method.
Specifically, we randomize a set of $\convChannelInputVec^{(i)}$, $\convChannelWeightVec^{(i)}$, and conduct the transmission process to obtain the corresponding output tone series $\widehat{\convCalibOutputVec}^{(i)}$.
The channel estimation can be formulated as solving the following optimization problem:
\begin{align}
    \left\{ \convChannelInputVec^{\star}, \convChannelWeightVec^{\star}, \convChannelOutputVec^{\star} \right\}
    & = \arg \min_{\convChannelInputVec, \convChannelWeightVec, \convChannelOutputVec} \sum_{i} \abs{\widehat{\convCalibOutputVec}^{(i)} - \convChannelFunc{\convCalibInputVec^{(i)}, \convCalibWeightVec^{(i)}}}^2 \\
    & = \arg \min_{\convChannelInputVec, \convChannelWeightVec, \convChannelOutputVec} \sum_{i} \abs{\widehat{\convCalibOutputVec}^{(i)} - \convChannelOutputVec \odot \left( \left( \convChannelInputVec \odot \convCalibInputVec^{(i)} \right) \convolution \left( \convChannelWeightVec \odot \convCalibWeightVec^{(i)} \right) \right)}^2.
\end{align}
We adopt the Adam optimizer~\cite{kingma2014adam} with a learning rate of $10^{-3}$, and run iterations until the MMSE converges.
Generally, smaller values of $\convInputLen$, $\convWeightLen$, and $\convOutputLen$ may not be sufficient to accurately describe the physical channel, while larger values may introduce too many parameters and therefore lead to over-fitting.
We empirically select $\convInputLen=\convWeightLen=1,000$ and $\convOutputLen=1,999$ to optimize $\convChannelInputVec^{\star}$, $\convChannelWeightVec^{\star}$, and $\convChannelOutputVec^{\star}$ once, and use linear interpolation respectively for amplitudes and phases to obtain the physical channels at other lengths.
Under this setting, we randomize {100} such training sequences for the optimization.

\subsection{Phase Calibration with Preambles and Postambles}
\label{ssec: supplementary-phase-calibration}

The input and weight waveforms of {\name} are structured into (wired) input and (wireless) weight packets. These two packets are expected to be aligned in time and fed into the computing mixer for computing the target convolution operations, as shown in Fig.~\ref{fig:supplementary-phase-offset}(a).
Typically, the two transmitters ($\convWaveInput(\waveIdx)$ and $\convWaveWeight(\waveIdx)$) and the receiver ($\convWaveOutputSub(\waveIdx)$) are not synchronized in time and frequency, requiring a detection process at the receiver. This process identifies the starting point of the transmitted waveforms in time and aligns the carrier frequency at the receiver.
Specifically, we follow the detection process of the low-sampling-rate Wi-Fi monitoring system~\cite{gao2023swirls} in the implementation of {\name}.
In this section, we mainly focus on the output packet and its phase calibration.

\myparatight{Packet formulation and time-frequency representation}
As shown in Fig.~\ref{fig:supplementary-phase-offset}(a), a packet consists of $(\symNum+2)$ symbols concatenated over time: $\symNum$ payload symbols sandwiched between two pre-defined preamble and postamble symbols.
These $\symNum$ payload symbols are used to perform $\symNum$ equal-dimension convolution operations (denoting its output length as $\convOutputSubLen$, and tone spacing as $\freqOffset$), while the preamble and postamble symbols are used for phase estimation and calibration.
The waveforms of these $\symNum+2$ symbols in the time domain can be converted into the frequency domain using $\convOutputSubLen$-point FFT; by concatenating these frequency-domain representations, we obtain the time-frequency representation of the output packet, denoted as $\packMat \in \mathbb{C}^{(\symNum+2) \times \convOutputSubLen}$, as shown in Fig.~\ref{fig:supplementary-phase-offset}(b).
Specifically, $\packMat$ can be further divided into $\packPayMat \in \mathbb{C}^{\symNum \times \convOutputSubLen}$ for the payload, as well as $\packPreMat \in \mathbb{C}^{\convOutputSubLen}$ and $\packPostMat \in \mathbb{C}^{\convOutputSubLen}$ for the preamble and postamble symbols.

\myparatight{Time-frequency phase offset formulation}
In practice, the phases of the experimentally observed $\packExpMat$ are offset from the expected one $\packMat$, and this offset varies linearly over time (due to the carrier frequency offset of the receiver), and over frequency (due to the packet detection timing offset on the receiver).
Denoting the carrier frequency offset as $\freqOffset$ and the timing offset as $\timeOffset$, the experimental $\packExpMat$ with the phase offset can be formulated as
\begin{align}
    \packExpScalar[\symIdx, \convIdx] = \packScalar[\symIdx, \convIdx] \cdot \eu^{\iu 2 \pi \cdot \left( \symIdx \cdot \frac{\freqOffset}{\freqSub} + \convIdx \cdot \timeOffset \freqSub \right)},\ \forall \symIdx = 0, 1, \dots, \symNum+1,\ \forall \convIdx = 0, 1, \dots, \convOutputSubLen-1,
    \label{eq: supplementary-phase-offset-formulation}
\end{align}
where $\freqSub$ is the tone spacing (see definition in {\supplementary}~\ref{ssec: supplementary-theory-linear-convolution}). This is demonstrated in Fig.~\ref{fig:supplementary-phase-offset}(c).
Breaking it down, the experimental preamble and postamble symbols ($\packPreExpMat$ and $\packPostExpMat$) are with phase offsets given by
\begin{align}
    \packPreExpScalar[\convIdx] &= \packPreScalar[\convIdx] \cdot \eu^{\iu 2 \pi \cdot \left( 0 \cdot \frac{\freqOffset}{\freqSub} + \convIdx \cdot \timeOffset \freqSub \right)},\ \forall \convIdx = 0, 1, \dots, \convOutputSubLen-1; \\
    \packPostExpScalar[\convIdx] &= \packPostScalar[\convIdx] \cdot \eu^{\iu 2 \pi \cdot \left( (\symNum+1) \cdot \frac{\freqOffset}{\freqSub} + \convIdx \cdot \timeOffset \freqSub \right)},\ \forall \convIdx = 0, 1, \dots, \convOutputSubLen-1.
\end{align}
Similarly, the experimental payload symbols with the phase offset can be expressed as
\begin{align}
    \packPayExpScalar[\symIdx, \convIdx] = \packPayScalar[\symIdx, \convIdx] \cdot \eu^{\iu 2 \pi \cdot \left( (\symIdx+1) \cdot \frac{\freqOffset}{\freqSub} + \convIdx \cdot \timeOffset \freqSub \right)},\
    \forall \symIdx = 0, 1, \dots, \symNum-1,\ \forall \convIdx = 0, 1, \dots, \convOutputSubLen-1.
\end{align}
Therefore, to correct this phase offset, we need to estimate the two unknown variables, $\freqOffset$ and $\timeOffset$.
In contrast to the relatively stable channel calibration discussed in {\supplementary}~\ref{ssec: supplementary-channel-calibration}, the values of $\freqOffset$ and $\timeOffset$ fluctuate unpredictably due to variations in packet detection and synchronization and, as a result, this estimation and calibration must be conducted for every packet.

\myparatight{Packet preamble and postamble design}
We pre-define the frequency-domain preambles on the input and weight packets as vectors with uniform amplitudes $\mathcal{U}(0, 1)$ and random phases following the independent uniform distribution $\mathcal{U}(-\pi, \pi)$. 
In this way, the transmitted time-domain preamble waveforms have a low PAPR to alleviate the saturation issue (see {\supplementary}~\ref{ssec: supplememtary-waveform-papr}).
In addition, this yields a relatively flat power distribution in the frequency domain after the convolution operation, ensuring good SNR values of $\packPreMat$ for any choices of $\convOutputSubStart$ and $\convOutputSubEnd$ (see definition in {\supplementary}~\ref{ssec: supplementary-theory-linear-convolution}).
The postamble $\packPostMat$ is the duplication of the preamble $\packPreMat$.

\myparatight{Carrier frequency offset estimation}
The carrier frequency offset $\freqOffset$ is estimated by comparing the phase offset between the preamble and postamble symbols.
Specifically, we have 
\begin{align}
    \freqOffset = \frac{\freqSub}{2 \pi (\symNum+1)} \cdot \ang{} \left[ \sum_{\convIdx=0}^{\convOutputSubLen-1} \packPreExpScalar[\convIdx] \cdot \conj{\packPostExpScalar}[\convIdx] \right].
\end{align}
where $\ang(\cdot)$ and $\conj{(\cdot)}$ denote the angle and conjugate of a complex number.

\myparatight{Timing offset estimation}
The timing offset $\timeOffset$ is estimated by independently by examining the phase offset over frequency on both the preamble or postamble symbol, and then averaging the two results to improve the accuracy.
This process can be written as
\begin{align}
    \timeOffset &= \frac{1}{2 \pi \freqSub} \cdot \left( \frac{\sum_{\convIdx=0}^{\convOutputSubLen-2} \ang{} \left[ \packPreExpScalar[\convIdx] \cdot \conj{\packPreExpScalar}[\convIdx+1] \right]}{\convOutputSubLen-1} + \frac{\sum_{\convIdx=0}^{\convOutputSubLen-2} \ang{} \left[ \packPostExpScalar[\convIdx] \cdot \conj{\packPostExpScalar}[\convIdx+1] \right]}{\convOutputSubLen-1} \right).
\end{align}

\subsection{Waveform PAPR Alleviation}
\label{ssec: supplememtary-waveform-papr}

Consider a DAC generating a waveform of $\convWave(\waveIdx)$ from the time-domain I/Q sample sequence $\convSampVec$.

\myparatight{Saturation and quantization error of a DAC}
In practice, a DAC has a maximum output waveform range for $\convWave(\waveIdx)$, beyond which the generated analog waveforms in analog might be saturated and will be truncated to the maximum output voltage.
This maximum voltage corresponds to a maximum amplitude of the input I/Q waveform for $\convSampVec$, which can be formulated as a normalized constraint given by
\begin{align}
    \max_{\convIdx} \abs{\convSampScalar[\convIdx]} \leq 1.
    \label{eq: supplementary-papr-dac-constraint}
\end{align}
At the same time, a DAC has a finite resolution for the input I/Q samples $\convSampVec$, which introduces quantization errors when the input sample amplitude $\convSampScalar_{\convIdx}$ is too small.
Specifically, a $\bit$-bit DAC has a constant quantization step of $2^{-(\bit-1)}$ with $\convSampScalar[\convIdx] \in \left[ -1, +1\right]$, corresponding to a mean squared quantization error of $\left(3 \cdot 4^{\bit} \right)^{-1}$~\cite{lathi2019modern}.
The average power of the output waveform $\convWave(\waveIdx)$ is proportional to the mean squared value of $\convSampVec$, i.e., $\frac{1}{\convLen} \sum_{\convIdx=0}^{\convLen-1} \abs{\convSampScalar[\convIdx]}^2$.
A smaller average power of $\convSampVec$ leads to a smaller signal-to-quantization-noise ratio (SQNR) of the output waveform $\convWave(\waveIdx)$, given by
\begin{align}
    \sqnr{\convSampVec}
    = 10 \log_{10} \left( \frac{\frac{1}{\convLen} \sum_{\convIdx=0}^{\convLen-1} \abs{\convSampScalar[\convIdx]}^2}{\left(3 \cdot 4^{\bit} \right)^{-1}} \right)
    = 10 \log_{10} \left( \frac{1}{\convLen} \sum_{\convIdx=0}^{\convLen-1} \abs{\convSampScalar[\convIdx]}^2 \right) + 4.77 + 6.02\bit.
    \label{eq: supplementary-papr-sqnr-definition}
\end{align}
Therefore, we notice a trade-off between the peak power and average power of the waveform on the amplitude of $\convSampVec$: a smaller $\abs{\convSampScalar[\convIdx]}$ is less likely to be saturated, while yielding a smaller SQNR.

\myparatight{Peak-to-average power ratio (PAPR) and waveform backoff}
The PAPR of the I/Q sample sequence $\convSampVec$ can be calculated as
\begin{align}
    \papr{\convSampVec} = 10 \log_{10} \left( \frac{\max_{\convIdx} \abs{\convSampScalar[\convIdx]}^2}{\frac{1}{\convLen} \sum_{\convIdx=0}^{\convLen-1} \abs{\convSampScalar[\convIdx]}^2} \right).
    \label{eq: supplementary-papr-papr-definition}
\end{align}
Putting equations~\eqref{eq: supplementary-papr-dac-constraint}, \eqref{eq: supplementary-papr-sqnr-definition}, and~\eqref{eq: supplementary-papr-papr-definition} together, we have
\begin{align}
    \sqnr{\convSampVec} \leq - \papr{\convSampVec} + 4.77 + 6.02\bit.
\end{align}
This indicates that a smaller PAPR of $\convSampVec$ leads to a higher upper bound of the corresponding SQNR, which is beneficial for alleviating the quantization error introduced by the finite-resolution DAC.
In practice, we define a backoff coefficient $\backoff$ that scales the amplitude of the transmitted I/Q waveform $\convSampVec$ so that its mean square becomes $1/\backoff$.
Under this definition, the scaled waveform $\convSampVec^{\prime}$ is given by:
\begin{align}
    \convSampVec^{\prime} = \sqrt{\frac{1}{\backoff \cdot \frac{1}{\convLen} \sum_{\convIdxTemp=0}^{\convLen-1} \abs{\convSampScalar[\convIdxTemp]}^2}} \cdot \convSampVec,\ \text{where}~ \convSampScalar^{\prime}[\convIdx] = \sqrt{\frac{1}{\backoff \cdot \frac{1}{\convLen} \sum_{\convIdxTemp=0}^{\convLen-1} \abs{\convSampScalar[\convIdxTemp]}^2}} \cdot \convSampScalar[\convIdx],\ \forall \convIdx = 0, 1, \dots, \convLen-1.
\end{align}
In this way, the scaled I/Q waveform $\convSampVec^{\prime}$ with a backoff coefficient of $\backoff$ can tolerate a maximum PAPR of $10 \log_{10}\backoff$ (dB).
However, plugging in equation~\eqref{eq: supplementary-papr-sqnr-definition}, its SQNR is reduced to
\begin{align}
    \sqnr{\convSampVec^{\prime}} = - 10 \log_{10} \backoff + 4.77 + 6.02\bit.
\end{align}
To summarize, a higher PAPR of $\convSampVec$ requires a larger backoff coefficient $\backoff$ to avoid saturation, which leads to a smaller SQNR.
In addition, a larger backoff coefficient $\backoff$ requires a higher amplification gain to compensate for the reduced amplitude of the transmitted I/Q waveform $\convSampVec^{\prime}$, which is limited by the hardware constraints of the employed radios (see {\supplementary}~\ref{ssec: supplementary-experiment-testbed}).
Therefore, mitigating the PAPR of $\convSampVec$ is critical to the computing accuracy of {\name}'s analog computing.

\myparatight{Mitigating the impact of PAPR using Zadoff-Chu phases over input channels}
In today's DL workloads, the input tensor $\inputTensor$ is usually nonnegative for two main reasons:
(\emph{i}) for the first {\convtwodim} layer, the input image RGB pixels are non-negative within the range of $[0, 255]$ for the three input channels; and
(\emph{ii}) for the following {\convtwodim} or FC layers, the input tensor goes through ReLU, which yields nonnegative values.
Such a nonnegative input tensor $\inputTensor$ results in a strong direct current (DC) component in the time domain I/Q sample sequence after IFFT, which leads to a high PAPR.
To mitigate this issue, we employ a $\chanInNum$-point Zadoff-Chu sequence, denoted as $\zadoffVec = \left[ \zadoffScalar_{\chanInIdx} \right] \in \left[ -\pi, +\pi \right]^{\chanInNum}$, where
\begin{align}
    \zadoffScalar_{\chanInIdx} = \eu^{-\iu \pi \frac{\chanInIdx (\chanInIdx+1)}{\chanInNum}},\ \forall \chanInIdx = 0, 1, \dots, \chanInNum-1.
\end{align}
Specifically, we respectively multiply this Zadoff-Chu sequence to all the $\chanInNum$ input channels on the input tensor $\inputTensor$ before mapping to the frequency-domain tone series; also, we multiply the element-wise inverse Zadoff-Chu sequence to all the $\chanInNum$ weight channels on the weight tensor $\weightTensor$.
It can be easily verified that this operation does not change the final convolution output $\outputTensor$.
On the other hand, this operation effectively spreads the energy of the input tensor $\inputTensor$ over the time domain I/Q sample sequence after IFFT, thereby mitigating the PAPR issue.

%% file: tex/Supplementary_experiment.tex
\section*{\textbf{Supplementary Information: Experiment}}

\subsection{Additive White Gaussian Noise Channel based Simulations}
\label{ssec: supplementary-gaussian-simulation}

We simulate {\name}'s performance on the ML tasks for both classification models and the generative models.

In the simulation, we employ the additive white Gaussian noise (AWGN) channel with no channel distortions, where the computing accuracy is impacted by the Gaussian noise.
Also, we consider an ideal analog multiplication on the computing mixer.
Specifically, given the two waveforms $\convWaveInput(\waveIdx)$ and $\convWaveWeight(\waveIdx)$, the simulated process of the computing mixer in equation~\eqref{eq: supplementary-linear-convolution-up-convertion} can be rewritten as 
\begin{align}
    \convWaveOutput(\waveIdx) = \convWaveInput(\waveIdx) \cdot \convWaveWeight(\waveIdx) + \convWaveNoise(\waveIdx), ~\forall \waveIdx \in [0, \waveLen).
    \label{eq: supplementary-simulation-channel-model}
\end{align}
where $\convWaveNoise(\waveIdx)$ is a complex-valued Gaussian noise.
Theoretically, the PSD of the Gaussian noise is given by {-174}\thinspace{dBm/Hz}. However, this PSD is not practical, and is usually much higher than this number due to the noise figure of the radio RX, insertion loss of the frequency mixer, etc.
To align the simulation results with experiments, we define the equivalent noise PSD, denoted as $\PSDNoise$, to include the factors mentioned above. This equivalent noise PSD may vary over different convolutional layers and, thus, different ML tasks.

\subsection{Software-Defined Radio Testbed for Experiments}
\label{ssec: supplementary-experiment-testbed}

We implement and evaluate {\name} on a software-defined radio (SDR) testbed, which consists of a central radio that transmits/broadcasts the model weights, and one edge client that performs the analog computing locally, as shown in Fig.~\ref{fig:supplementary-experiment-setup}(a).
There are two main setups for our experiments: (\emph{i}) a wired setup, where the model weights are delivered via a cable, and (\emph{ii}) a wireless setup, where the model weights are delivered over a wireless channel.
Specifically in the wireless setup, the central radio is implemented by a radio transmitter ($\weightTensor$) and a directional antenna; the edge client is comprised of one radio transmitter ($\inputTensor$), one radio receiver ($\outputTensor$), a frequency mixer for analog computing, and a directional antenna.
In the SDR testbed, we utilize the USRP X310, a high-performance SDR with a UBX-160 daughterboard, as the transmitter and receiver in our SDR testbed; the Mini-Circuits ZEM-4300+~\cite{ZEM-4300+} as the computing frequency mixer; and the Tupavco TP514 Yagi directional antenna for wireless transmission.

\myparatight{USRP X310 as the radio transmitter}
As a radio transmitter, the USRP X310 can be summarized into three main components: (\emph{i}) a pair of I/Q DACs, which supports a sampling rate from {0.196}\thinspace{MHz} to {200}\thinspace{MHz} with a resolution of 16 bits; (\emph{ii}) an internal frequency mixer for up-converting the baseband I/Q waveforms to the RF band at the carrier frequency from {10}\thinspace{MHz} to {6}\thinspace{GHz}; and (\emph{iii}) a pair of power amplifiers (PAs) with a tunable gain at {0--31.5}\thinspace{dB}, as shown in Fig.~\ref{fig:supplementary-experiment-setup}(b).
The maximum transmit power of the USRP X310 is approximately {+23.5}\thinspace{dBm} with a {0}\thinspace{dB}-PAPR-waveform, and is reduced to {+23.5-$\backoff$}\thinspace{dBm} given the backoff coefficient of $\backoff$, as discussed in {\supplementary}~\ref{ssec: supplememtary-waveform-papr}.

\myparatight{USRP X310 as the radio receiver}
As shown in Fig.~\ref{fig:supplementary-experiment-setup}(b), there are three components when the USRP X310 serves as a radio receiver: (\emph{i}) an internal frequency mixer that down-converts the RF signals with a carrier frequency from {10}\thinspace{MHz} to {6}\thinspace{GHz}; (\emph{ii}) low-pass filters (LPF) with a cutoff frequency of half of the set sampling rate to avoid frequency aliasing; and (\emph{iii}) a pair of I/Q ADCs, which supports a sampling rate from {0.196}\thinspace{MHz} to {200}\thinspace{MHz} with a resolution of 14 bits.

\myparatight{ZEM-4300+ frequency mixer as the computing mixer}
The Mini-Circuits ZEM-4300+~\cite{ZEM-4300+} is a passive double-balanced diode mixer for either frequency down-conversion or up-conversion.
In {\name}, we consider the down-conversion case, where its LO and RF ports are configured as inputs and the IF port as output, as shown in Fig.~\ref{fig:supplementary-experiment-setup}(c).
For the conventional usage in a communication system, the input power to the LO port is typically around {+7}\thinspace{dBm} to drive the diodes much higher than its voltage threshold, where the frequency mixer behaves as an on-off switch that modulates the RF waveform, $\convWave_{\textrm{RF}}(\waveIdx)$ with a switching pattern determined by the (conjugated) LO waveform $\convWave_{\textrm{LO}}(\waveIdx)$ as
\begin{align}
    \convWave_{\textrm{IF}}(\waveIdx) \propto \textsf{sgn}(\conj{\convWave}_{\textrm{LO}}(\waveIdx)) \cdot \convWave_{\textrm{RF}}(\waveIdx),
\end{align}
where $\textsf{sgn}(\cdot)$ is the sign function.
In {\name}'s computing usage, the input power to the LO port is reduced to approximately {-3}\thinspace{dBm} (see measurements in {\supplementary}~\ref{ssec: supplementary-experiment-benchmark}), which drives the diodes around the threshold voltage, where its behavior is similar to the analog multiplier that
\begin{align}
    \convWave_{\textrm{IF}}(\waveIdx) \propto \conj{\convWave}_{\textrm{LO}}(\waveIdx) \cdot \convWave_{\textrm{RF}}(\waveIdx).
\end{align}
On the other hand, ZEM-4300+ supports the input carrier frequency range of {0.3--4.3}\thinspace{GHz} for the LO and RF ports, and a carrier frequency range of {0--1.0}\thinspace{GHz} for the IF port, which conforms {\name}'s computing usage (see {\supplementary}~\ref{ssec: supplementary-experiment-benchmark}).

\myparatight{Tupavco TP514 Yagi directional antenna}
For the wireless setup, we use a pair of Tupavco TP514 Yagi directional antenna as the TX/RX antenna to establish the wireless link between the central radio and the edge client.
This Yagi antenna is designed for dual frequency bands at {0.80--0.96}\thinspace{GHz} and {1.7--2.5}\thinspace{GHz}, covering the ISM band at {0.915}\thinspace{GHz} utilized in our experiments.
In addition, it provides an antenna gain of {9}\thinspace{dBi} in the designated direction for both TX and RX.
Given the backoff coefficient $\backoff=0.04$ (tolerating PAPR up to {14}\thinspace{dB} without saturation), and the maximum transmission power of {+23.5}\thinspace{dBm} by USRP X310, the wireless link distance is recommended to be around {1}\thinspace{m} to maintain a receiving power of {-3}\thinspace{dBm} on the edge client.

\subsection{Frequency Mixer Computing Accuracy Benchmark}
\label{ssec: supplementary-experiment-benchmark}

We first benchmark and optimize the computing accuracy of the employed frequency mixer, ZEM-4300+~\cite{ZEM-4300+}, over two varying factors: (\emph{i}) the carrier frequency to the RF port, and (\emph{ii}) the input power levels to the RF (for $\inputTensor$) and LO (for $\weightTensor$) ports.
Note that these benchmarks are performed in the wired setup to avoid the impact of wireless channel impairments.

\myparatight{General-purpose {\convtwodim} operation setup}
We employ a set of default general-purpose {\convtwodim} operations to benchmark the computing accuracy.
Specifically, we consider a {\convtwodim} layer (see definition in {\supplementary}~\ref{ssec: supplementary-theory-convolution-basics}) with the number of input channels $\chanInNum \in \left\{ 64, 128, 256\right\}$, and the number of output channels $\chanOutNum = 1$, and a kernel dimension of $\kernelSizeX=\kernelSizeY=3$; the input image size is set to $\imageSizeX=\imageSizeY=16$.
Note that we vary the number of input channels $\chanInNum$ for their different downsampling ratios (see {\autorefmethod}).
We randomize $\inputTensor$ and $\weightTensor$ with i.i.d. Gaussian distributed amplitudes from $\mathcal{N}(0, 1)$, and uniformly distributed phases from $\mathcal{U}(-\pi, +\pi)$.
This randomization is repeated for 100 times to obtain statistically meaningful results.
In addition, limited by the unlicensed ISM band~\cite{ISM}, the wireless transmission of $\weightTensor$ is always constrained to the carrier frequency of {915}\thinspace{MHz} and a bandwidth of {25}\thinspace{MHz}.

\myparatight{Computing accuracy metrics: normalized RMSE and resolution bit}
To quantify the computing accuracy, we consider two metrics:
(\emph{i}) the normalized root mean square error (RMSE), and
(\emph{ii}) the equivalent resolution bit.
For the digital computed $\outputTensor$, we first normalize it so that its standard deviation equals $1/3$, i.e., $\outputTensor \leftarrow \outputTensor / (3 \cdot \textsf{std}(\outputTensor))$. After this normalization, the real and imaginary components of $\outputTensor$ mostly fall within the range of $[-1, +1]$.
Then, we normalize the analog computed $\outputTensorEst$ by the same standard deviation as $\outputTensorEst \leftarrow \outputTensorEst / (3 \cdot \textsf{std}(\outputTensor))$.
Hence, we define the normalized RMSE as:
\begin{align}
    \textsf{RMSE} = \sqrt{\mathbb{E} \left[ \frac{1}{\chanOutNum \cdot \imageSizeX \cdot \imageSizeY} \sum_{\chanOutIdx=1}^{\chanOutNum} \sum_{\imageIdxX=1}^{\imageSizeX} \sum_{\imageIdxY=1}^{\imageSizeY} \abs{ \outputScalar{\chanOutIdx, \imageIdxX, \imageIdxY} - \outputScalarEst{\chanOutIdx, \imageIdxX, \imageIdxY} }^2 \right] }.
    \label{eq: supplementary-experiment-benchmark-rmse}
\end{align}
Next, we define the resolution bit as $-\log_2(\textsf{RMSE}/2)$, where the division by $2$ comes from $\outputTensor$'s range of $[-1, +1]$.

\myparatight{Benchmark the carrier frequency to RF port}
We vary the carrier frequency of $\inputTensor$ to benchmark the computing accuracy, while fixing that of $\weightTensor$ at {0.915}\thinspace{GHz} of the ISM band.
In the down-conversion, the carrier frequency of $\outputTensor$ is the difference between $\inputTensor$ and $\weightTensor$.
Specifically, the carrier frequency of $\inputTensor$ is varied from {1.00}\thinspace{GHz} to {3.00}\thinspace{GHz} with a step of {0.05}\thinspace{GHz}, while fixing the input PSDs to the RF (for $\inputTensor$) and LO ports (for $\weightTensor$) to {-102}\thinspace{dBm/Hz} and {-77}\thinspace{dBm/Hz}, respectively. These correspond to the power levels of {-28}\thinspace{dBm} and {-3}\thinspace{dBm} over a bandwidth of {25}\thinspace{MHz}.
As shown in Fig.~\ref{fig:supplementary-mixer-carrier}, the trendency is similar over the three considered numbers of input channels $\chanInNum \in \left\{ 64, 128, 256\right\}$.
Specifically for $\chanInNum=128$, the normalized RMSE is around {0.056--0.064} (approximately 5-bit resolution) when the carrier frequency of $\inputTensor$ is set to {1.00--2.50}\thinspace{GHz}, excluding an abnomality at around {1.80}\thinspace{GHz} where the carrier frequencies of leaked $\weightTensor$ at {0.915}\thinspace{GHz} is overlapped with that of $\inputTensor$ at {0.885}\thinspace{GHz}.
When further increasing the carrier frequency of $\inputTensor$ to {3.00}\thinspace{GHz}, the normalized RMSE gradually increases to around {0.098} (approximately {4.35}-bit resolution).

\myparatight{Benchmark the input power levels to RF and LO ports}
We follow the above-mentioned {\convtwodim} operation setup to benchmark the computing accuracy over varying input power levels to the RF ($\inputTensor$) and LO ($\weightTensor$) ports.
Specifically, we vary the PSD combination of $\inputTensor$ and $\weightTensor$ respectively from {$-$116}\thinspace{dBm/Hz} to {$-$66}\thinspace{dBm/Hz} with a step of {1}\thinspace{dB}, corresponding to power levels of {$-$42}\thinspace{dBm} to {$+$8}\thinspace{dBm} over the bandwidth of {25}\thinspace{MHz}.
Similarly, the computing accuracy trend is similar for the three selected numbers of input channels of $\chanInNum \in \left\{ 64, 128, 256\right\}$, as shown in Fig.~\ref{fig:supplementary-both-power}.
Specifically for $\chanInNum=128$, an minimized normalized RMSE of around {0.043} (approximately {5.54}-bit resolution) is achieved when the input power levels to the RF and LO ports are around {-81}\thinspace{dBm/Hz} and {-81}\thinspace{dBm/Hz} (corresponding to {-7.02}\thinspace{dBm} power), respectively.
This indicates the optimal operating point for the diodes in the frequency mixer as an analog multiplier.
In {\name}, the input power level to the LO port is fixed, depending on the central radio's broadcast power, while that to the RF port is adjustable by the edge client, but is minimized to improve the energy efficiency.
For example, under a fixed LO power of {-77}\thinspace{dBm/Hz} ({-3.02}\thinspace{dBm}), we have an normalized RMSE of {0.056} (resolution bit of {5.16}) with at the RF power of {-90}\thinspace{dBm/Hz} ({-16.02}\thinspace{dBm}); when reducing that to {-110}\thinspace{dBm/Hz} ({-36.02}\thinspace{dBm}), the normalized RMSE is increased to around {0.165} (resolution bit of {3.60}).

\myparatight{Default settings for the frequency mixer}
In the implementation of {\name}, we empirically select the carrier frequency of $\inputTensor$ to be {1.00}\thinspace{GHz}; given the carrier frequency of $\weightTensor$ at {0.915}\thinspace{GHz}, that of $\outputTensor$ is set to {0.085}\thinspace{GHz} after the down-conversion.
In addition, we select the input power levels to the LO ports to be {-77}\thinspace{dBm/Hz} ({-3.02}\thinspace{dBm}).

\subsection{General-Purpose {\convtwodimbf} Benchmarks over Scalability}
\label{ssec: supplementary-experiment-scalability}

We benchmark the complex-valued {\convonedim} and {\convtwodim} layers with randomized $\inputTensor$ and $\weightTensor$ over varying numbers of input/output channels and PSDs.
Throughout this section, all the elements in $\inputTensor$ and $\weightTensor$ are i.i.d. with a uniformly distributed amplitudes from $\mathcal{U}(0, 1)$ and a uniformly distributed phase from $\mathcal{U}(-\pi, +\pi)$.
Similar to the {\supplementary}~\ref{ssec: supplementary-experiment-benchmark}, we fix the kernel size of the {\convtwodim} layer to $\kernelSizeX \times \kernelSizeY = 3 \times 3$, and the input image size to $\imageSizeX \times \imageSizeY = 16 \times 16$. As for {\convonedim} layers, we configure the kernel size to $\kernelSize = 3$, and the input image size to $\imageSize = 64$.

\myparatight{Impact over varying number of input/output channels}
We first examine the computing accuracy over a varying number of input and output channels, with a fixed input PSD to the RF at {-77}\thinspace{dBm/Hz}.
In particular, we adjust the number of input and output channels equally, setting $\chanInNum = \chanOutNum \in \left\{ 2^0, 2^1, \dots, 2^{10}\right\}$.
A detailed comparison regarding the real/imaginary components, amplitude, and phase between the $\outputTensor$ by full precision and $\outputTensorEst$ by {\name} is shown in Fig.~\ref{fig:supplementary-scalability-scatter}.
Specifically, Fig.~\ref{fig:supplementary-scalability-scatter}(a) shows the results of {\convonedim}, while Fig.~\ref{fig:supplementary-scalability-scatter}(b) shows the results of {\convtwodim}.
Over a different number of input/output channels, the normalized RMSE is stable between {0.031--0.076} (approximately {4.72--6.01}-bit resolution) for {\convonedim}, and {0.048--0.065} (approximately {4.94--5.49}-bit resolution) for {\convtwodim}.
This illustrates {\name}'s scalability over the number of input/output channels, which is crucial for supporting increasingly large CNN models.

\myparatight{Impact over varying input PSD}
For each number of input/output channels, we further examine the computing accuracy over varying input PSD to the RF port.
This varying PSD potentially impacts the SNR of the output waveform, and thus the computing accuracy. 
Fig.~\ref{fig:supplementary-scalability-energy}(a) shows the results of {\convonedim}, and Fig.~\ref{fig:supplementary-scalability-energy}(b) shows the results of {\convtwodim}.
For instance, with $\chanInNum=\chanOutIdx=64$, the normalized RMSE is {0.326} (equivalent to {2.62}-bit resolution) at a PSD of {-140}\thinspace{dBm/Hz} for {\convtwodim}. The computing accuracy improves at a rate of approximately {6.02}\thinspace{dB/bit} as the input PSD decreases, until reaching a PSD of {-115}\thinspace{dBm/Hz}, where Gaussian noise becomes the dominant factor affecting accuracy. Beyond this point, increasing the PSD leads to a saturation in computing accuracy, maintaining an RMSE of approximately {0.057--0.063} (corresponding to a resolution bit of {5}) within the PSD range of {-115---100}\thinspace{dBm/Hz}, limited by the imperfections of the frequency mixer.
This trend is similar for the other numbers of input/output channels ranging from $2^0$ to $2^{10}$.
Comparing the results of {\convonedim} and {\convtwodim} layers, the {\convonedim} layers fluctuate more.

\myparatight{Energy efficiency with different computing accuracy requirements}
The energy efficiency per MAC of {\name} is discussed in {\autorefmethod}.
Specifically, the energy per MAC is given by $\energyMAC = \energyMACEnc + \energyMACDAC + \energyMACADC + \energyMACDec$, where $\energyMACDAC$ by the DACs is determined by the PSD that impacts the computing accuracy, while the other three terms, $\energyMACEnc$ by digital encoding, $\energyMACADC$ by ADC, and $\energyMACDec$ by digital decoding are constant, and can be scaled down by the computing scale, $\chanInNum$ and $\chanOutNum$.
Fig.~\ref{fig:supplementary-scalability-scale-1d} and Fig.~\ref{fig:supplementary-scalability-scale} respectively showcase the energy scaling of {\convonedim} and {\convtwodim} layers.
Specifically for {\convtwodim}, we first show the energy term $\energyMACEnc$, $\energyMACADC$ and $\energyMACDec$ over varying number of input/output channels in Fig.~\ref{fig:supplementary-scalability-scale}(a), whose energy per MAC is reduced from {44.44}\thinspace{fJ/MAC}, {5.56}\thinspace{fJ/MAC} and {44.44}\thinspace{fJ/MAC} to {0.10}\thinspace{fJ/MAC}, {0.01}\thinspace{fJ/MAC} and {0.04}\thinspace{fJ/MAC} when $\chanInNum = \chanOutNum$ is increased from $2^0$ to $2^{10}$.
Then, we show the minimum $\energyMACDAC$ to achieve a computing accuracy of 3/4/5 bits and the corresponding total energy per MAC $\energyMAC$ in Fig.~\ref{fig:supplementary-scalability-energy}(b).
Overall, the term $\energyMACDAC$ increases slightly with the number of input/output channels.
Specifically for $\chanInNum=\chanOutNum=2^{10}$, the minimum $\energyMACDAC$ to achieve a computing accuracy of 3/4/5 bits is {1.40}\thinspace{aJ/MAC}, {8.83}\thinspace{aJ/MAC} and {0.22}\thinspace{fJ/MAC}, respectively; these correspond to the total energy per MAC $\energyMAC$ of {0.15}\thinspace{fJ/MAC}, {0.16}\thinspace{fJ/MAC}, and {0.37}\thinspace{fJ/MAC}.
The total energy per MAC $\energyMAC$ can be scaled up by increasing the computing scale, and will converge to the term $\energyMACDAC$. 
To conclude, the energy efficiency of {\name} is approximately {1--3} orders of magnitude higher than the state-of-the-art digital computing with {100}\thinspace{fJ/MAC}.

\subsection{Complex-Valued CNN-based Classification and Generative Models}
\label{ssec: supplementary-experiment-model}

The CNN models, both for classification and the generation, are trained digitally, and then directly evaluated in {\name}'s analog computing on the software-defined radio testbed.
Note that all parameters in these models are complex-valued, in order to showcase {\name}'s capability to support complex MAC operations.
A summary of the six models is presented in Fig.~\ref{fig:supplementary-model-architecture} and Table~\ref{tab:supplementary-model-spec}.

\myparatight{Complex-valued layers}
There are mainly two types of layers used in the CNN models in {\name}: (\emph{i}) linear layers, including the {\convtwodim} layers and the fully-connected layers (see {\supplementary}~\ref{ssec: supplementary-theory-convolution-basics}), and (\emph{ii}) non-linear layers, such as BatchNorm, ReLU/LeakyReLU, max pooling and etc.
These non-linear layers are implemented by extending their real-valued definitions: two parallel layers are defined respectively for the real and imaginary components. For example, there are two sets of mean and variance parameters in the BatchNorm layer, one for the real component and the other for the imaginary component.

\myparatight{6-layer CNN models for DeepSig classification}
DeepSig~\cite{o2018over} is a wireless signal dataset for modulation classification. Originally, there are {28} classes of modulation types, but we pick and merge them into {10} representative classes, which are
\begin{itemize}
    \item \emph{OOK}: on-off keying.
    \item \emph{ASK}: 4/8-amplitude shift keying.
    \item \emph{BPSK}: binary phase shift keying.
    \item \emph{QPSK}: quadrature phase shift keying.
    \item \emph{PSK}: 8/16/32/64/128-phase shift keying.
    \item \emph{QAM}: 16/32/64/128/256-quadrature amplitude modulation.
    \item \emph{AM-SSB}: amplitude modulation -- single sideband, either with carrier or carrier suppressed.
    \item \emph{AM-DSB}: amplitude modulation -- double sideband, either with carrier or carrier suppressed.
    \item \emph{FM}: frequency modulation.
    \item \emph{GMSK}: gaussian minimum-shift keying.
\end{itemize}
All of these wireless signals are represented as complex-valued {1,024}-point vectors. The dataset provides wireless signals with SNRs varying from {-20}\thinspace{dB} to {+30}\thinspace{dB}. We train the classification model on the signals with SNRs of {10--30}\thinspace{dB}, and test the model using the {30}\thinspace{dB} signals only.
We consider one-shot, three-shot, and five-shot classifications, which infer 1/3/5 different wireless signals of the same modulation, and vote for the final prediction. This multi-shot setup is based on the fact that a wireless signal can exist for a longer time than the vector length provided in the dataset.
Given the one-dimensional input and the complex-valued nature of the signals, we build a 6-layer-CNN model with five {\convonedim} layers (namely, \textsf{Conv1} to \textsf{Conv5}) and one FC layer (namely, \textsf{FC1}), where the kernel size of all {\convonedim} layers is $3$, and the number of channels are $1 \rightarrow 256 \rightarrow 256 \rightarrow 512 \rightarrow 512 \rightarrow 512$; each {\convonedim} layer is followed by a BatchNorm layer, a ReLU layer, and a max pooling layers with a kernel size of $4$ that shrinks the signals' dimension by $4$. The final FC layer maps the feature into a 10-point complex-valued vector for the 10-classification task.
To summarize, this 6-layer CNN model has a total of {2.18}\thinspace{M} parameters and {368.1}\thinspace{M} MACs.

\myparatight{9-layer CNN models for SVHN and CIFAR-10 classification}
The SVHN~\cite{netzer2011reading} dataset is composed of $32 \times 32$ RGB images of house numbers (``0'' to ``9''), and the CIFAR-10 dataset is composed of $32 \times 32$ RGB images of ten classes of objects (plane, car, bird, cat, deer, dog, frog, horse, ship, and truck).
Given the same dimensions of these two datasets, we tailor the standard VGG11 architecture~\cite{simonyan2014very} to build a 9-layer CNN model architecture that is applied to the image classification tasks on both datasets.
Specifically, the 9-layer CNN model consists of eight {\convtwodim} layers (namely, \textsf{Conv1} to \textsf{Conv8}) and one FC layer (namely, \textsf{FC1}), where the kernel size of all {\convtwodim} layers, and the number of channels are $3 \rightarrow 64 \rightarrow 128 \rightarrow 256 \rightarrow 256 \rightarrow 512 \rightarrow 512 \rightarrow 512 \rightarrow 512$; each {\convtwodim} layer is followed by a 2D BatchNorm layer, a ReLU layer, and \textsf{Conv1/2/4/6/8} layers have additional 2D max pooling layers with a kernel size of $2 \times 2$ and a stride of $2$.
Over these eight {\convtwodim} layers, the input image with a size of $3 \times 32 \times 32$ is converted into a feature of $512 \times 1 \times 1$, which is then flattened as a one-dimensional vector into the \textsf{FC1} layer that projects it into a {10}-point complex-valued vector.
Finally, we extract the real components of the output vector as the model's prediction of the input image being ten classes.
To summarize, this 9-layer CNN model architecture has a total of {9.22}\thinspace{M} parameters and {611.1}\thinspace{M} MACs.

\myparatight{4-layer CNN model for the MNIST and FMNIST generation}
MNIST~\cite{lecun2002gradient} dataset contains $28 \times 28$ grayscale images of handwritten digits (``0'' to ``9''), and FMNIST~\cite{xiao2017fashion} dataset contains $28 \times 28$ grayscale images of ten classes of fashion clothes (coat, boot, shirt, sneaker, pullover, bag, scandal, dress, trousers, and T-shirt).
Similarly, we employ the same model architecture to generate fake images for both datasets following InfoGAN~\cite{chen2016infogan}.
The generator is a complex-valued 4-layer CNN model with one FC layer (\textsf{FC1}) and three {\convtwodim} layers (\textsf{Conv1} to \textsf{Conv3}).
It takes a 74-point real-valued vector as input, including a 62-point random seed vector, a 10-point one-hot vector to represent the image's being ten classes, and a 2-point latent code vector continuously ranging from 0 to 1 to represent the image's hidden features, such as the rotation and thickness of the digits or fashion products.
This input is first projected by the \textsf{FC1} layer into a {6,272}-point complex-valued vector, which is then reshaped into a feature of $128 \times 7 \times 7$.
Then, this feature is processed by three {\convtwodim} layers with a kernel size of $3 \times 3$ with a channel number of $128 \rightarrow 512$, $128 \rightarrow 256$, and $64 \rightarrow 1$, respectively.
The \textsf{Conv1} and \textsf{Conv2} layers are followed by PixelShuffle, 2D BatchNorm, and LeakyReLU with a negative slope of $0.2$. Note that the PixelShuffle layers perform 2D up-sampling by $2 \times 2$ while reducing the channel number by $4$.
The \textsf{Conv3} layer is followed by one 2D BatchNorm, and performs Tanh as the output activation function on the real component only, which converts the output into the range of $[-1, +1]$.
In summary, this 4-layer CNN generative model has a total of {1.35}\thinspace{M} parameters and {350.5}\thinspace{M} MACs.
The discriminator is a real-valued 5-layer CNN model with four {\convtwodim} layers and one FC layer, which (\emph{i}) distinguishes the real images from the fake images by the generator, (\emph{ii}) classifies the input images into ten classes, and (\emph{iii}) estimates the latent code vector of the input images.
The four {\convtwodim} layers have a kernel size of $3 \times 3$ and $2 \times 2$ stride, and the channel numbers are $1 \rightarrow 16 \rightarrow 32 \rightarrow 64 \rightarrow 128$; they are all followed by 2D BatchNorm, LeakyReLU with a negative slope of $0.2$, and Dropout with a probability of {25}\%.
The last FC layer projects the feature into a 14-point real-valued vector composed of a 2-point vector indicating the input image's being real or fake, a 10-point vector indicating the input image's being one of ten classes, and a 2-point vector indicating the input image's latent code vector.

\myparatight{7-layer CNN model for CelebA generation}
We further employ a complex-valued 7-layer CNN model to generate fake images of RGB human faces with $128 \times 128$ based on the CelebA~\cite{liu2015deep} dataset.
The input is still a 74-point real-valued vector, where the 10-point one-hot vector for the label is randomly generated.
The first \textsf{FC1} layer projects the input into a {16,384}-point complex-valued vector, which is reshaped into a feature of $1024 \times 8 \times 8$.
Then, this feature is processed by six {\convtwodim} layers with a kernel size of $3 \times 3$ and a channel number of $1024 \rightarrow 2048$, $512 \rightarrow 1024$, $256 \rightarrow 512$, $128 \rightarrow 256$, $64 \rightarrow 32$ and $32 \rightarrow 3$, respectively.
The \textsf{Conv1} to \textsf{Conv5} layers are followed by PixelShuffle, 2D BatchNorm and LeakyReLU with a negative slope of $0.2$, while the last \textsf{Conv6} layer is followed by one 2D BatchNorm and performs Tanh as the output activation function on the real component only.
This 7-layer-CNN generator has a total of {26.4}\thinspace{M} parameters and {5,195}\thinspace{M} MACs.
The discriminator is a real-valued 6-layer CNN model with five {\convtwodim} layers and one FC layer.
The five {\convtwodim} layers have a kernel size of $3 \times 3$ and $2 \times 2$ stride, and the channel numbers are $3 \rightarrow 64 \rightarrow 128 \rightarrow 256 \rightarrow 512 \rightarrow 1024$; they are all followed by 2D BatchNorm, LeakyReLU with a negative slope of $0.2$, and Dropout with probability of {25}\%. The last FC layer projects the feature into a 14-point real-valued vector with the same composition as the discriminator for MNIST and FMNIST.

\myparatight{Model training}
The two classification models for SVHN and CIFAR-10 are trained for {200} epochs using SGD~\cite{bottou2010large} with a learning rate of $10^{-2}$ and a weight decay of $10^{-2}$. Over the epochs, we select the model with the best classification accuracy as the final model for evaluation in {\name}'s analog computing.
The three generative models for SVHN, CIFAR-10 and CelebA are trained with the InfoGAN loss~\cite{chen2016infogan} for {200} epochs using Adam~\cite{kingma2014adam} with a learning rate of $2 \times 10^{-4}$ until the loss converges.
These training processes are all performed on a single NVIDIA A100 GPU.

\subsection{Waveform Amplitude Selection against PAPR}
\label{ssec: supplementary-experiment-waveform-amplitude}

As discussed in {\supplementary}~\ref{ssec: supplememtary-waveform-papr}, balancing the trade-off between PAPR and SQNR, a.k.a. selecting the backoff coefficient $\backoff$, is crucial for {\name}'s computing accuracy.
Hereby, we benchmark the computing accuracy over varying $\backoff$ to select a proper $\backoff$ for each {\convonedim}, {\convtwodim}, and FC layer in the employed CNN models.
At the same time, we identify the key layers that are particularly sensitive to computing errors and replace them from {\name}'s analog computing to full precision digital computing if the errors exceed acceptable levels.

\myparatight{Definition of the single-layer and end-to-end inference RMSEs}
We define the single-layer and end-to-end inference RMSEs as the metrics to benchmark the computing accuracy on these CNN models.
When examining the $\layerIdx$-th layer, we feed the input $\inputTensor$ of this layer by inferring the previous $\layerIdx-1$ layers in full precision. Then, we infer the $\layerIdx$-th layer by {\name} ($\outputTensorEst$) and full precision ($\outputTensor$). The RMSE following equation~\eqref{eq: supplementary-experiment-benchmark-rmse} between $\outputTensorEst$ and $\outputTensor$ is defined as the single-layer inference RMSE.
By comparing to the randomized single layer in {\supplementary}~\ref{ssec: supplementary-experiment-scalability}, this metric quantifies how the real $\inputTensor$ and $\weightTensor$ to the $\layerIdx$-th layer impact the computing accuracy.
Then, we infer the resting layers after the $\layerIdx$-th layer in full precision with $\outputTensorEst$ and $\outputTensor$, respectively. The RMSE between the final outputs of the whole CNN model is defined as the end-to-end inference RMSE.
This end-to-end inference RMSE provides a fair comparison between the sensitivity of different layers.
In the rest of this section, all the benchmark measurements are based on three signals/images in the dataset, and the optimized $\backoff$ is then applied to the whole testing dataset.

\myparatight{Benchmark the backoff coefficient for 6-layer CNN model for the DeepSig classification}
We first benchmark the backoff coefficient $\backoff$ using the end-to-end inference RMSE for the 6-layer CNN model for the DeepSig classification.
Fig.~\ref{fig:supplementary-wave-amp-deepsig} shows the end-to-end RMSE as a function of the backoff coefficient $\backoff$ for each layer in the 6-layer CNN model for the DeepSig classification.
The \textsf{Conv1} layer encounters the highest end-to-end RMSE ($>0.30$) over all the choices of $\backoff$. This is because of the highest PAPR of its input wireless signals, and because $\chanInNum=1$, where the Zadoff-Chu phases cannot apply (see {\supplementary}~\ref{ssec: supplememtary-waveform-papr}). Therefore, we infer the \textsf{Conv1} layer in digital full precision. Fortunately, this \textsf{Conv1} layer only occupies {0.85\%} of the MACs, so its digital inference does not largely change the overall energy efficiency (or energy per MAC) of the whole model's inference.
For the following \textsf{Conv2} layer, the RMSE first drops because of its better robustness to overcome the saturation issue, and then increases because of a too low SQNR. In observation of this, we select the $\backoff$ as {34}\thinspace{dB} to achieve the minimum RMSE.
Then, the RMSE gradually decreases over the layers. This is because a larger $\chanInNum$ enables the Zadoff-Chu phases to reduce the PAPR more effectively, and a smaller $\backoff$ is good enough to avoid the saturation. 

\myparatight{Benchmark the backoff coefficient for 9-layer CNN model for the SVHN/CIFAR-10 classification}
We then benchmark the 9-layer CNN model for the SVHN/CIFAR-10 classification.
The end-to-end inference RMSE on the SVHN classification is shown in Fig.~\ref{fig:supplementary-wave-amp-svhn}. Most layers (\textsf{Conv1/2/3/4}) have decreasing end-to-end inference RMSE with the increasing $\backoff$, which is expected as a larger $\backoff$ tolerates a higher PAPR without the saturation issue. Also, the optimized RMSE decreases over the layers. This is because the later layers have a larger $\chanInNum$, and thus the Zadoff-Chu phases can alleviate the PAPR issue more effectively; to avoid saturation, a smaller $\backoff$ is required, which leads to a better SQNR and thus a higher computing accuracy.
A similar trend can be found in the model for CIFAR-10 classification, as shown in Fig.~\ref{fig:supplementary-wave-amp-cifar10}.

\myparatight{Benchmark the backoff coefficient for the generative models}
We further benchmark the backoff coefficient $\backoff$ for the three generative models.
Similarly, most layers have a decreasing end-to-end inference RMSE with the increasing $\backoff$, and then increase.
Different from the three models for classifications, the generative models have a decreasing number of channels over the layers, which leads to a reverse trend of the end-to-end inference RMSE over the layers.
For example, in the 7-layer CNN model for CelebA generation, the end-to-end inference RMSE of \textsf{FC1} is {0.054}, while that of the \textsf{Conv6} is increased to {0.200}. To maintain a good model inference accuracy, we perform \textsf{Conv6} in digital full precision.

\myparatight{The single-layer RMSE measurements}
After picking the optimal $\backoff$ for all the layers in all the models, we showcase the comparison between the full precision output and {\name} output, as well as the single-layer RMSE of the six CNN models in Fig.~\ref{fig:supplementary-scatter-DL-deepsig} (DeepSig classification), Fig.~\ref{fig:supplementary-scatter-DL-svhn} (SVHN classification), Fig.~\ref{fig:supplementary-scatter-DL-cifar10} (CIFAR-10 classification), Fig.~\ref{fig:supplementary-scatter-DL-mnist} (MNIST generation), Fig.~\ref{fig:supplementary-scatter-DL-fmnist} (FMNIST generation), and Fig.~\ref{fig:supplementary-scatter-DL-celeba} (CelebA generation).
Note that the digital inferred layers (\textsf{Conv1} in the DeepSig classification, and \textsf{Conv6} in the CelebA generation) are skipped.
Generally, the distribution of these DL outputs follows specific patterns, unlike that in the scalability measurement in {\supplementary}~\ref{ssec: supplementary-experiment-scalability} that follows a Gaussian distribution. Therefore, the corresponding single-layer RMSE can be different.

\subsection{Power Spectral Density Trade-off on the CNN Models for Classification}
\label{ssec: supplementary-experiment-energy-classification}

In {\name}, the PSD of $\inputTensor$ determines the SNR, and therefore the computing accuracy. This further affects the overall energy efficiency of {\name} as a trade-off.
We provide additional results by simulations and experiments across a larger PSD or energy efficiency range for the six CNN models.

\myparatight{Single/multi-shot classification accuracy over PSD for DeepSig}
Due to the continuity of the wireless signals in the DeepSig dataset, we can infer multiple signals of the same modulation type to improve the classification accuracy. 
Hereby, we consider the one-shot, three-shot, and five-shot classifications, which infer 1/3/5 different wireless signals of the same modulation, and vote for the final prediction, as shown in Fig.~\ref{fig:supplementary-classification-energy-deepsig-1shot}, Fig.~\ref{fig:supplementary-classification-energy-deepsig-3shot}, and Fig.~\ref{fig:supplementary-classification-energy-deepsig-5shot}, respectively.
Specifically for the one-shot classification in Fig.~\ref{fig:supplementary-classification-energy-deepsig-1shot}(a), the simulated and experimental results by {\name} exhibit a consistent trend as the PSD varies. At very low PSD values (e.g., below approximately {-125}\thinspace{dBm/Hz}), the classification accuracy remains below {20\%}, indicating that {\name} is heavily corrupted by the Gaussian noises.
As the PSD increases to {-112}\thinspace{dBm/Hz} ({3.34}\thinspace{fJ/MAC}), the experimental classification accuracy is increased to {87.6\%}; The classification accuracy is further increased up to {88.7\%} at the PSD of {-107}\thinspace{dBm/Hz} (({4.48}\thinspace{fJ/MAC})), only {4.2\%} gap from the full-precision accuracy of {92.9\%}, as shown in Fig.~\ref{fig:supplementary-classification-energy-deepsig-1shot}(b).
A similar trend can be observed for the three-shot and five-shot classifications, as shown in Fig.~\ref{fig:supplementary-classification-energy-deepsig-3shot} and Fig.~\ref{fig:supplementary-classification-energy-deepsig-5shot}, respectively. The experimental classification accuracy at {-107}\thinspace{dBm/Hz} (({4.48}\thinspace{fJ/MAC})) is improved to {92.8\%} and {95.2\%} by inferring more signals of the same modulation type.

\myparatight{Classification accuracy over PSD for SVHN and CIFAR-10}
We then evaluate the classification accuracy over varying PSD for the 9-layer CNN model for the SVHN and CIFAR-10 classification.
Fig.~\ref{fig:supplementary-classification-energy-svhn} shows the results for the SVHN classification, where the experimental classification accuracy is around {93.3\%} at the PSD of {-117}\thinspace{dBm/Hz} ({0.76}\thinspace{fJ/MAC}), only {1.4\%} gap from the full-precision accuracy of {94.7\%}.
A similar trend can be observed for the CIFAR-10 classification, as shown in Fig.~\ref{fig:supplementary-classification-energy-cifar10}, where the experimental classification accuracy is around {90.0\%} at the PSD of {-112}\thinspace{dBm/Hz} ({0.88}\thinspace{fJ/MAC}), only {0.9\%} gap from the full-precision accuracy of {90.9\%}.

\subsection{Power Spectral Density Trade-off on the CNN Models for Generation}
\label{ssec: supplementary-experiment-energy-generation}

We evaluate the trade-off between the input PSD and the FID of the generated images by the three generative models on the MNIST, FMNIST and CelebA datasets, respectively.

\myparatight{Frechet inception distance (FID) as the metric for generative images}
FID is a widely used metric to evaluate the quality of generated images, which maps a set of real and fake images into a feature space by a pre-trained Inception network, and then measures the distance between the distribution of the real and fake images.
Generally, a smaller FID indicates a higher fidelity of the generated images, and thus a better performance of the generative model.

\myparatight{PSD-FID trade-off on MNIST and FMNIST datasets}
We evaluate the trade-off between the PSD (or energy efficiency) and the FID of the generated images by the four-layer generative models on the MNIST and FMNIST datasets. 
These FIDs are calculated across {100} generated images with {10} images per label.
Fig.~\ref{fig:supplementary-generative-energy-mnist}(a) shows the trade-off on the MNIST dataset, where the FID is around {81.33} when PSD is {-102}\thinspace{dBm/Hz} ({2.89}\thinspace{fJ/MAC}), while the FID is increased to around {108.29} when PSD is reduced to around {-117}\thinspace{dBm/Hz} ({1.11}\thinspace{fJ/MAC}). We further show examples of digit 1--9 in Fig.~\ref{fig:supplementary-generative-energy-mnist}(b) at the PSD of {-132-- -102}\thinspace{dBm/Hz} at the step of {5}\thinspace{dB}, corresponding to the FIDs of {81.33--295.04}.
Similarly, Fig.~\ref{fig:supplementary-generative-energy-fmnist}(a) shows the trade-off on the FMNIST dataset, where the FID is around {170.97} when PSD is {-102}\thinspace{dBm/Hz} ({2.89}\thinspace{fJ/MAC}), while the FID is increased to around {188.15} when PSD is reduced to around {-117}\thinspace{dBm/Hz} ({1.11}\thinspace{fJ/MAC}). We further show examples of the ten classes of fashion clothes in Fig.~\ref{fig:supplementary-generative-energy-fmnist}(b) at the PSD of {-132 -- -102}\thinspace{dBm/Hz} at the step of {5}\thinspace{dB}, corresponding to the FIDs of {170.97--346.73}.

\myparatight{PSD-FID trade-off on CelebA dataset}
We repeat the same experiments for the 7-layer CNN model on the CelebA dataset to generate fake human faces of {128}$\times${128} with RGB.
Each reported FID is calculated across {100} generated images.
Due to the higher complexity of the model and dataset, the digital FID at full precision is {112.55}. In comparison, the FID is around {183.22} when PSD is {-102}\thinspace{dBm/Hz} ({2.77}\thinspace{fJ/MAC}), while the FID is increased to around {206.83} when PSD is reduced to around {-112}\thinspace{dBm/Hz} ({1.17}\thinspace{fJ/MAC}). We further show examples of the generated human faces in Fig.~\ref{fig:supplementary-generative-energy-celeba}(b) at the PSD of {-132 -- -102}\thinspace{dBm/Hz} at the step of {5}\thinspace{dB}, corresponding to the FIDs of {183.22 -- 393.33}.

%% file: tex/Supplementary_figure.tex
\begin{figure*}[!t]
    \centering
    \includegraphics[width=0.95\columnwidth]{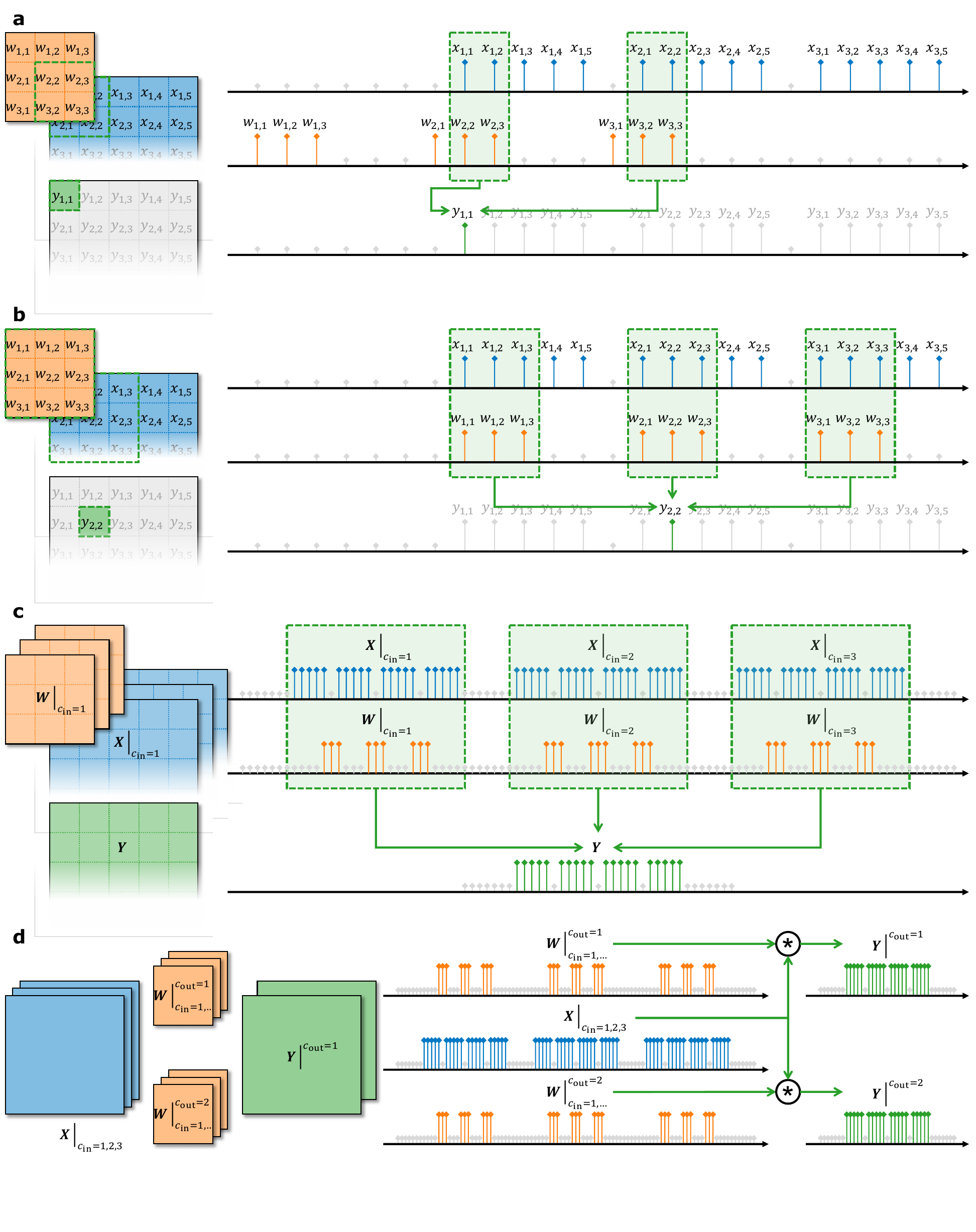}
    \caption{
    \textbf{The tone mapping algorithm of {\namebf} to achieve {\convtwodimbf} layers.}
    \textbf{a--b}, For the single-input/output-channel case, the tone series of $\inputTensor$ is the flattened 2D image, and that of $\weightTensor$ is aligned in rows; the correlation result of each alignment generates one output element of $\outputTensor$.
    \textbf{c}, For the multiple-input-channel case, the tone series of $\inputTensor$ and $\weightTensor$ are aligned for each channel, and are therefore summed up after the correlation.
    \textbf{d}, For the multiple-output-channel case, the inference is performed independently for each output channel using the same tone series of $\inputTensor$.
    }
    \label{fig:conv-algorithm}
\end{figure*}

\begin{figure*}[!t]
    \centering
    \includegraphics[width=0.8\columnwidth]{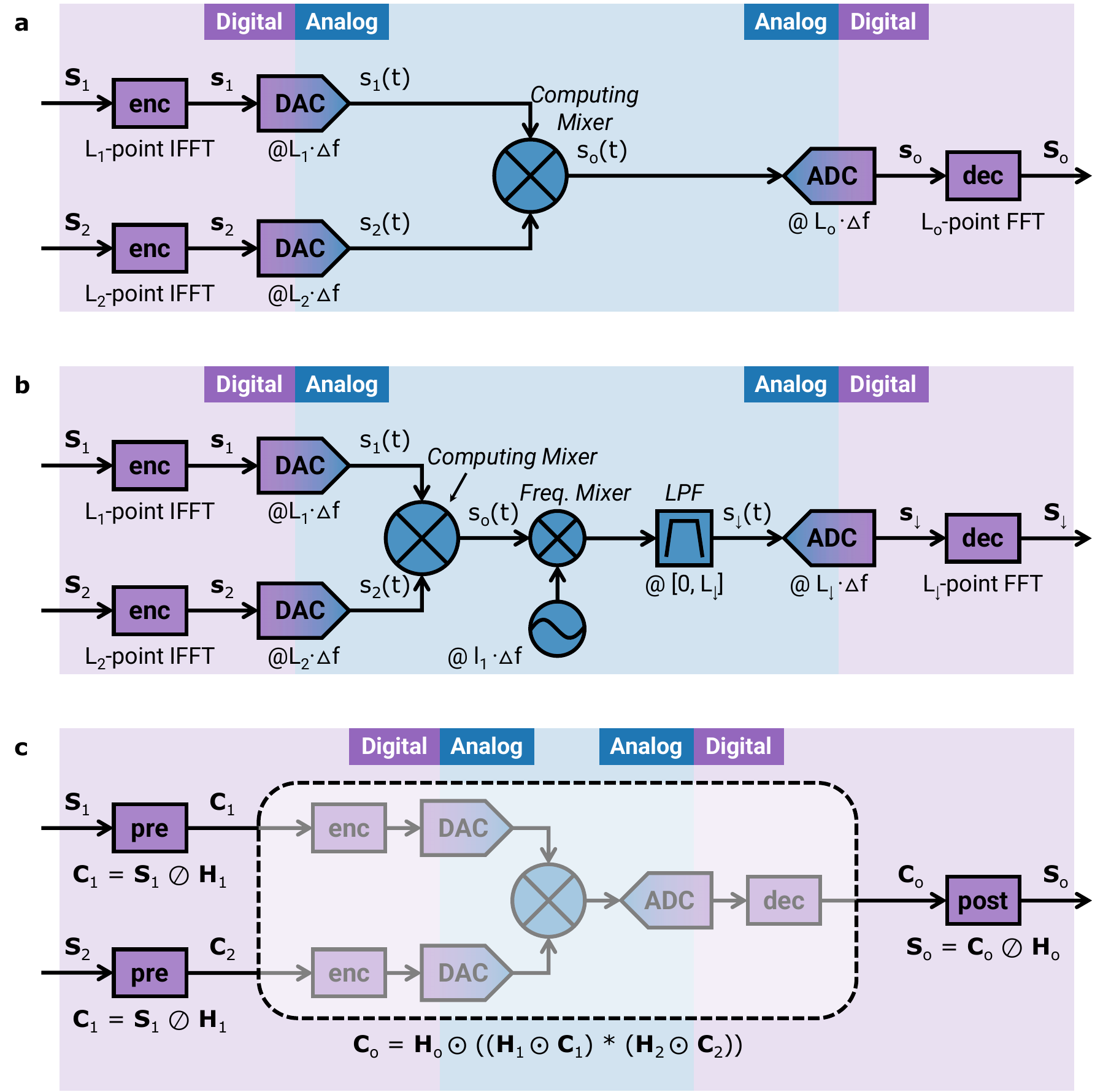}
    \caption{
    \textbf{The hardware diagrams for the frequency mixer-driven convolutions.}
    \textbf{a}, The hardware diagram for the full convolutions from the input vectors $\convInputVec$ and $\convWeightVec$ to the output $\convOutputVec$.
    \textbf{b}, The hardware diagram for the partial convolution, whose output $\convOutputSubVec$ is a segment of $\convOutputVec$.
    \textbf{c}, The hardware diagram with the pre- and post-channel calibration, which compensates the physical channel (wired or wireless) distortion $\convCalibInputVec$, $\convCalibWeightVec$, and $\convCalibOutputVec$.
    }
    \label{fig:supplementary-convolution-diagram}
\end{figure*}

\begin{figure*}[!t]
    \centering
    \includegraphics[width=0.7\columnwidth]{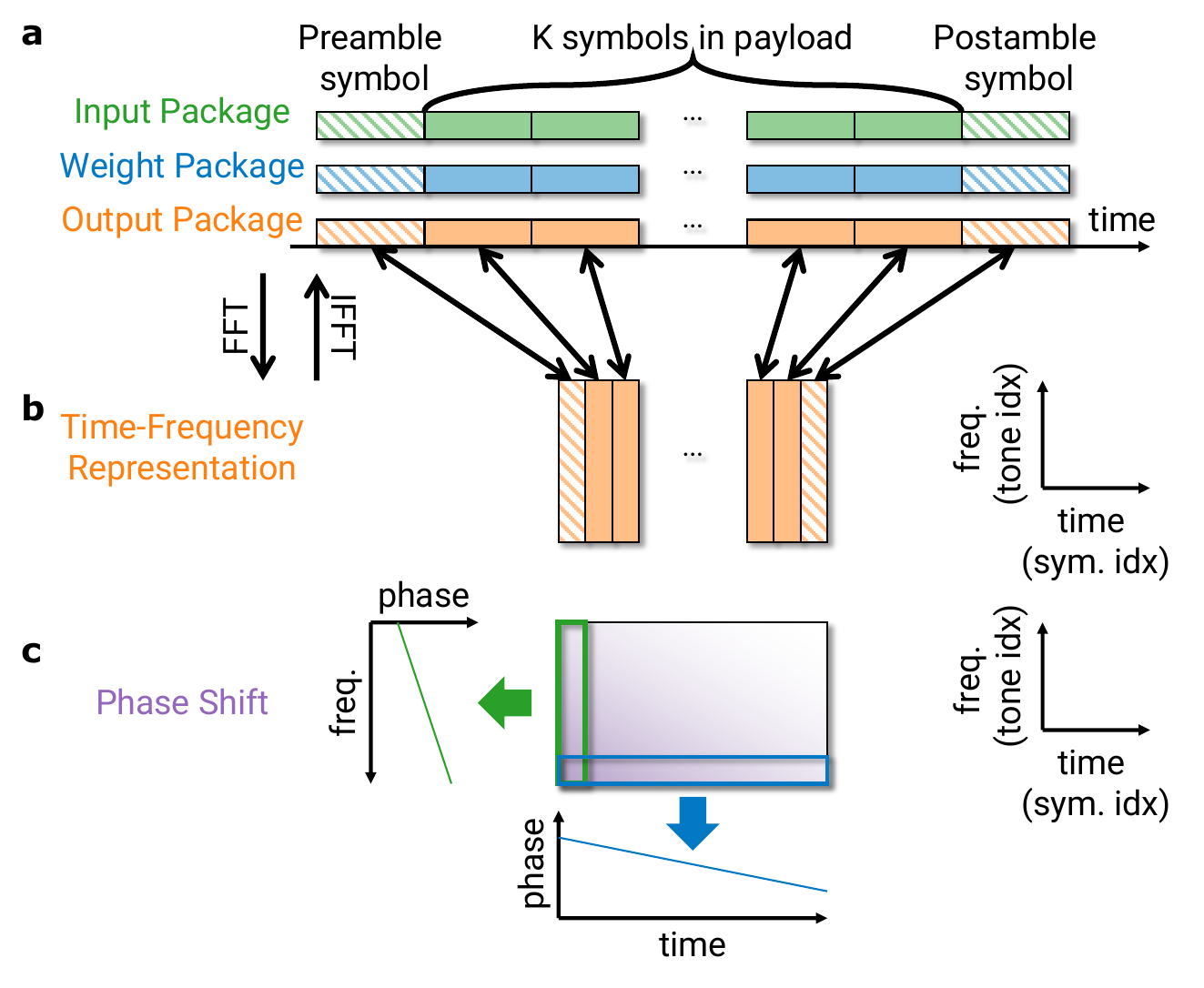}
    \caption{
    \textbf{During the wireless transmission, the phase offset on the output packet.}
    \textbf{a}, The packet structure of {\name} with preamble/postamble symbols before and after the payload in time.
    \textbf{b}, After conducting the FFT for each symbol, the time-domain output packet becomes a time-frequency representation.
    \textbf{c}, The phase offset on the output packet in the time-frequency representation is linear over time and frequency.
    }
    \label{fig:supplementary-phase-offset}
\end{figure*}

\begin{figure*}[!t]
    \centering
    \includegraphics[width=0.9\columnwidth]{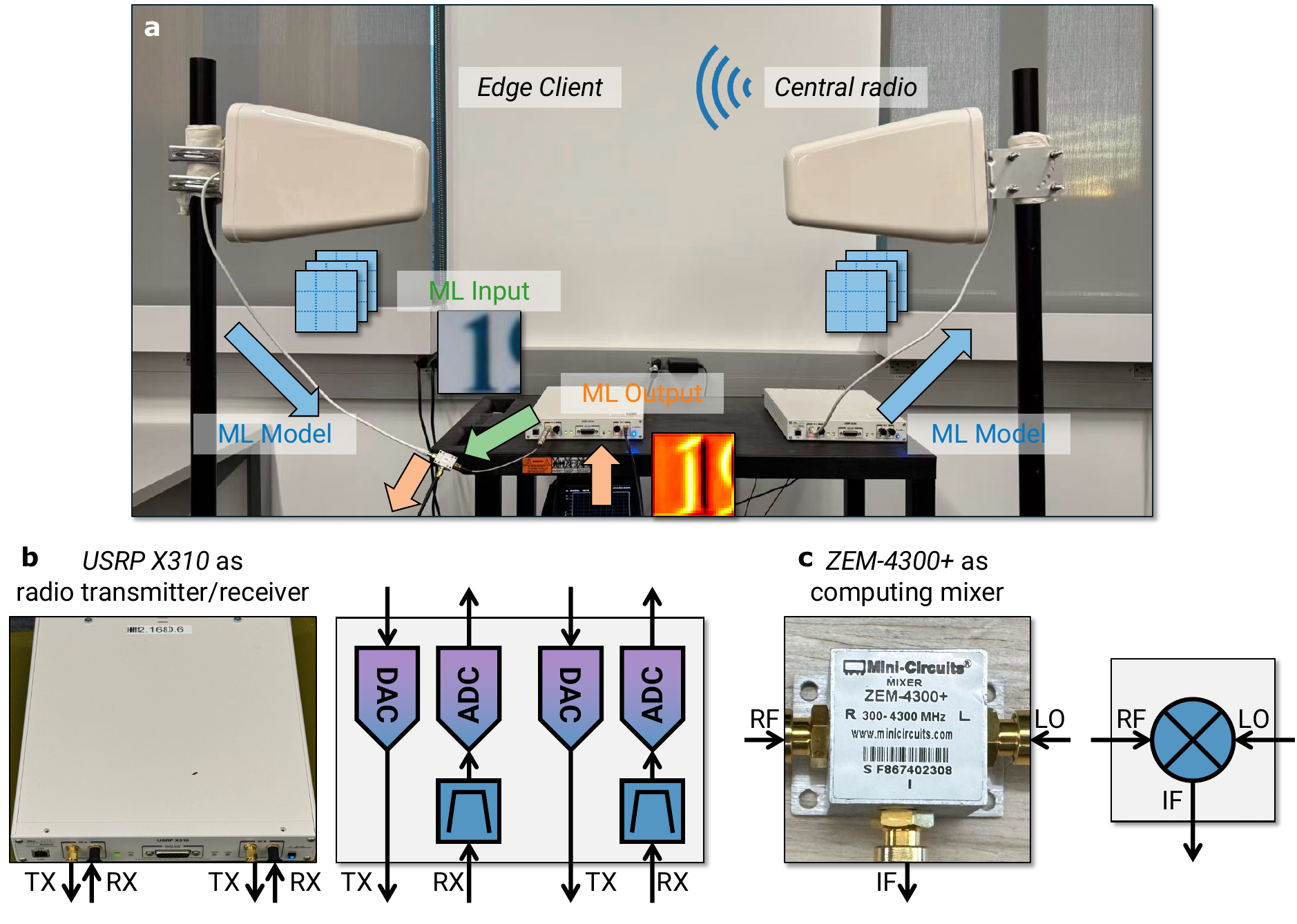}
    \caption{
    \textbf{The experiment setup on the software-defined radio testbed.}
    \textbf{a}, In {\name}'s implementation with a central radio broadcasting ML model ($\weightTensor$) and an edge client performing local ML inference from ML input $\inputTensor$ to the ML output $\outputTensor$ using a frequency mixer.
    \textbf{b}, The hardware diagram of a USRP X310, which is equivalent to a DAC as a transmitter, and a low-pass filter followed by an ADC as a receiver.
    \textbf{c}, The employed ZEM-4300+ as the computing mixer, including an LO port receiving wireless $\weightTensor$, an RF port receiving wired $\inputTensor$, and an IF port sending $\outputTensor$.
    }
    \label{fig:supplementary-experiment-setup}
\end{figure*}

\begin{figure*}[!t]
    \centering
    \includegraphics[width=1.0\columnwidth]{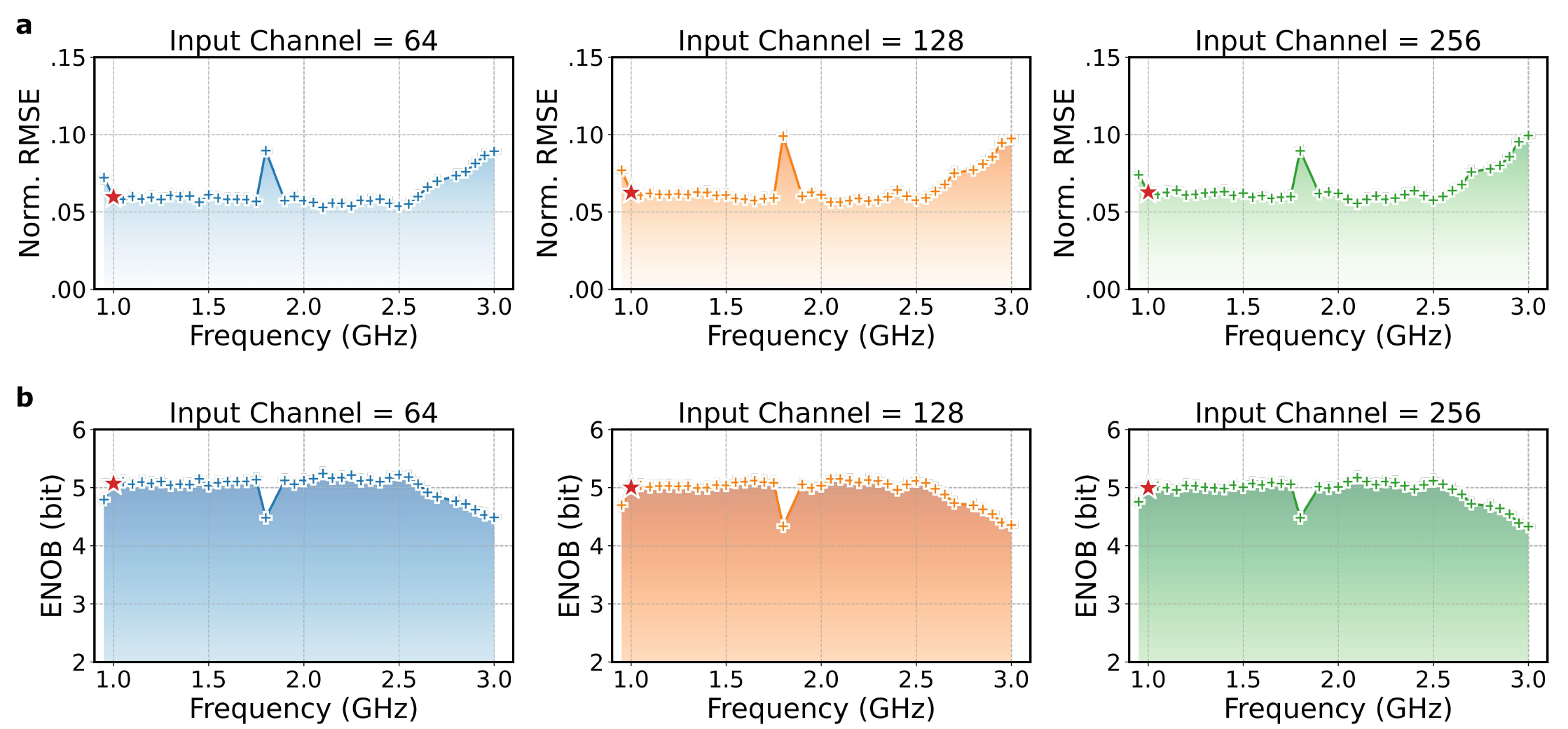}
    \caption{
    \textbf{The computing accuracy measurement over different carrier frequencies of $\inputTensor$ with that of $\weightTensor$ fixed at {0.915}\thinspace{GHz}, where we select {1.0}\thinspace{GHz} (red star) in {\namebf}'s implementation.}
    \textbf{a}, The normalized RMSE with the number of input channels $\chanInNum \in \left\{ 64, 128, 256\right\}$.
    \textbf{b}, The computing accuracy in ENOB with the number of input channels $\chanInNum \in \left\{ 64, 128, 256\right\}$.
    }
    \label{fig:supplementary-mixer-carrier}
\end{figure*}

\begin{figure*}[!t]
    \centering
    \includegraphics[width=1.0\columnwidth]{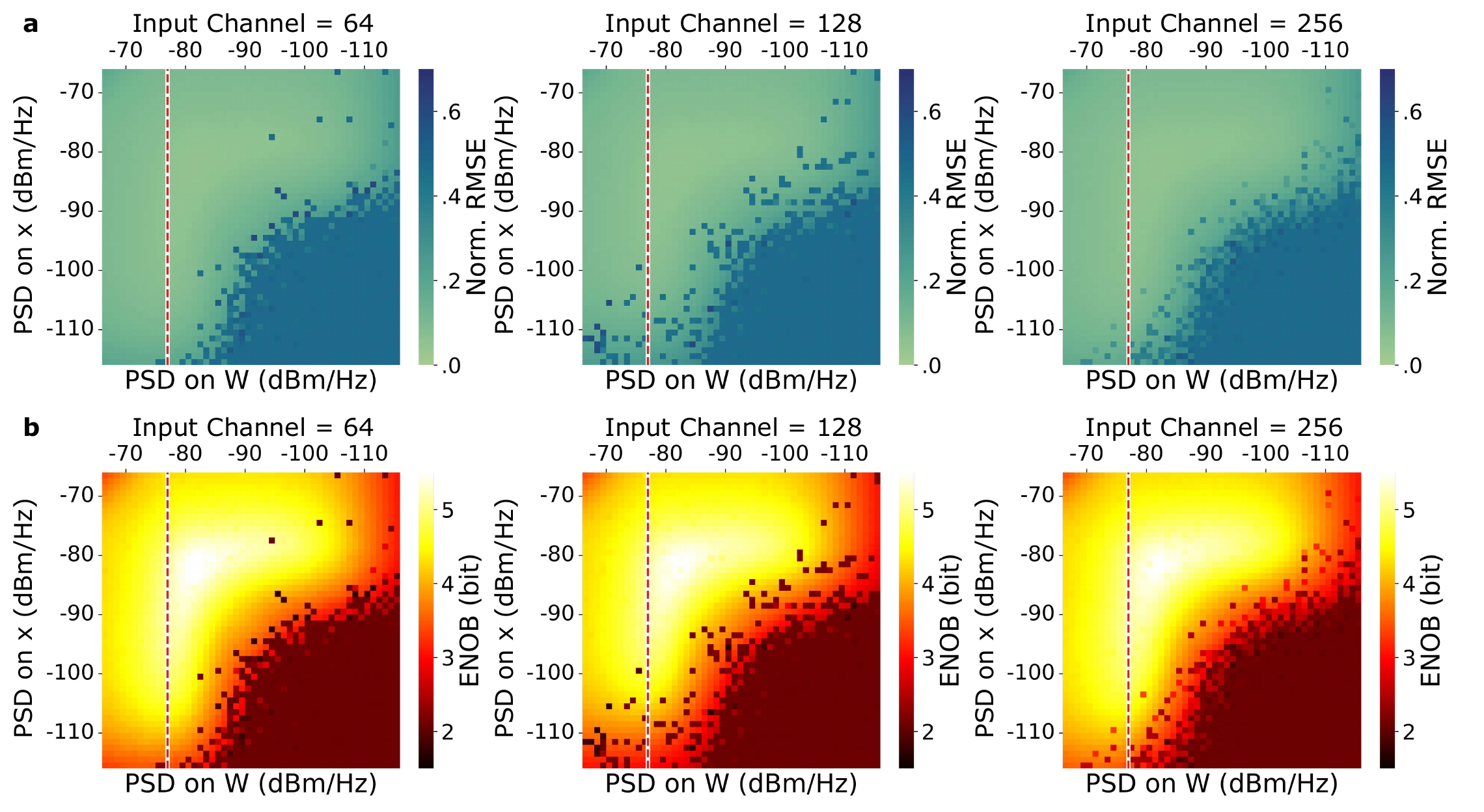}
    \caption{
    \textbf{The computing accuracy measurement over different input PSD to the LO port ($\weightTensor$) and the RF port ($\inputTensor$), where we fix the PSD of $\weightTensor$ to {-77}\thinspace{dBm/Hz} (red dashed line).}
    \textbf{a}, The normalized RMSE with the number of input channels $\chanInNum \in \left\{ 64, 128, 256\right\}$.
    \textbf{b}, The computing accuracy in ENOB with the number of input channels $\chanInNum \in \left\{ 64, 128, 256\right\}$.
    }
    \label{fig:supplementary-both-power}
\end{figure*}

\begin{figure*}[!t]
    \centering
    \includegraphics[width=0.95\columnwidth]{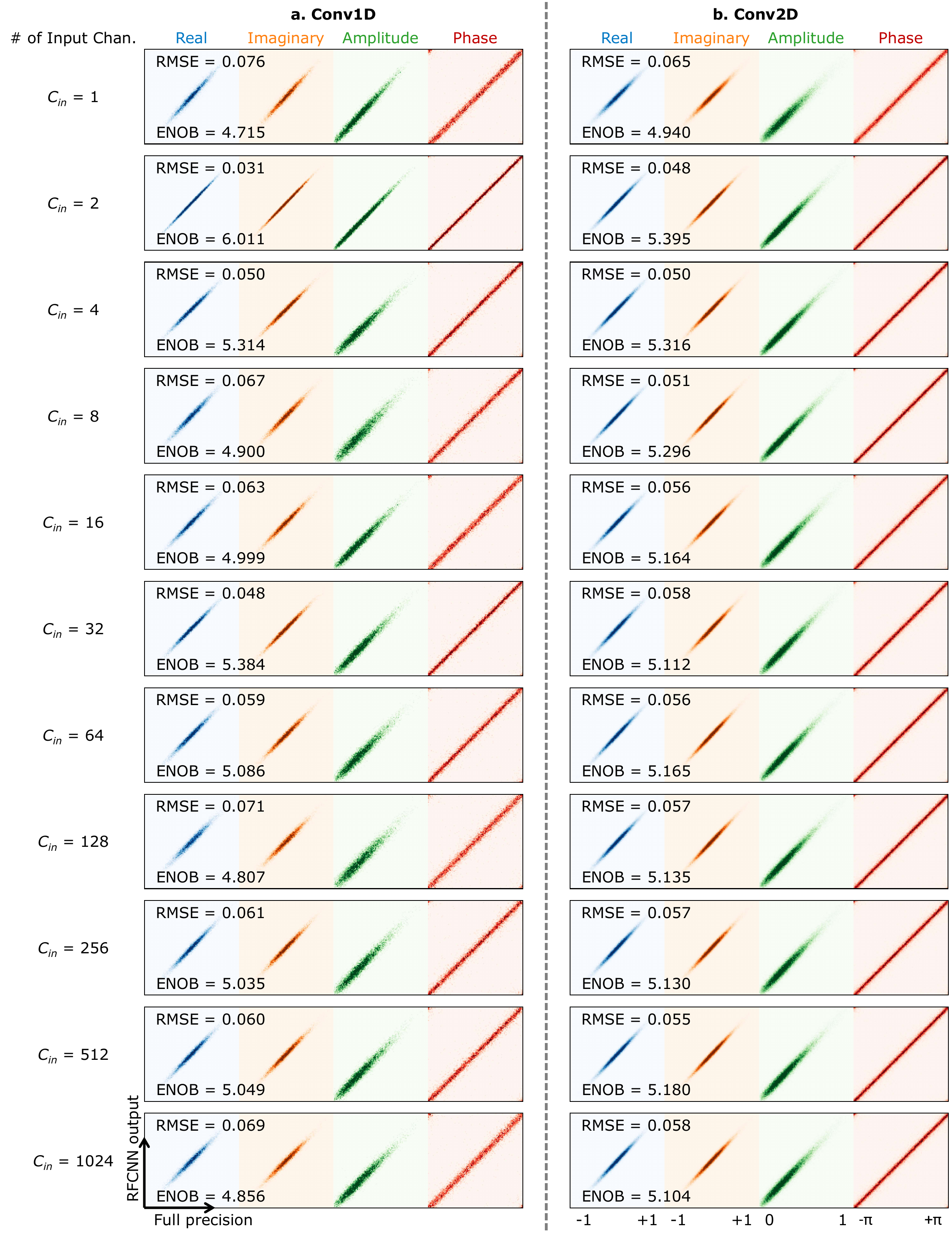}
    \caption{
    The comparison between the digital full precision output (\texttt{torch.complex64} in PyTorch) and {\name}'s output.
    Specifically, we break down the complex values into real/imaginary components, or amplitudes and phases; we further vary the number of input channels $\chanInNum \in \left\{ 1, 2, 4, 8, \dots, 1024\right\}$.
    }
    \label{fig:supplementary-scalability-scatter}
\end{figure*}

\begin{figure*}[!t]
    \centering
    \includegraphics[width=0.95\columnwidth]{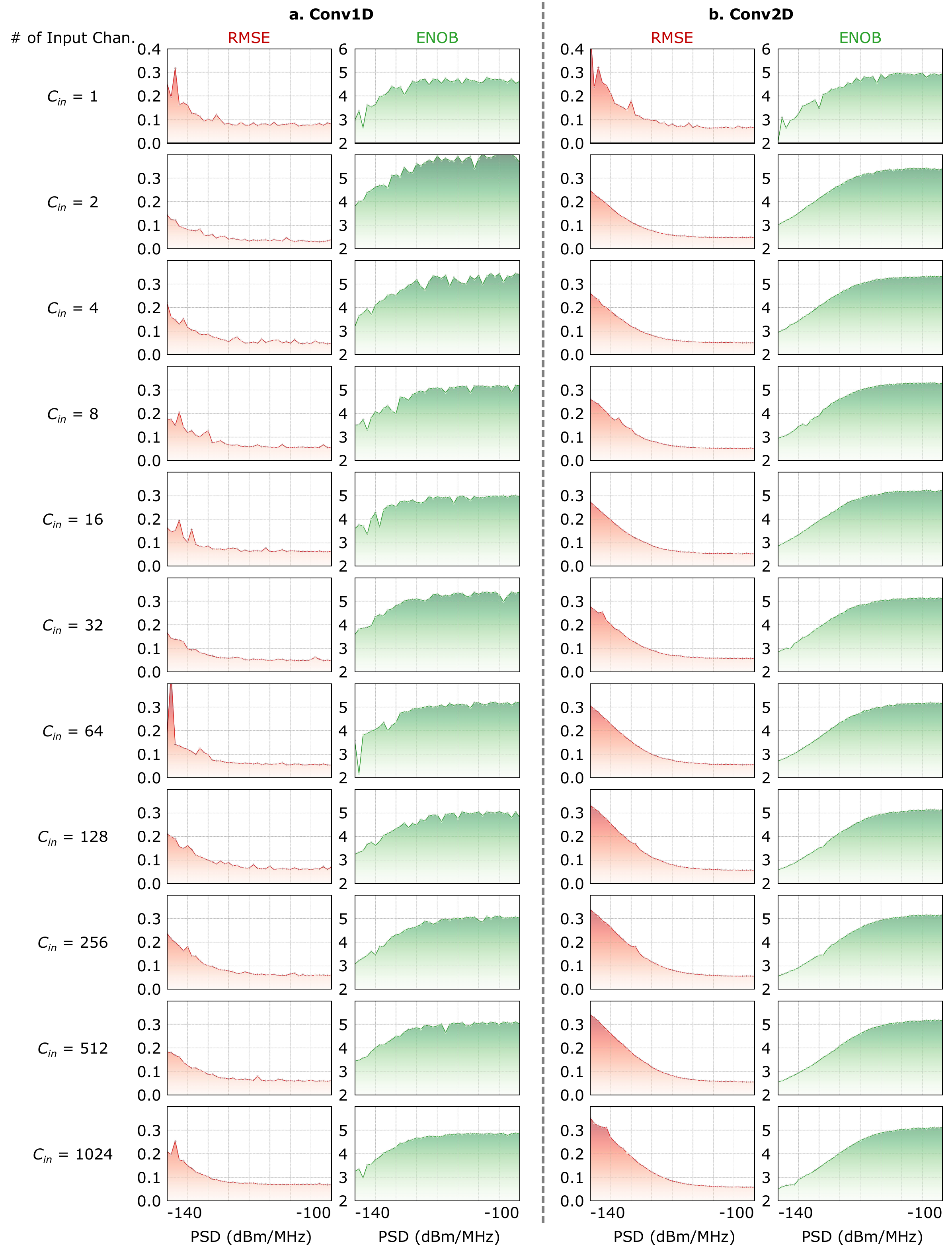}
    \caption{
    The computing accuracy measurement over the PSD of $\inputTensor$ with the number of input channels $\chanInNum \in \left\{ 1, 2, 4, 8, 16, 32, 64, 128, 256, 512, 1024\right\}$ of {\convonedim} (\textbf{a}) and {\convtwodim} (\textbf{b}).
    }
    \label{fig:supplementary-scalability-energy}
\end{figure*}

\begin{figure*}[!t]
    \centering
    \includegraphics[width=0.8\columnwidth]{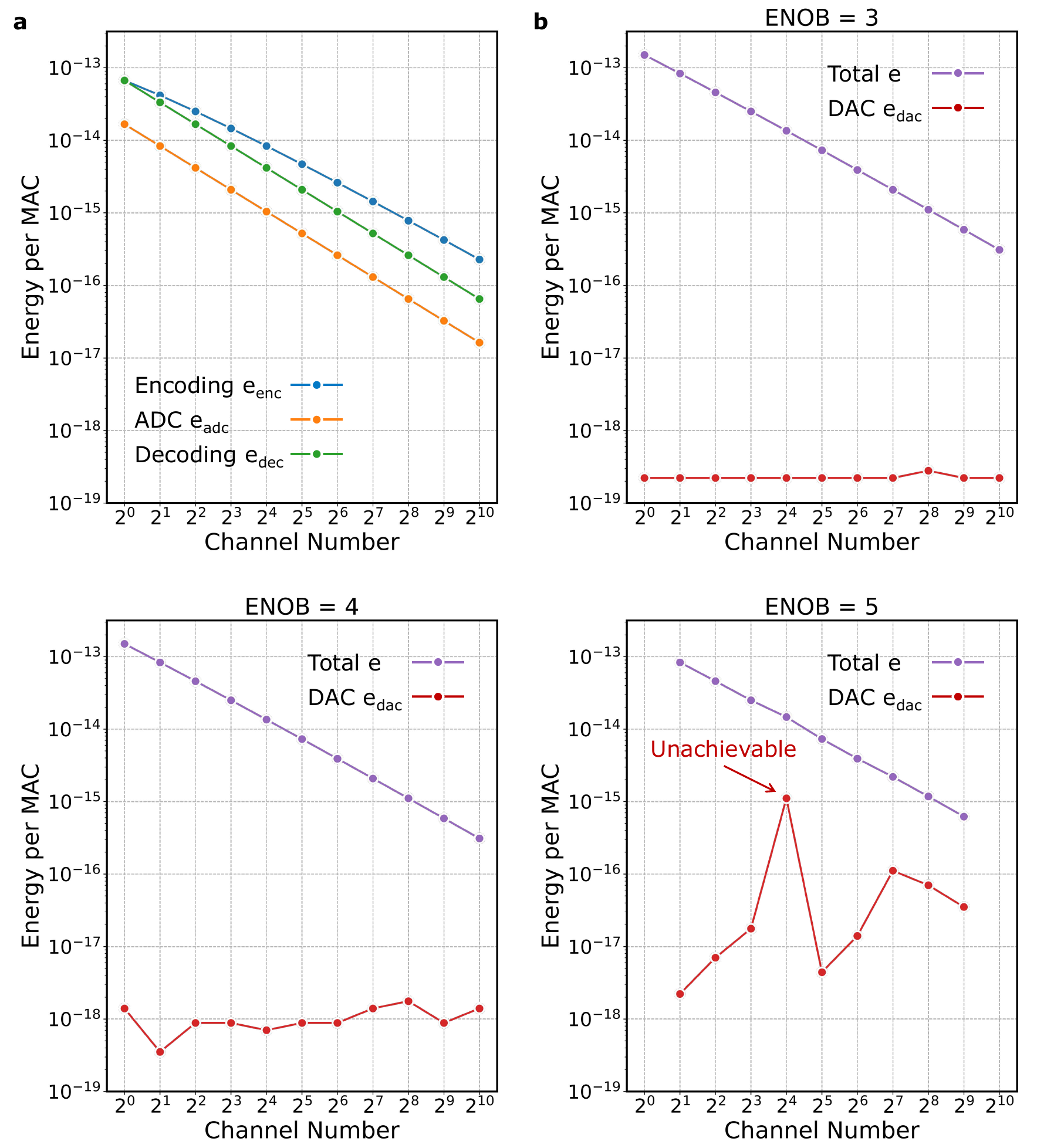}
    \caption{
    \textbf{For {\convonedimbf} layers, {\namebf}'s energy efficiency per MAC over different numbers of input/output channels.}
    \textbf{a}, The scaling of the energy efficiency terms ($\energyMACEnc$ for encoding, $\energyMACADC$ for ADC sampling, and $\energyMACDec$ for decoding) over the number of channels.
    \textbf{b}, The minimum energy per MAC by the DACs $\energyMACDAC$ to achieve 3/4/5 ENOBs, and the corresponding total energy per MAC of {\name}.
    }
    \label{fig:supplementary-scalability-scale-1d}
\end{figure*}

\begin{figure*}[!t]
    \centering
    \includegraphics[width=0.8\columnwidth]{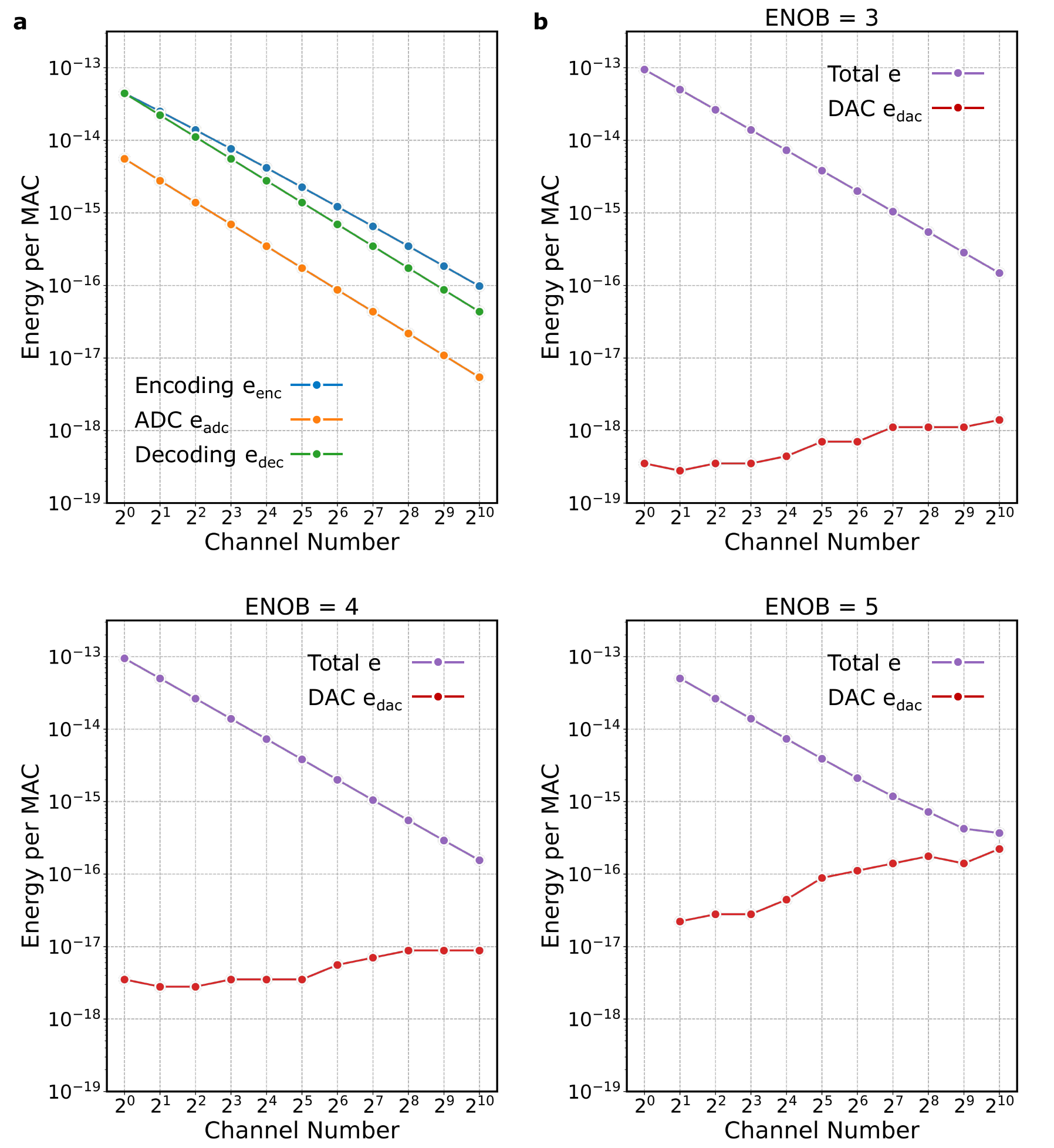}
    \caption{
    \textbf{For {\convtwodimbf} layers, {\namebf}'s energy efficiency per MAC over different numbers of input/output channels.}
    \textbf{a}, The scaling of the energy efficiency terms ($\energyMACEnc$ for encoding, $\energyMACADC$ for ADC sampling, and $\energyMACDec$ for decoding) over the number of channels.
    \textbf{b}, The minimum energy per MAC by the DACs $\energyMACDAC$ to achieve 3/4/5 ENOBs, and the corresponding total energy per MAC of {\name}.
    }
    \label{fig:supplementary-scalability-scale}
\end{figure*}

\begin{figure*}[!t]
    \centering
    \includegraphics[width=1.0\columnwidth]{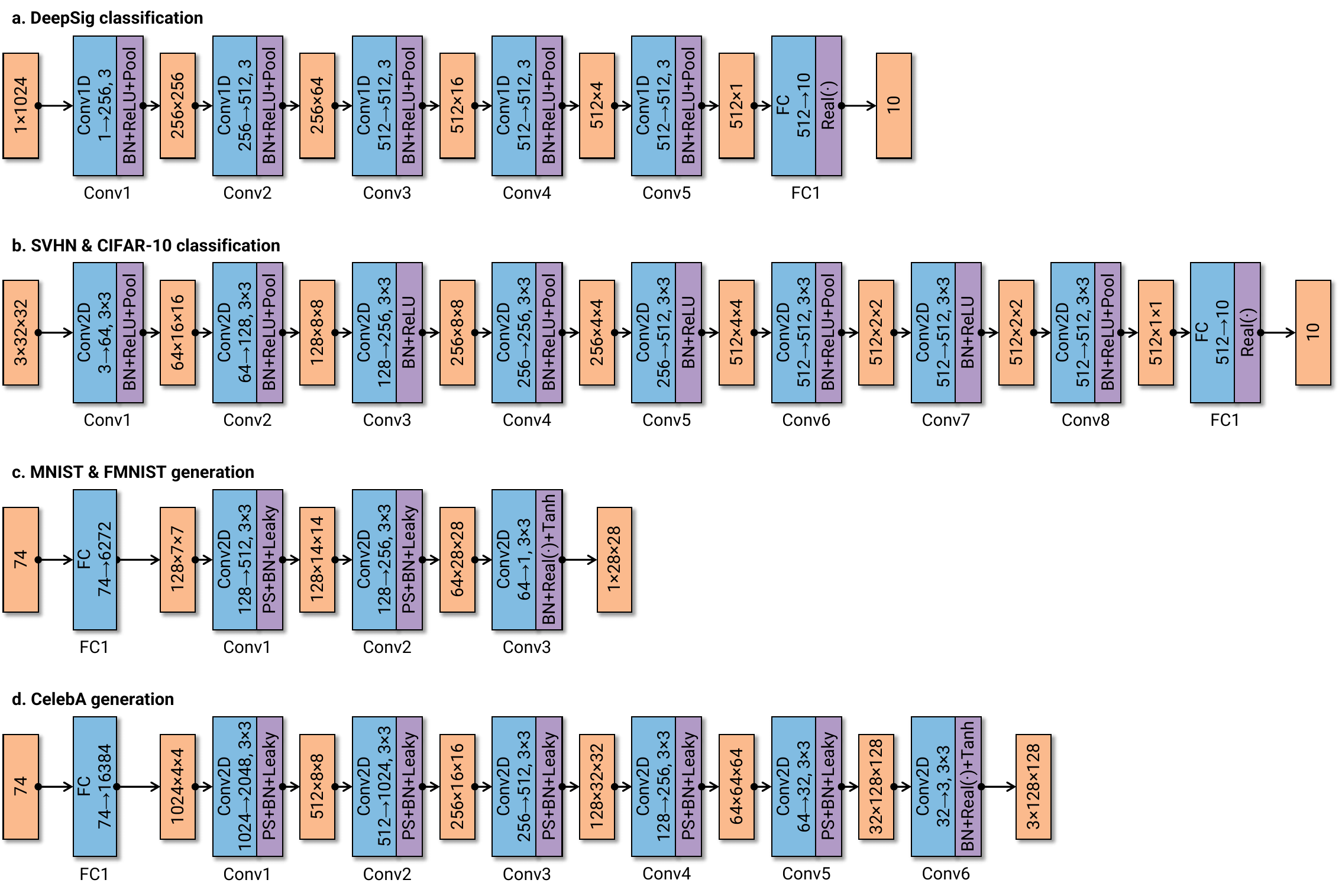}
    \caption{
    \textbf{The four CNN model architectures employed in {\namebf}'s evaluations on six ML tasks.}
    \textbf{a}, the 6-layer CNN model with {\convonedim} layers for the wireless signal modulation classification task on the DeepSig dataset.
    \textbf{b}, the 9-layer CNN model with {\convtwodim} for both the RGB digit classification task on the SVHN dataset, and the object classification task on the CIFAR-10 dataset.
    \textbf{c}, the 4-layer CNN model with {\convtwodim} trained from InfoGAN~\cite{chen2016infogan} to generate gray images of digits (MNIST) and fashion clothes (FMNIST).
    \textbf{d}, the 7-layer CNN model with {\convtwodim} trained from InfoGAN to generate RGB images of human faces (CelebA).
    }
    \label{fig:supplementary-model-architecture}
\end{figure*}

\begin{figure*}[!t]
    \centering
    \includegraphics[width=0.9\columnwidth]{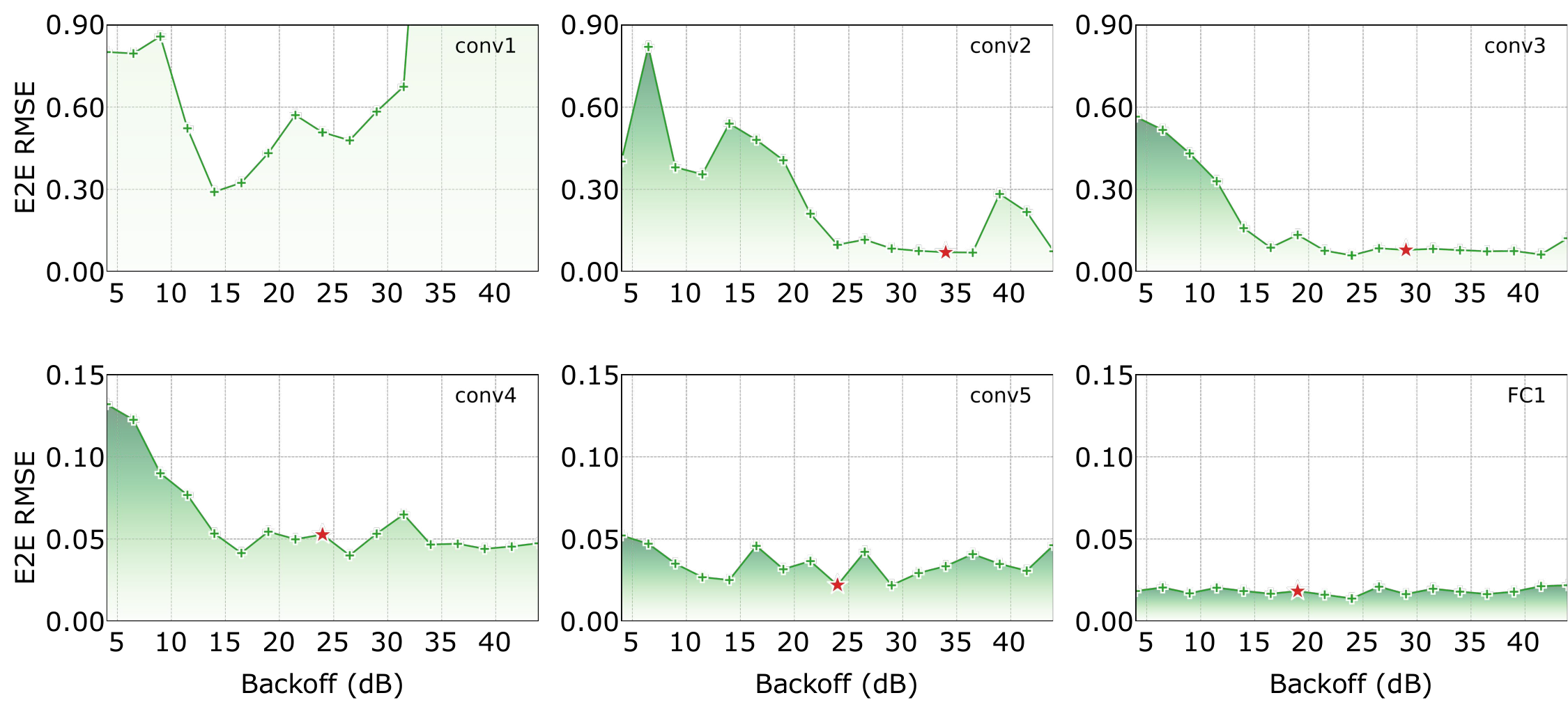}
    \caption{
    The backoff coefficient $\backoff$ measurement per layer for the 6-layer CNN model for DeepSig classification, where the red stars refer to the selected $\backoff$. Note that the end-to-end RMSE in \textsf{Conv1} is too large, and is inferred in digital full precision.
    }
    \label{fig:supplementary-wave-amp-deepsig}
\end{figure*}

\begin{figure*}[!t]
    \centering
    \includegraphics[width=0.9\columnwidth]{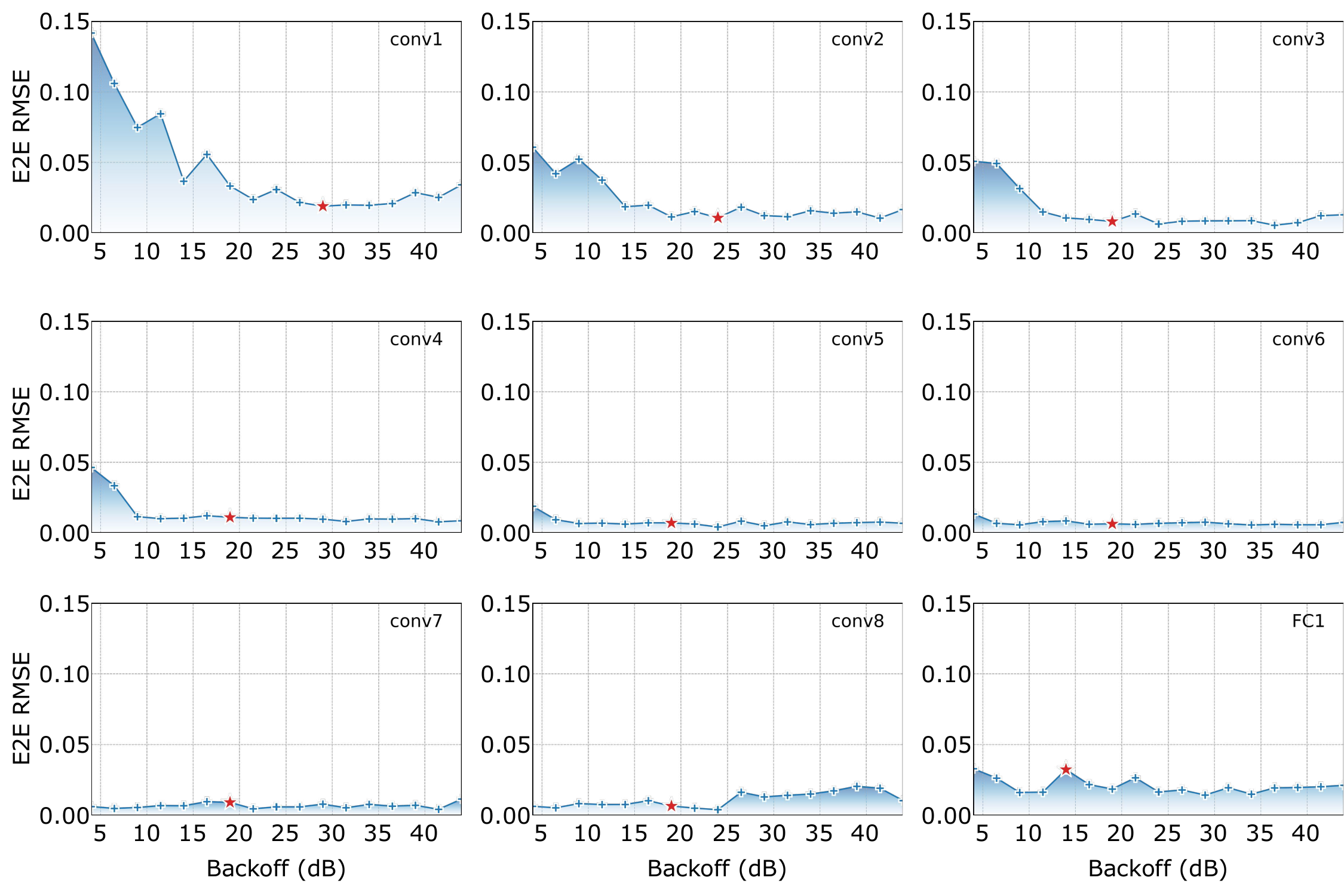}
    \caption{
    The backoff coefficient $\backoff$ measurement per layer for the 9-layer CNN model for SVHN classification, where the red stars refer to the selected $\backoff$.
    }
    \label{fig:supplementary-wave-amp-svhn}
\end{figure*}

\begin{figure*}[!t]
    \centering
    \includegraphics[width=0.9\columnwidth]{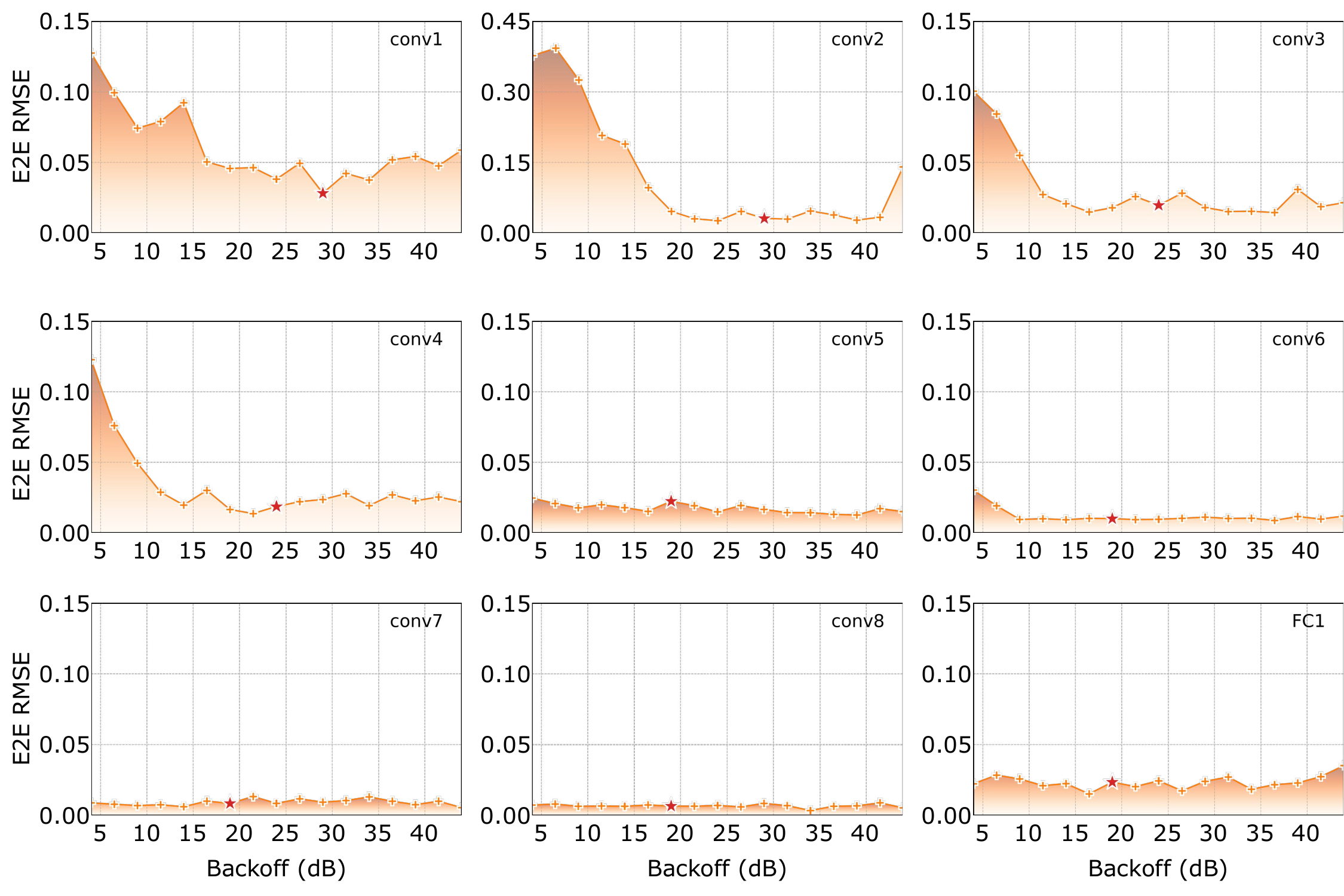}
    \caption{
    The backoff coefficient $\backoff$ measurement per layer for the 9-layer CNN model for CIFAR-10 classification, where the red stars refer to the selected $\backoff$.
    }
    \label{fig:supplementary-wave-amp-cifar10}
\end{figure*}

\begin{figure*}[!t]
    \centering
    \includegraphics[width=0.9\columnwidth]{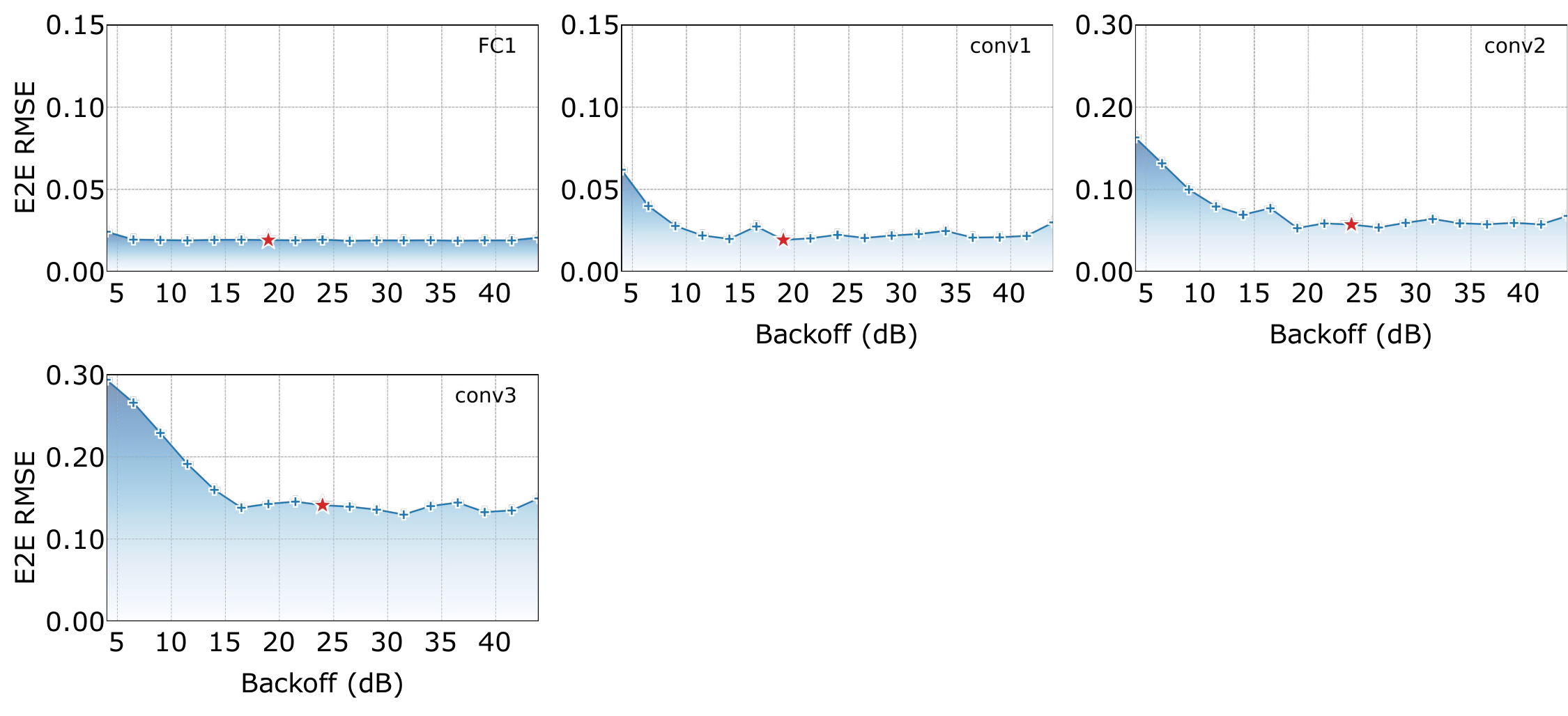}
    \caption{
    The backoff coefficient $\backoff$ measurement per layer for the 4-layer CNN model for MNIST generation, where the red stars refer to the selected $\backoff$.
    }
    \label{fig:supplementary-wave-amp-mnist}
\end{figure*}

\begin{figure*}[!t]
    \centering
    \includegraphics[width=0.9\columnwidth]{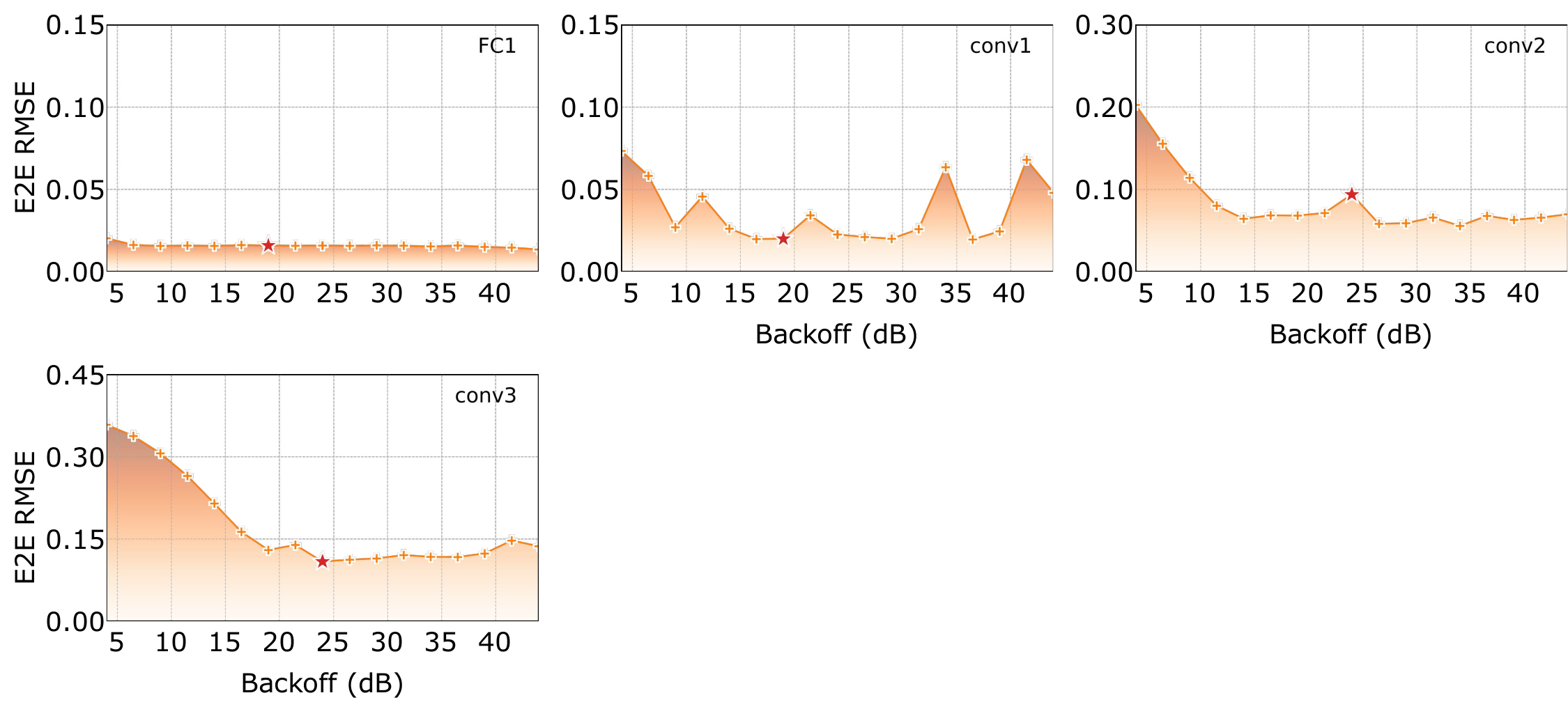}
    \caption{
    The backoff coefficient $\backoff$ measurement per layer for the 4-layer CNN model for FMNIST generation, where the red stars refer to the selected $\backoff$.
    }
    \label{fig:supplementary-wave-amp-fmnist}
\end{figure*}

\begin{figure*}[!t]
    \centering
    \includegraphics[width=0.9\columnwidth]{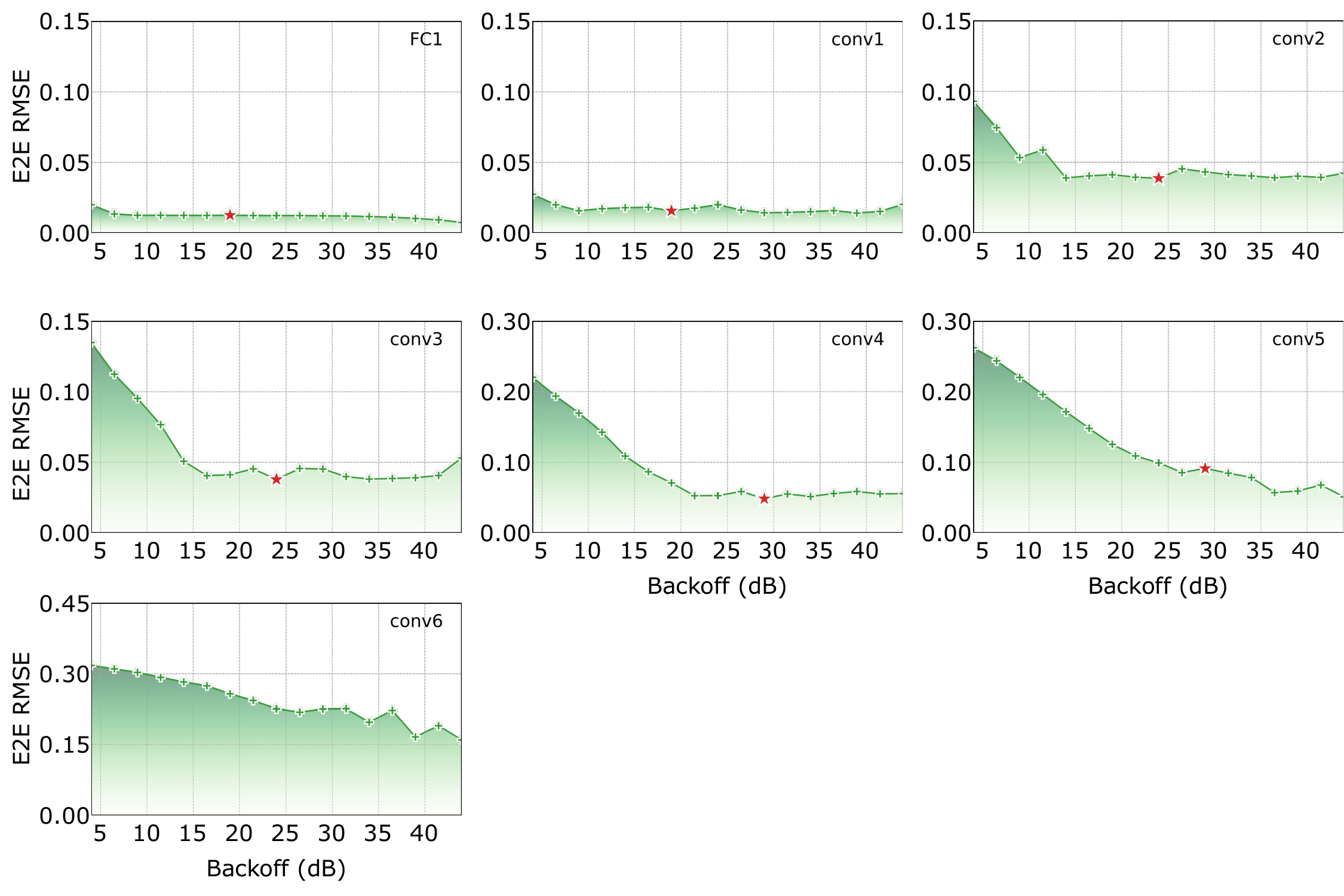}
    \caption{
    The backoff coefficient $\backoff$ measurement per layer for the 7-layer CNN model for CelebA generation, where the red stars refer to the selected $\backoff$. Note that the end-to-end RMSE in \textsf{Conv3} is too large, and is inferred in digital full precision.
    }
    \label{fig:supplementary-wave-amp-celeba}
\end{figure*}

\begin{figure*}[!t]
    \centering
    \includegraphics[width=0.6\columnwidth]{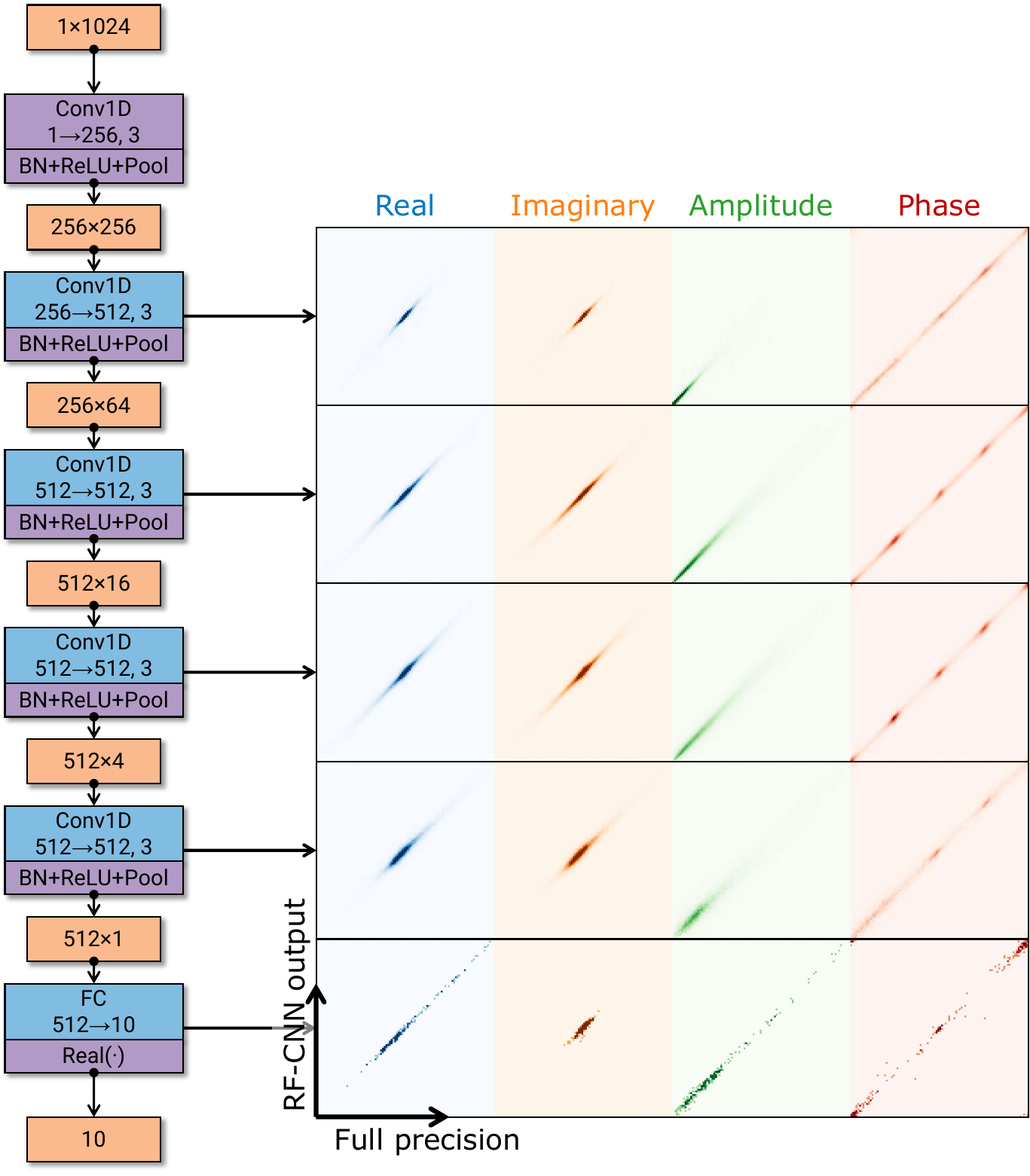}
    \caption{The per-layer comparison between the full precision output and {\name} output regarding real/imaginary/amplitude/angle on the 6-layer CNN model for the DeepSig classification.}
    \label{fig:supplementary-scatter-DL-deepsig}
\end{figure*}

\begin{figure*}[!t]
    \centering
    \includegraphics[width=0.6\columnwidth]{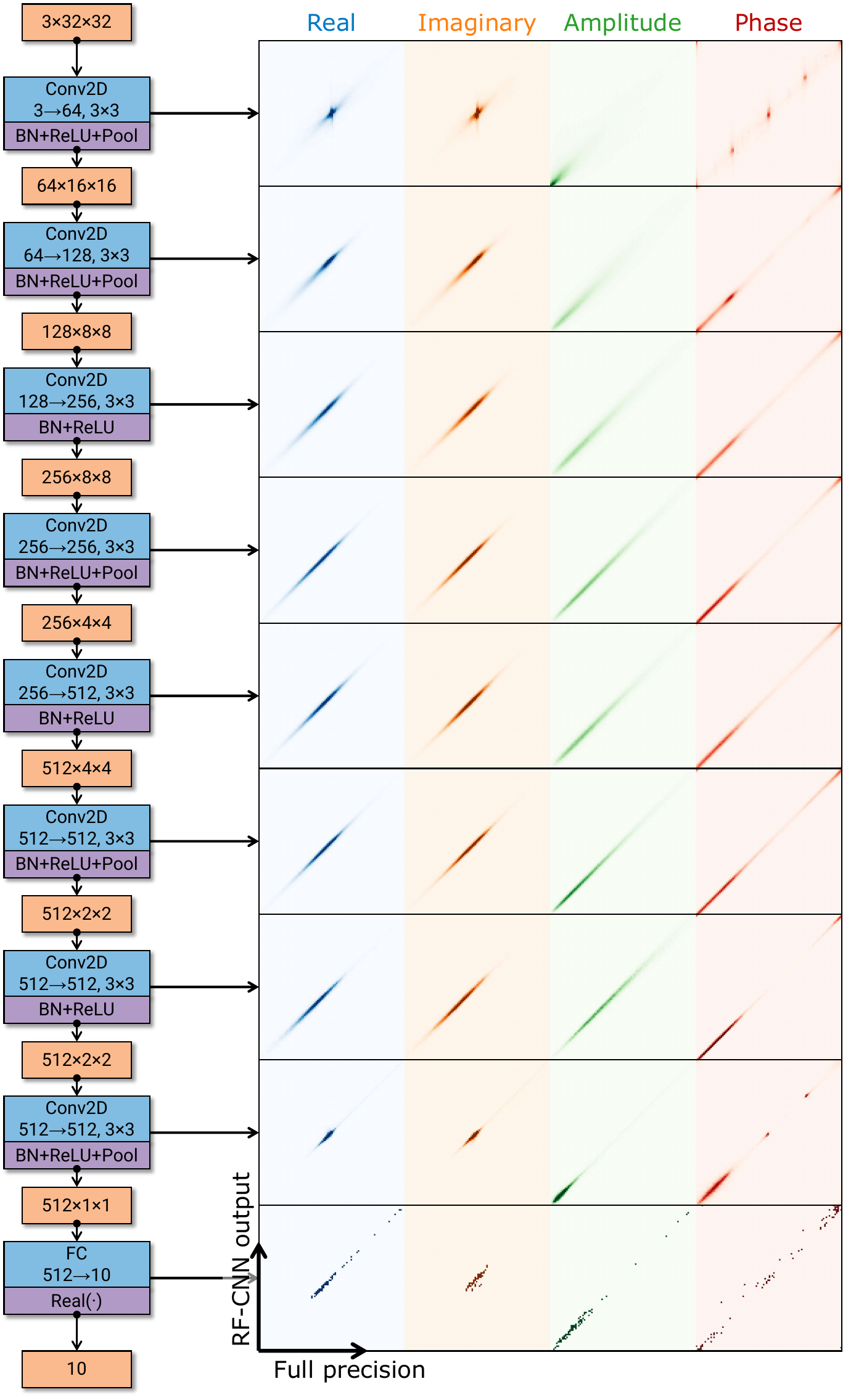}
    \caption{The per-layer comparison between the full precision output and {\name} output regarding real/imaginary/amplitude/angle on the 6-layer CNN model for the SVHN generation.}
    \label{fig:supplementary-scatter-DL-svhn}
\end{figure*}

\begin{figure*}[!t]
    \centering
    \includegraphics[width=0.6\columnwidth]{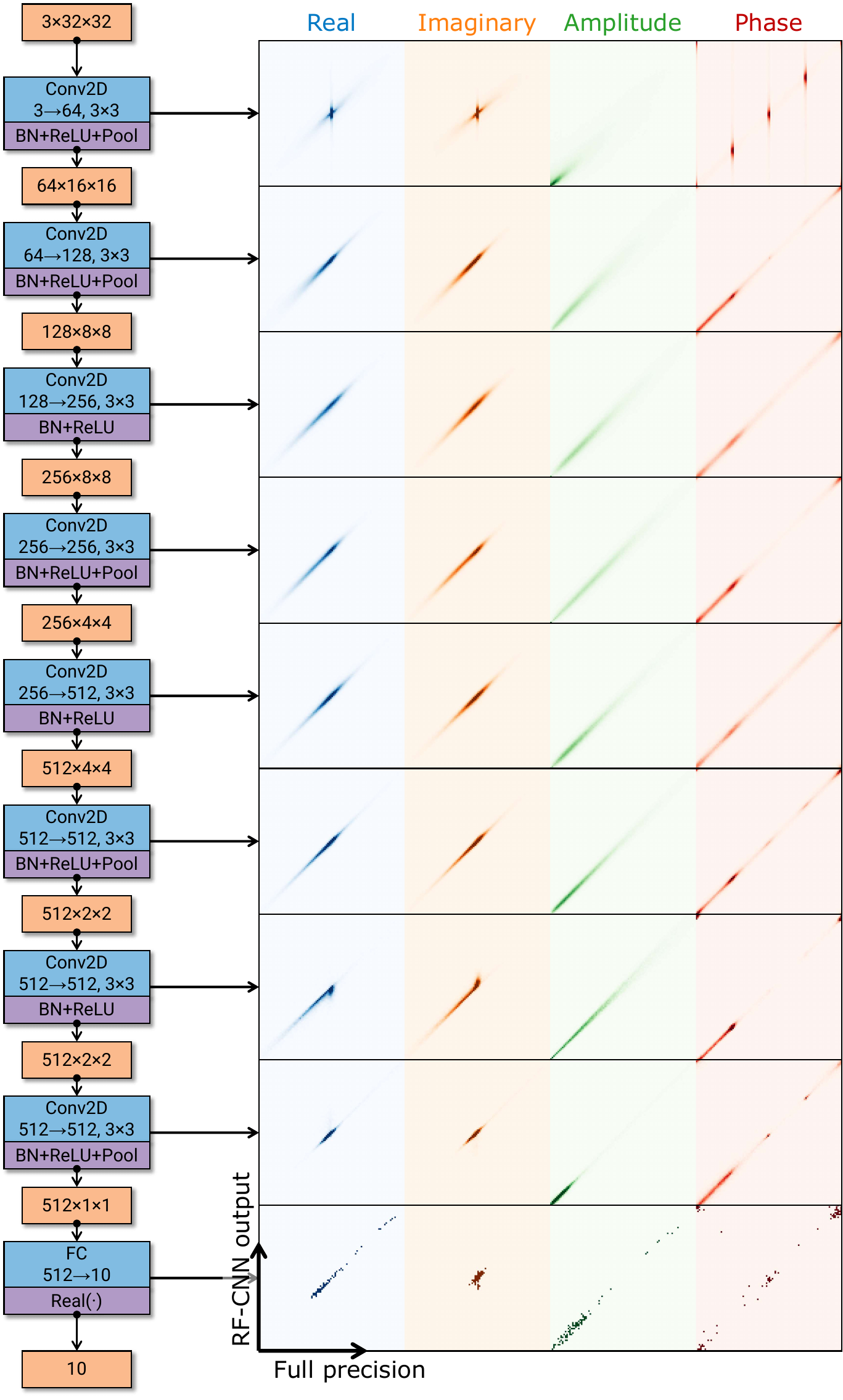}
    \caption{The per-layer comparison between the full precision output and {\name} output regarding real/imaginary/amplitude/angle on the 6-layer CNN model for the CIFAR-10 generation.}
    \label{fig:supplementary-scatter-DL-cifar10}
\end{figure*}

\begin{figure*}[!t]
    \centering
    \includegraphics[width=0.6\columnwidth]{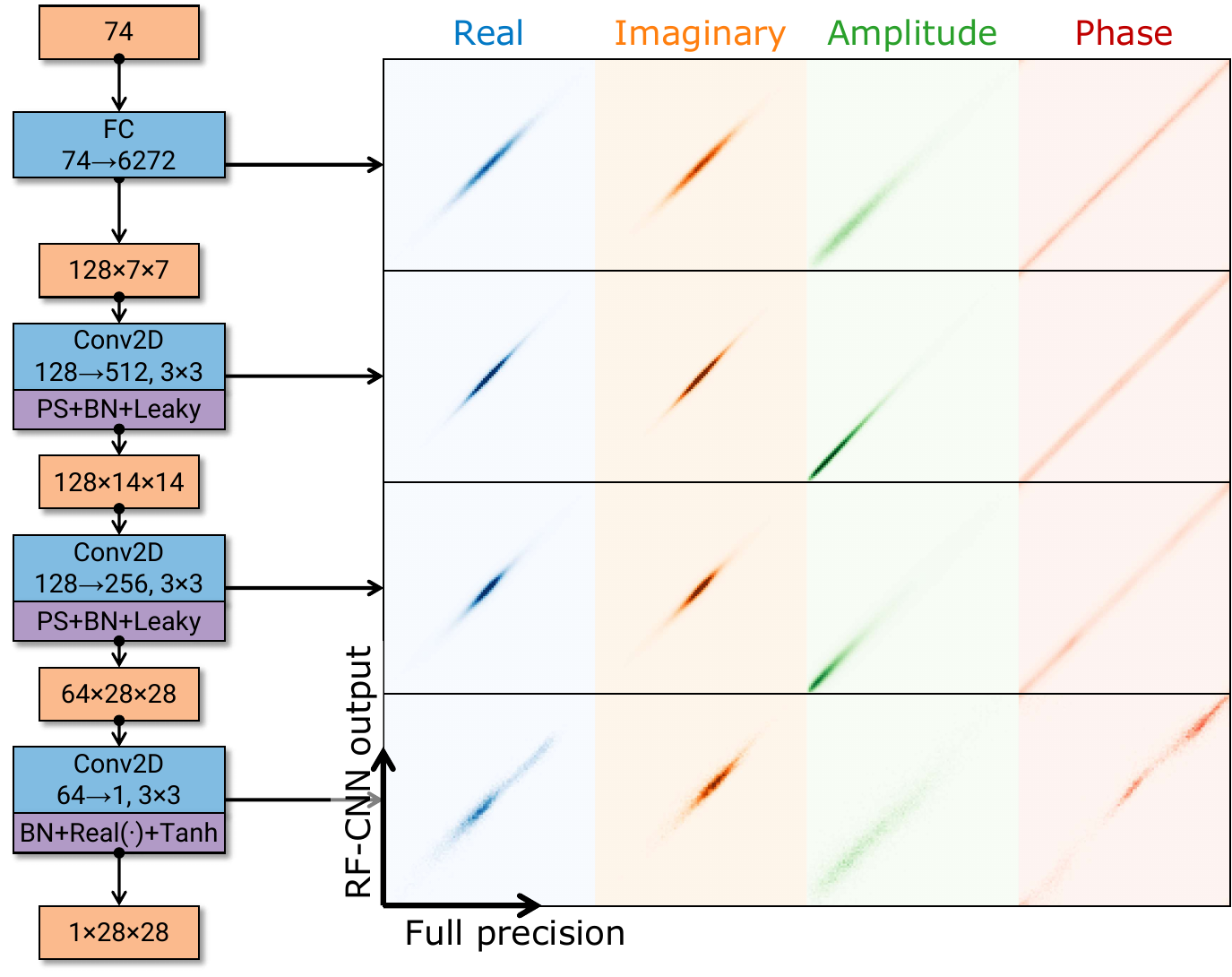}
    \caption{The per-layer comparison between the full precision output and {\name} output regarding real/imaginary/amplitude/angle on the 6-layer CNN model for the MNIST generation.}
    \label{fig:supplementary-scatter-DL-mnist}
\end{figure*}

\begin{figure*}[!t]
    \centering
    \includegraphics[width=0.6\columnwidth]{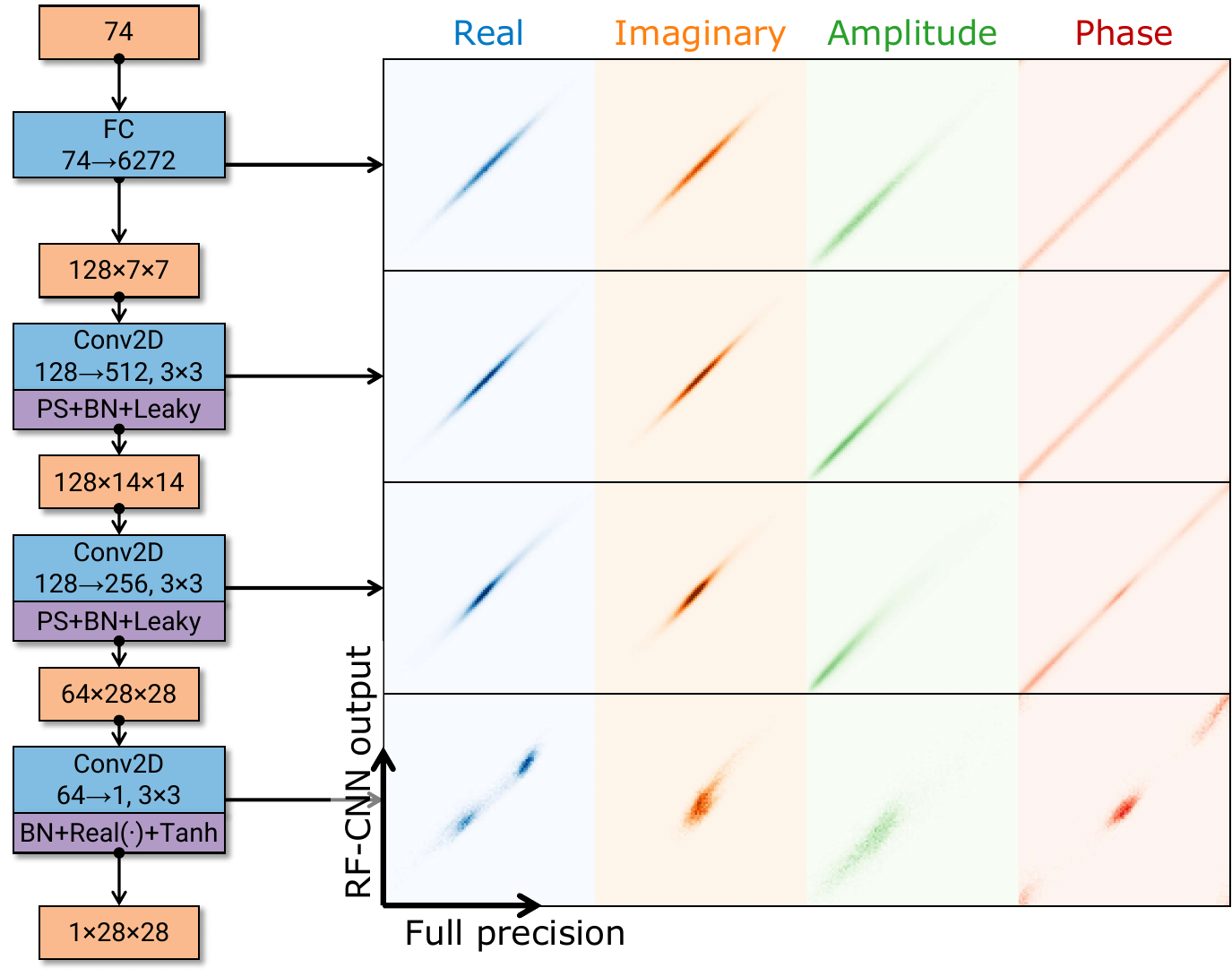}
    \caption{The per-layer comparison between the full precision output and {\name} output regarding real/imaginary/amplitude/angle on the 6-layer CNN model for the FMNIST generation.}
    \label{fig:supplementary-scatter-DL-fmnist}
\end{figure*}

\begin{figure*}[!t]
    \centering
    \includegraphics[width=0.6\columnwidth]{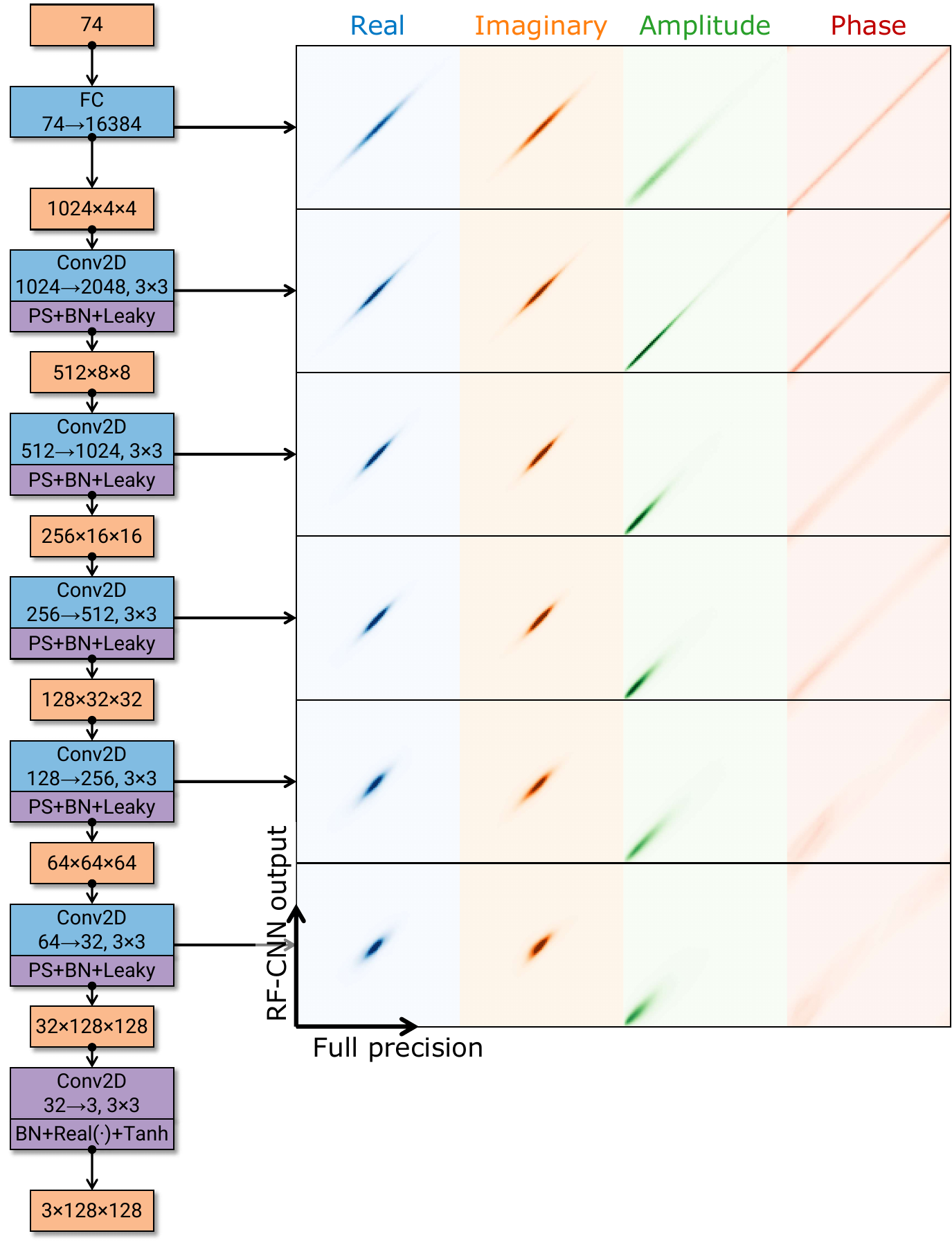}
    \caption{The per-layer comparison between the full precision output and {\name} output regarding real/imaginary/amplitude/angle on the 7-layer CNN model for the CelebA generation.}
    \label{fig:supplementary-scatter-DL-celeba}
\end{figure*}

\begin{figure*}[!t]
    \centering
    \includegraphics[width=0.9\columnwidth]{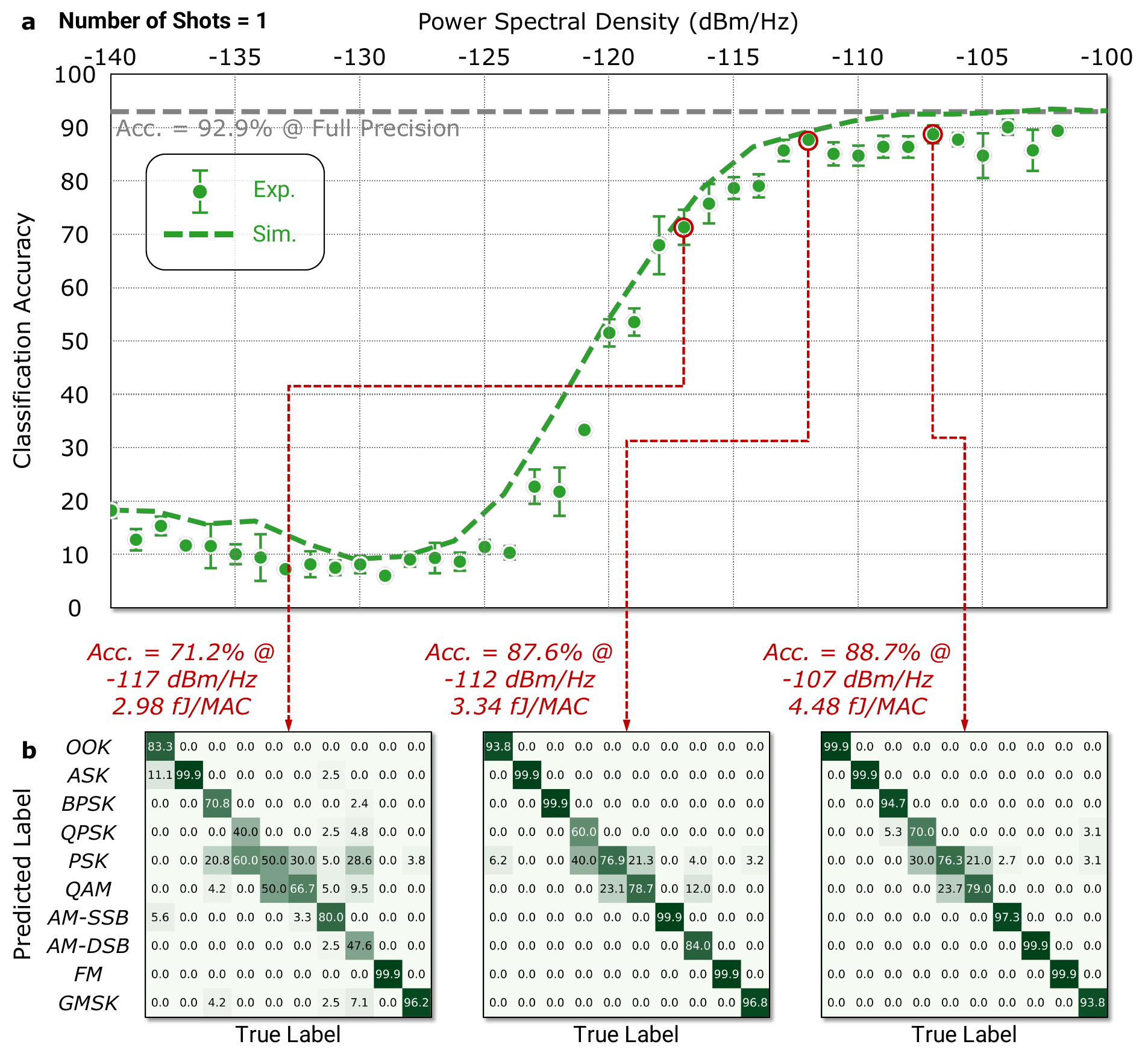}
    \caption{
    \textbf{The evaluation results of the 6-layer CNN model for the one-shot DeepSig classification.}
    \textbf{a}, The classification accuracy over PSD of $\inputTensor$ by experiments and simulations.
    \textbf{b}, The confusion matrices of three selected points at {-107/-112/-117}\thinspace{dBm/Hz}.
    }
    \label{fig:supplementary-classification-energy-deepsig-1shot}
\end{figure*}

\begin{figure*}[!t]
    \centering
    \includegraphics[width=0.9\columnwidth]{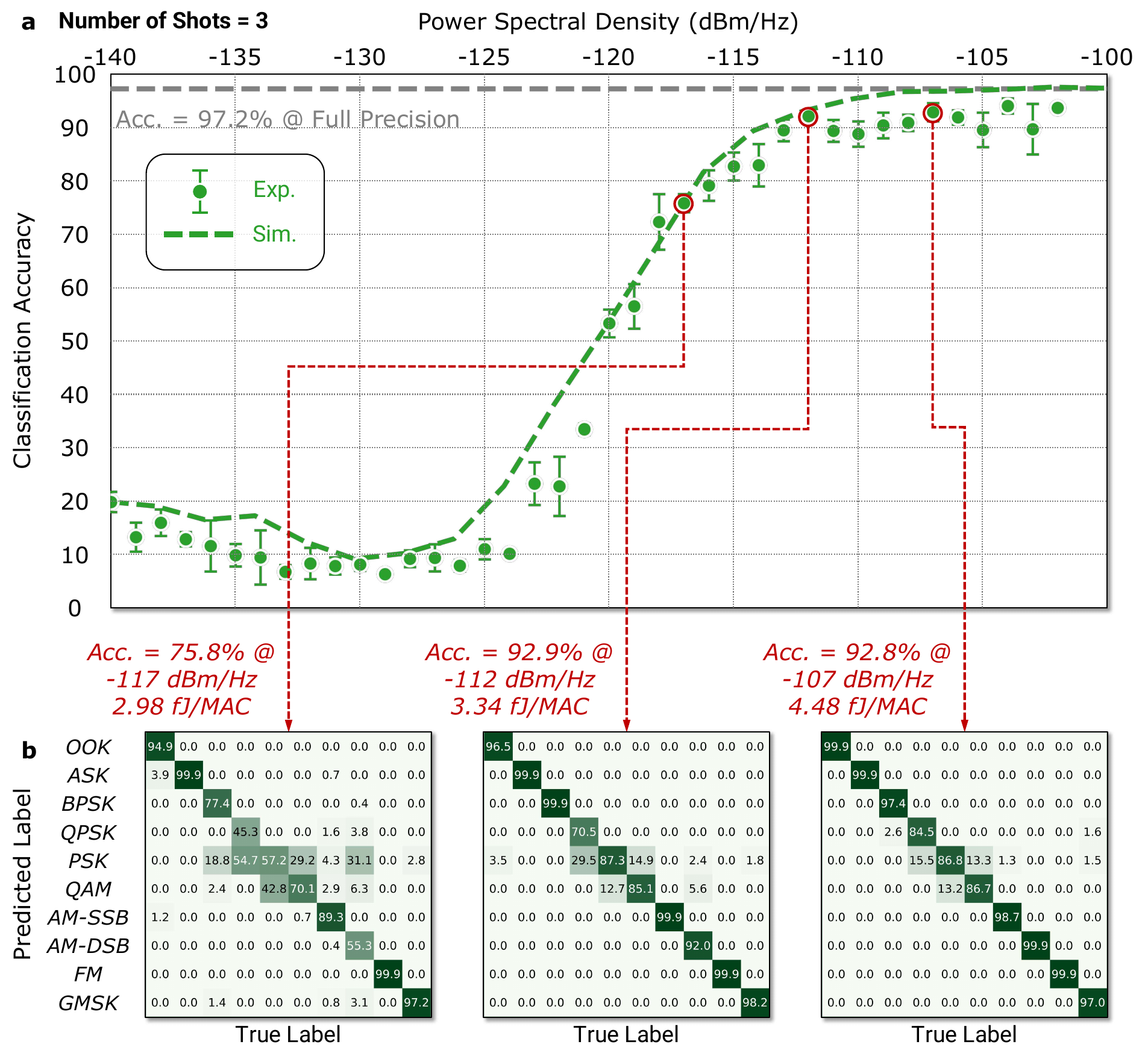}
    \caption{
    \textbf{The evaluation results of the 6-layer CNN model for the three-shot DeepSig classification.}
    \textbf{a}, The classification accuracy over PSD of $\inputTensor$ by experiments and simulations.
    \textbf{b}, The confusion matrices of three selected points at {-107/-112/-117}\thinspace{dBm/Hz}.
    }
    \label{fig:supplementary-classification-energy-deepsig-3shot}
\end{figure*}

\begin{figure*}[!t]
    \centering
    \includegraphics[width=0.9\columnwidth]{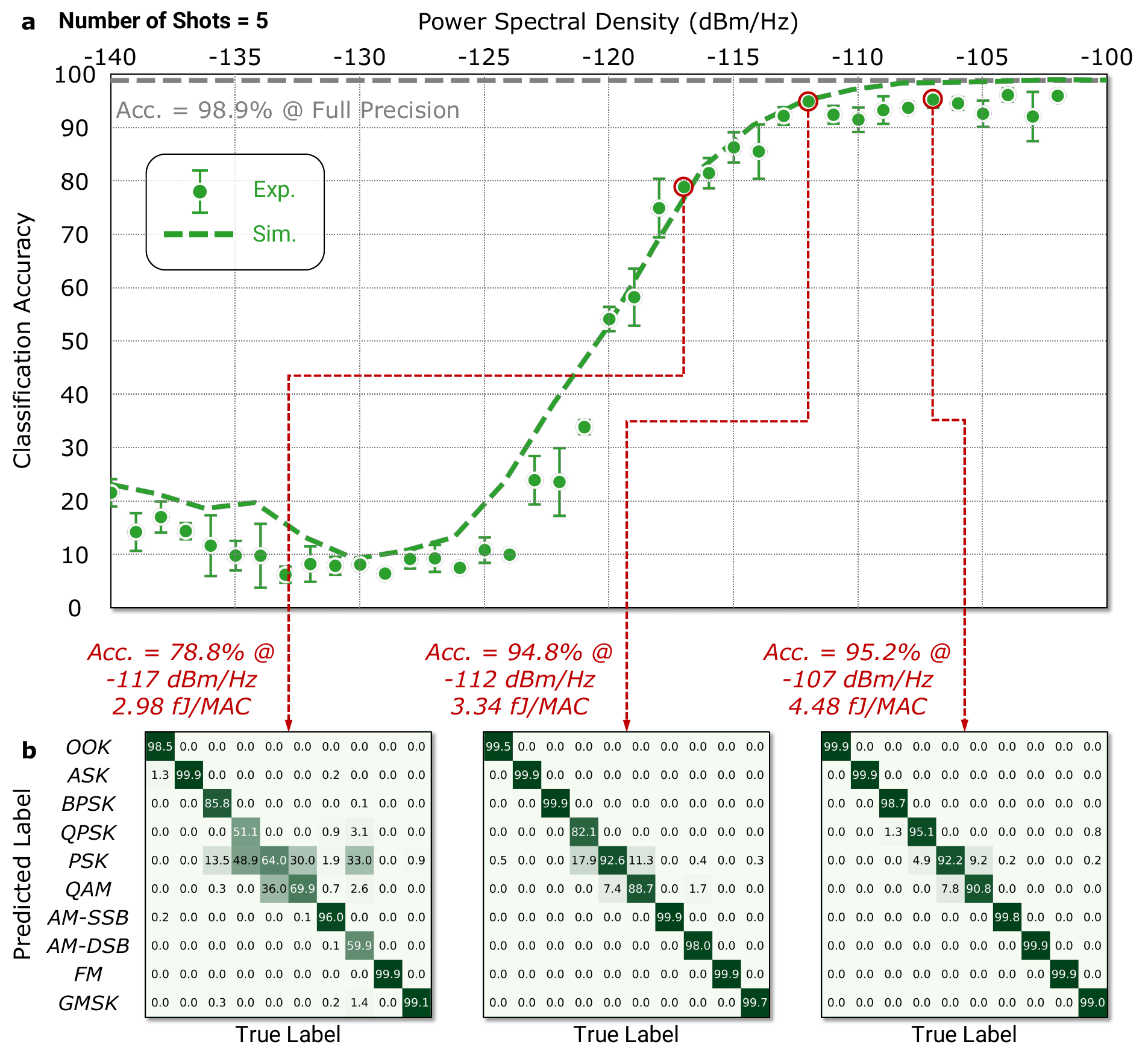}
    \caption{\textbf{The evaluation results of the 6-layer CNN model for the five-shot DeepSig classification.}
    \textbf{a}, The classification accuracy over PSD of $\inputTensor$ by experiments and simulations.
    \textbf{b}, The confusion matrices of three selected points at {-107/-112/-117}\thinspace{dBm/Hz}.}
    \label{fig:supplementary-classification-energy-deepsig-5shot}
\end{figure*}

\begin{figure*}[!t]
    \centering
    \includegraphics[width=0.9\columnwidth]{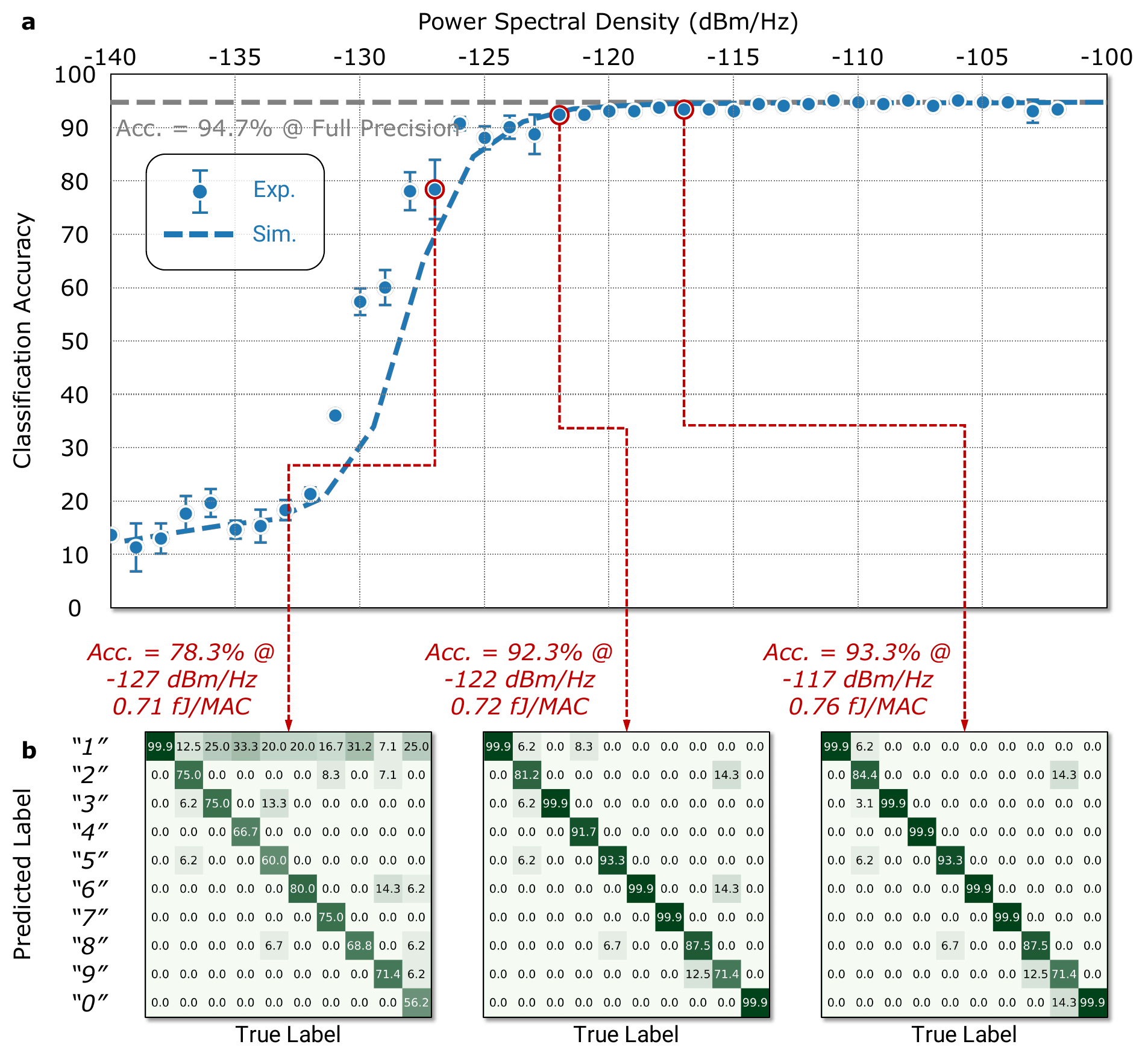}
    \caption{
    \textbf{The evaluation results of the 9-layer CNN model for the SVHN classification.}
    \textbf{a}, The classification accuracy over PSD of $\inputTensor$ by experiments and simulations.
    \textbf{b}, The confusion matrices of three selected points at {-117/-122/-127}\thinspace{dBm/Hz}.
    }
    \label{fig:supplementary-classification-energy-svhn}
\end{figure*}

\begin{figure*}[!t]
    \centering
    \includegraphics[width=0.9\columnwidth]{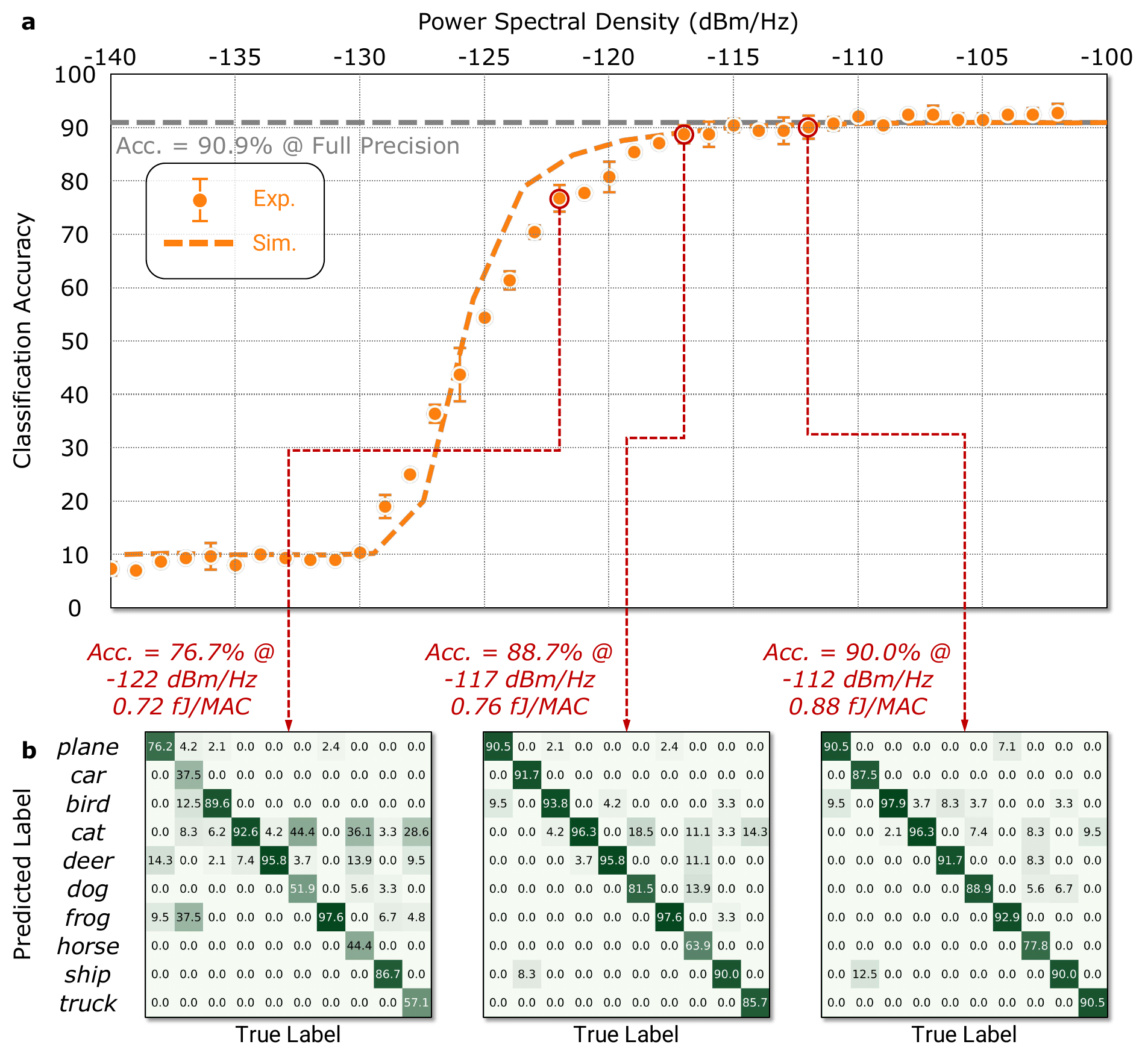}
    \caption{
    \textbf{The evaluation results of the 9-layer CNN model for the CIFAR-10 classification.}
    \textbf{a}, The classification accuracy over PSD of $\inputTensor$ by experiments and simulations.
    \textbf{b}, The confusion matrices of three selected points at {-112/-117/-122}\thinspace{dBm/Hz}.
    }
    \label{fig:supplementary-classification-energy-cifar10}
\end{figure*}

\begin{figure*}[!t]
    \centering
    \includegraphics[width=0.9\columnwidth]{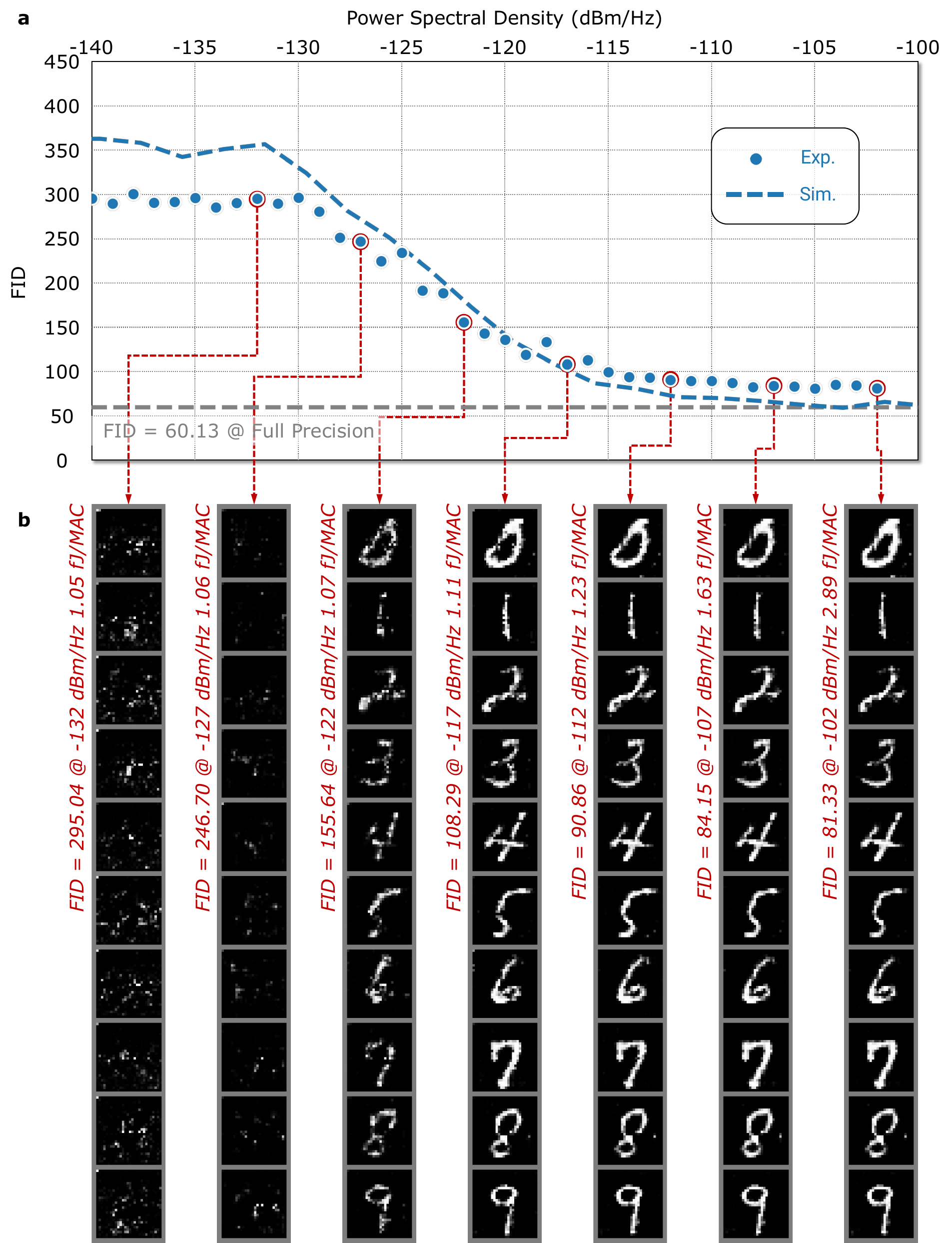}
    \caption{
    \textbf{The evaluation results of the 4-layer CNN model for the MNIST generation.}
    \textbf{a}, The FID over PSD of $\inputTensor$ by experiments and simulations, which is computed over {100} generated images.
    \textbf{b}, The generated examples of digit ``0'' to ``9'' of selected points at {-132/-127/-122/-117/-112/-107/-102}\thinspace{dBm/Hz}.
    }
    \label{fig:supplementary-generative-energy-mnist}
\end{figure*}

\begin{figure*}[!t]
    \centering
    \includegraphics[width=0.9\columnwidth]{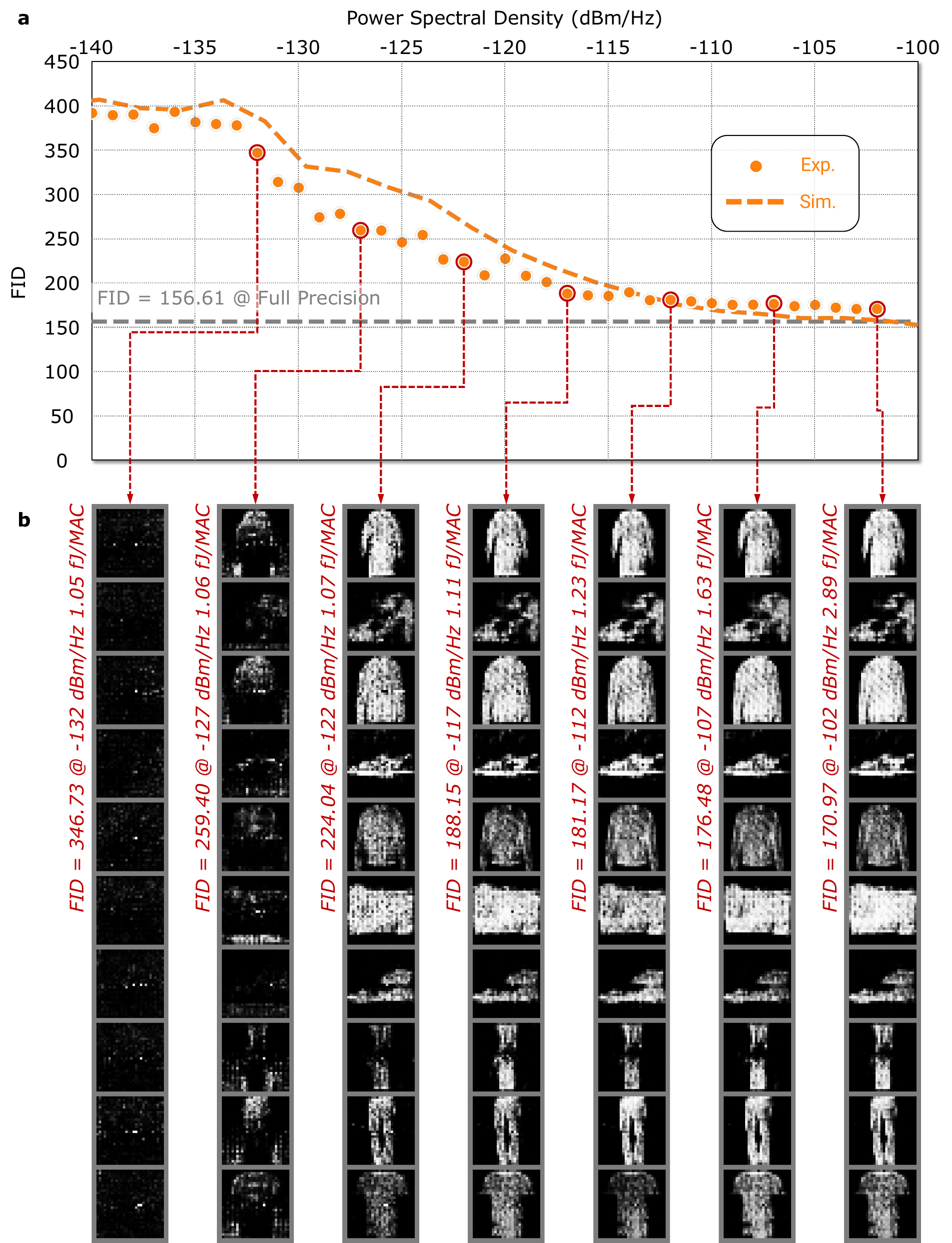}
    \caption{
    \textbf{The evaluation results of the 4-layer CNN model for the FMNIST generation.}
    \textbf{a}, The FID over PSD of $\inputTensor$ by experiments and simulations, which is computed over {100} generated images.
    \textbf{b}, The generated examples of ten fashion products of selected points at {-132/-127/-122/-117/-112/-107/-102}\thinspace{dBm/Hz}.
    }
    \label{fig:supplementary-generative-energy-fmnist}
\end{figure*}

\begin{figure*}[!t]
    \centering
    \includegraphics[width=0.9\columnwidth]{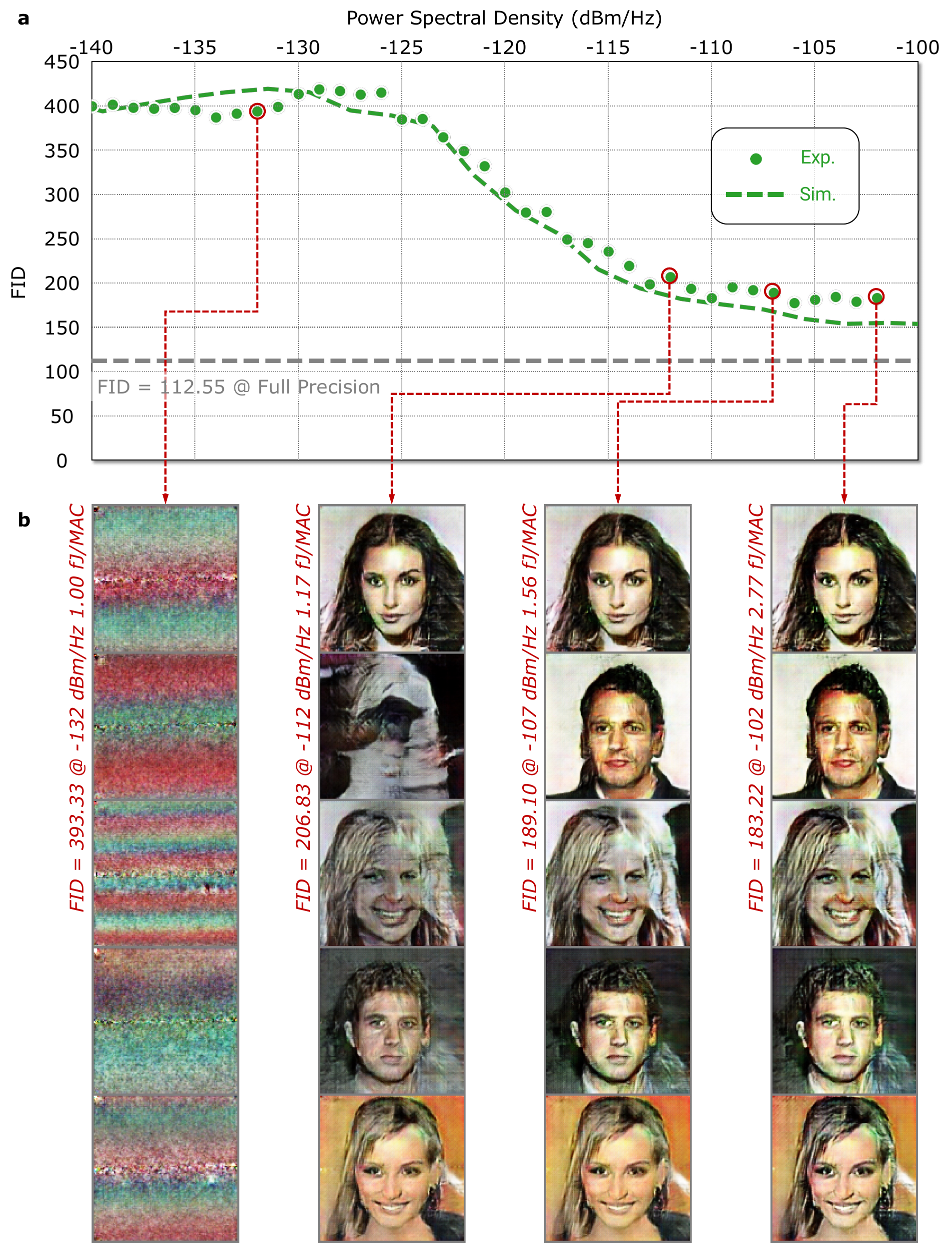}
    \caption{
    \textbf{The evaluation results of the 7-layer CNN model for the CelebA generation.}
    \textbf{a}, The FID over PSD of $\inputTensor$ by experiments and simulations, which is computed over {100} generated images.
    \textbf{b}, The generated examples of human faces of selected points at {-132/-112/-107/-102}\thinspace{dBm/Hz}.
    }
    \label{fig:supplementary-generative-energy-celeba}
\end{figure*}

\clearpage

\begin{table*}[!t]
    \centering
    \includegraphics[width=1.0\columnwidth]{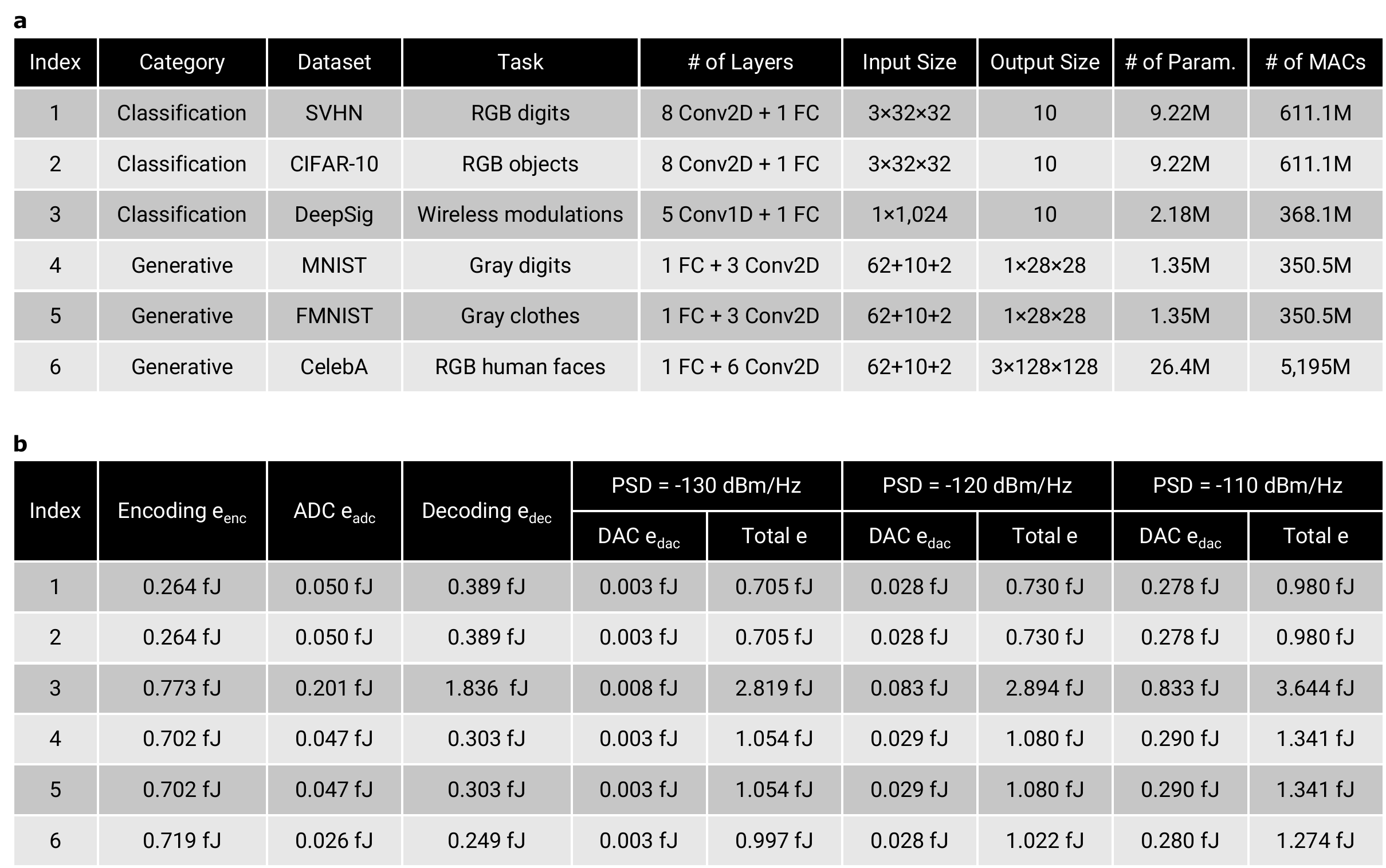}
    \caption{
    \textbf{The overview of the CNN models implemented by {\namebf}'s evaluation.}
    \textbf{a}, The model architecture summary of the six CNN models.
    \textbf{b}, The energy efficiency w.r.t. energy per MAC of the six CNN models.
    }
    \label{tab:supplementary-model-spec}
\end{table*}